\documentclass[journal]{IEEEtran}
\ifCLASSINFOpdf
\else
\fi

\usepackage{comment}
\usepackage{hyperref}
\usepackage[table]{xcolor} 
\usepackage{graphicx}
\usepackage{comment}
\usepackage{subfig}
\usepackage{amssymb}
\usepackage[percent]{overpic}
\usepackage{amsmath}
\usepackage[acronym]{glossaries}

\makeglossaries

\newacronym{dl}{DL}{Deep Learning}
\newacronym{rnn}{RNN}{Recurrent Neural Network}
\newacronym{cnn}{CNN}{Convolutional Neural Network}
\newacronym{gcnn}{GCNN}{Graph Convolutional Neural Network}
\newacronym{snn}{SNN}{Siamese Neural Network}
\newacronym{mot}{MOT}{Multi-Object Tracking}
\newacronym{sot}{SOT}{Single-Object Tracking}
\newacronym{dcf}{DCF}{Discriminative Correlation Filter}
\newacronym{hog}{HOG}{Histogram of Gradients}
\newacronym{vot}{VOT}{Visual Object Tracking}
\newacronym{dnn}{DNN}{Deep Neural Networks}
\newacronym{gan}{GAN}{Generative Adversarial Network}
\newacronym{gsd}{GSD}{Ground Sampling Distance}
\newacronym{ohem}{OHEM}{Online Hard Example Mining}
\newacronym{aerialmpt}{AerialMPT}{Aerial Multi-Pedestrian Tracking}
\newacronym{lstm}{LSTM}{Long Short-Term Memory}
\newacronym{ocr}{OCR}{Optical Character Recognition}
\newacronym{lrn}{LRN}{Local Response Normalization}
\newacronym{se}{SE}{Squeeze-and-Excitation}
\newacronym{relu}{ReLU}{Rectified Linear Unit}
\newacronym{mse}{MSE}{Mean Squared Error}
\newacronym{mae}{MAE}{Mean Absolute Error}
\newacronym{sift}{SIFT}{Scale-Invariant Feature Transform}
\newacronym{mosse}{MOSSE}{Minimum Output Sum of Squared Errors}
\newacronym{kcf}{KCF}{Kernelized Correlation Filter}
\newacronym{iou}{IoU}{Intersection over Union}
\newacronym{fps}{fps}{Frames per Second}
\newacronym{adi}{ADI}{Accumulative Difference Image}
\newacronym{smsot-cnn}{SMSOT-CNN}{Stack of Multiple Single Object Tracking CNNs}
\newacronym{dlr}{DLR}{German Aerospace Center}
\newacronym{dlr-acd}{DLR-ACD}{DLR's Aerial Crowd Dataset}
\newacronym{eot}{EOT}{Euclidean Online Tracking}
\newacronym{tum}{TUM}{Technical University of Munich}
\newacronym{mota}{MOTA}{Multiple-Object Tracker Accuracy}
\newacronym{motp}{MOTP}{Multiple-Object Tracker Precision}
\newacronym{mt}{MT}{Mostly Tracked}
\newacronym{ml}{ML}{Mostly Lost}

\usepackage[backend=bibtex,bibstyle=ieee,citestyle=numeric-comp]{biblatex}
\bibliography{bib}

\newcommand{\etal}{\textit{et al.}}
\newcommand{\AerialMOTPedestrian}{%

\begin{table}

\centering
\caption{Statistics of the AerialMPT dataset.
}

\centering
\resizebox{\columnwidth}{!}{%
\centering
\rowcolors{2}{gray!25}{white}
\begin{tabular}{|c|c|c|c|c|c|c|}
\hline
\rowcolor{gray!50}
\multicolumn{7}{|c|}{\textbf{Train}} \\
\hline
Seq. & Image size & \#Fr. & \#Pedest. & \#Anno. & \#Anno./Fr. & GSD\\
\hline
Bauma1 & 462$\times$306 & 19 & 270 & 4,448 & 234.1 & 11.5  \\ 
Bauma2 & 310$\times$249 & 29 & 148 & 3,627 & 125.1 & 11.5 \\
Bauma4 & 281$\times$243 & 22 & 127  & 2,399 & 109.1 & 11.5 \\
Bauma5 & 281$\times$243 & 17 & 94 & 1,410 & 82.9 & 11.5 \\
Marienplatz & 316$\times$355 & 30 & 215 & 5,158 & 171.9 & 10.5 \\
Pasing1L & 614$\times$366 & 28 & 100 & 2,327 & 83.1 & 10.5 \\
Pasing1R & 667$\times$220 & 16 & 86 & 1,196 & 74.7 & 10.5 \\
OAC & 186$\times$163 & 18 & 92 & 1,287 & 71.5 & 8.0 \\
\hline
\multicolumn{2}{|c|}{\textbf{Total}} & 179 & 1,132  & 21,852  & 122.1 &   \\
\hline
\hline
\rowcolor{gray!50}
\multicolumn{7}{|c|}{\textbf{Test}} \\
\hline
Bauma3 & 611$\times$552 & 16 & 609 & 8,788 & 549.2 & 11.5\\
Bauma6 & 310$\times$249 & 26 & 270 & 5,314 & 204.4 & 11.5\\
Karlsplatz & 283$\times$275 & 27 & 146 & 3,374 & 125.0 & 10.0\\
Pasing7 & 667$\times$220 & 24 & 103 & 2,064 & 86.0 & 10.5\\
Pasing8 & 614$\times$366 & 27 & 83 & 1,932 & 71.6 & 10.5\\
Witt & 353$\times$1,202 & 8 & 185 & 1,416 & 177.0 & 13.0\\
\hline
\multicolumn{2}{|c|}{\textbf{Total}} & 128 & 1,396  & 22,888  & 178.8 & \\
\hline
\end{tabular} 
}

\label{tab:MPTDataset}
\end{table}
}

\newcommand{\KITAISPedestrian}{%
\begin{table}

\centering
\caption{Statistics of the KIT AIS pedestrian dataset.
}

\centering
\resizebox{\columnwidth}{!}{%
\centering
\rowcolors{2}{gray!25}{white}
\begin{tabular}{|c|c|c|c|c|c|c|}
\hline
\rowcolor{gray!50}
\multicolumn{7}{|c|}{\textbf{Train}} \\
\hline
Seq. & Image size & \#Fr. & \#Pedest. & \#Anno. & \#Anno./Fr. & GSD\\
\hline
AA\_Crossing\_01 & 309$\times$487 & 18 & 164 & 2,618 & 145.4 & 15.0 \\ 
AA\_Easy\_01 & 161$\times$168 & 14 & 8 & 112 & 8.0 & 15.0\\
AA\_Easy\_02 & 338$\times$507 & 12 & 16  & 185 & 15.4 & 15.0  \\
AA\_Easy\_Entrance & 165$\times$125 & 19 & 83 & 1,105 & 58.3&  15.0\\
AA\_Walking\_01 & 227$\times$297 & 13 & 40 & 445 & 34.2 & 15.0\\
Munich01 & 509$\times$579 & 24 & 100 & 1,308 & 54.5 & 12.0\\
RaR\_Snack\_Zone\_01 & 443$\times$535 & 4 & 237 & 930 & 232.5 & 15.0  \\
\hline
\multicolumn{2}{|c|}{\textbf{Total}} & 104 &  633  & 6,703  & 64.4 &   \\
\hline
\hline
\rowcolor{gray!50}
\multicolumn{7}{|c|}{\textbf{Test}} \\
\hline
AA\_Crossing\_02 & 322$\times$537 & 13 & 94 & 1,135 & 87.3 & 15.0 \\
AA\_Entrance\_01 &  835$\times$798 & 16 & 973  &
14,031 & 876.9 & 15.0 \\
AA\_Walking\_02 & 516$\times$445 & 17 & 188 & 2,671 & 157.1 & 15.0\\
Munich02 & 702$\times$790 & 31 & 230 & 6,125 & 197.6 & 12.0 \\
RaR\_Snack\_Zone\_02 & 509$\times$474 & 4 & 220 & 865 & 216.2 & 15.0 \\
RaR\_Snack\_Zone\_04 & 669$\times$542 & 4 & 311 & 1,230 & 307.5 & 15.0 \\
\hline
\multicolumn{2}{|c|}{\textbf{Total}} & 85 & 2,016  & 26,057  & 306.5 &  \\
\hline
\end{tabular} 
}
\label{tab:KITAISPED}
\end{table}
}

\newcommand{\KIAISVehicle}{%
\begin{table}

\centering
\caption{Statistics of the KIT AIS vehicle dataset. }

\centering
\resizebox{\columnwidth}{!}{%
\centering
\rowcolors{2}{gray!25}{white}
\begin{tabular}{|c|c|c|c|c|c|c|}
\hline
\rowcolor{gray!50}
\multicolumn{7}{|c|}{\textbf{Train}} \\
\hline
Seq. & Image size & \#Fr. & \#Vehic. & \#Anno. & \#Anno./Fr. & GSD\\
\hline
MunichAutobahn1 & 633$\times$988 & 16  & 16 & 161 &  10.1 & 15.0 \\ 
MunichCrossroad1 & 684$\times$547  & 20 & 30 & 509 & 25.5 & 12.0 \\
MunichStreet1 & 1,764$\times$430  & 25  & 57  & 1,338  & 53.5  & 12.0\\
MunichStreet3 & 1,771$\times$422 & 47  & 88 & 3,071 & 65.3 & 12.0\\
StuttgartAutobahn1 & 767$\times$669 & 23  & 43 & 764 & 33.2  & 17.0\\

\hline
\multicolumn{2}{|c|}{\textbf{Total}} & 131 &  234  & 5,843  & 44.6  & \\
\hline
\hline
\rowcolor{gray!50}
\multicolumn{7}{|c|}{\textbf{Test}} \\
\hline
MunichCrossroad2 & 895$\times$1,036  & 45 & 66 & 2,155 & 47.9 & 12.0 \\
MunichStreet2 & 1,284$\times$377 & 20 & 47  & 746 & 37.3 & 12.0 \\
MunichStreet4 & 1,284$\times$388 & 29 & 68 & 1,519 & 52.4 & 12.0 \\
StuttgartCrossroad1 & 724$\times$708  & 14 & 49 & 554  & 39.6 & 17.0\\
\hline
\multicolumn{2}{|c|}{\textbf{Total}} & 108  & 230  & 4,974  & 46.1 &  \\
\hline
\end{tabular} 
}
\label{tab:KITAISVEH}
\end{table}
}

\newcommand{\preliminaryExperimentsSOT}{%

\begin{table*}
\caption{Results of KCF, MOSSE, CSRT, Median Flow, and Stacked-DCFNet on the KIT AIS pedestrian dataset.}
\resizebox{\textwidth}{!}{%
\rowcolors{2}{gray!20}{white}
\begin{tabular}{|c||c|ccc|ccc|cccc|cccc|ccc|}
\hline

Sequences & \# Images & IDF1$\uparrow$ & IDP$\uparrow$ & IDR$\uparrow$ & Rcll$\uparrow$ & Prcn$\uparrow$ & FAR$\downarrow$ & GT & MT\%$\uparrow$ & PT\%$\uparrow$ & ML\%$\downarrow$ & FP$\downarrow$ & FN$\downarrow$ & IDS$\downarrow$ & FM$\downarrow$ & MOTA$\uparrow$ & MOTP$\uparrow$ & MOTAL$\uparrow$ \\
\hline
\hline

\rowcolor{gray!50}
\multicolumn{19}{|c|}{KCF} \\ 
\hline
AA\_Crossing\_02 & 13 & 8.1 & 8.1 & 8.0 & 9.1 & 9.2 & 78.1 & 94 & 1.1 & 6.4 & 92.5 & 1015 & 1032 & 0  & 8 & -80.4 & 97.3 & -80.4 \\
AA\_Walking\_02 & 17 & 6.5 & 6.3 & 6.7 & 7.8 & 7.3 & 154.9 & 188 & 1.6 & 10.6 & 87.8 & 2633 & 2463 & 3 & 14 & -90.9 & 96.9 & -90.8 \\
Munich02 & 31 & 4.3 & 4.1 & 4.4 & 5.6 & 5.2 & 201.7 & 230 & 0.9 & 3.9 & 95.2 & 6254 & 5781 & 29  & 75 & -97.0 & 62.2 & -96.5 \\
RaR\_Snack\_Zone\_02 & 4 & 29.3 & 29.1 & 29.5 & 29.8 & 29.5 & 154.5 & 220 & 1.8 & 98.2 & 0.0 & 618 & 607 & 0 & 8 & -41.6 & 95.1 & -41.6 \\
RaR\_Snack\_Zone\_04 & 4 & 25.8 & 25.7 & 25.9 & 26.9 & 26.8 & 226.5 & 311 & 0.3 & 99.7 & 0.0 & 906 & 899 & 0  & 11 & -46.7 & 97.9 & -46.7 \\
\hline
Overall & 69 & 9.0 & 8.8 & 9.3 & 10.3 & 9.8 & 165.6 & 1043 & 1.1 & 53.8 & 45.1 & 11426 & 10782 & 32  & 116 & -84.9 & 87.2 & -84.7 \\
\hline
\hline

\rowcolor{gray!50}
\multicolumn{19}{|c|}{MOSSE} \\
\hline
AA\_Crossing\_02 & 13 & 8.0 & 8.1 & 7.9 & 9.1 & 9.2 & 78.1 & 94 & 1.1 & 5.3 & 93.6 & 1015 & 1032 & 0  & 9 & -80.4 & 96.9 & -80.4 \\
AA\_Walking\_02 & 17 & 6.6 & 6.4 & 6.7 & 8.0 & 7.6 & 151.8 & 188 & 1.6 & 10.1 & 88.3 & 2580 & 2458 & 2  & 20 & -88.7 & 95.7 & -88.6 \\
Munich02 & 31 & 4.3 & 4.2 & 4.5 & 5.7 & 5.4 & 199.7 & 230 & 0.9 & 4.3 & 94.8 & 6190 & 5775 & 29  & 78 & -95.8 & 61.9 & -95.4 \\
RaR\_Snack\_Zone\_02 & 4 & 29.4 & 29.2 & 29.6 & 30.4 & 30.0 & 153.2 & 220 & 99.5 & 219 & 0 & 613 & 602 & 0  & 14 & -40.5 & 94.9 & -40.5 \\
RaR\_Snack\_Zone\_04 & 4 & 25.8 & 25.7 & 25.9 & 27.0 & 26.8 & 226.2 & 311 & 0.3 & 99.7 & 0 & 905 & 898 & 0 & 12 & -46.6 & 97.5 & -46.6 \\
\hline
Overall & 69 & 9.1 & 8.9 & 9.3 & 10.5 & 10.0 & 163.8 & 1043 & 0.8 & 54.0 & 45.2 & 11303 & 10765 & 31  & 133 & -85.8 & 86.7 & -83.5 \\
\hline
\hline

\rowcolor{gray!50}
\multicolumn{19}{|c|}{CSRT} \\
AA\_Crossing\_02 & 13 & 12.9 & 13.2 & 12.5 & 15.1 & 15.9 & 69.5 & 94 & 1.1 & 30.9 & 68.0 & 904 & 964 & 10  & 29 & -65.5 & 84.6 & -64.7 \\
AA\_Walking\_02 & 17 & 9.2 & 10.0 & 8.5 & 11 & 12.9 & 116.9 & 188 & 2.7 & 15.4 & 81.9 & 187 & 2378 & 12 & 41 & -63.9 & 88.0 & -63.5 \\
Munich02 & 31 & 9.2 & 9.9 & 8.7 & 10.9 & 12.5 & 151.4 & 230 & 1.8 & 14.3 & 83.9 & 4696 & 5455 & 66  & 137 & -66.8 & 61.2 & -65.8 \\
RaR\_Snack\_Zone\_02 & 4 & 43.2 & 42.0 & 42.5 & 43.8 & 43.3 & 124.2 & 220 & 17.3 & 82.7 & 0.0 & 497 & 486 & 0  & 16 & -13.6 & 87.9 & -13.6 \\
RaR\_Snack\_Zone\_04 & 4 & 45.6 & 45.5 & 45.0 & 47.9 & 47.6 & 162.0 & 311 & 16.7 & 83.3 & 0.0 & 648 & 641 & 3  & 31 & -5.0 & 85.2 & -4.8 \\
\hline
Overall & 69 & 16.0 & 16.9 & 15.2 & 17.5 & 19.4 & 126.5 & 1043 & 9.6 & 51.0 & 39.4 & 8732 & 9924 & 91  & 254 & -55.9 & 78.4 & -55.1 \\
\hline
\hline

\rowcolor{gray!50}
\multicolumn{19}{|c|}{Median Flow} \\
AA\_Crossing\_02 & 13 & 27.3 & 27.3 & 27.4 & 28.5 & 28.3 & 62.8 & 94 & 1.1 & 68.1 & 30.8 & 817 & 812 & 4  & 49 & -43.9 & 74.9 & -43.6 \\
AA\_Walking\_02 & 17 & 10.0 & 9.9 & 10.0 & 11.1 & 11.0 & 141.1 & 188 & 1.6 & 21.3 & 77.1 & 2398 & 2374 & 8  & 16 & -79.0 & 86.3 & -78.7 \\
Munich02 & 31 & 9.2 & 9.0 & 9.4 & 9.9 & 9.5 & 186.4 & 230 & 1.3 & 8.7 & 90.0 & 5778 & 5517 & 10  & 53 & -84.6 & 64.7 & -84.4 \\
RaR\_Snack\_Zone\_02 & 4 & 51.7 & 51.4 & 52.0 & 52.8 & 52.2 & 104.7 & 220 & 8.6 & 91.4 & 0.0 & 419 & 408 & 2  & 14 & 4.2 & 83.7 & 4.3 \\
RaR\_Snack\_Zone\_04 & 4 & 53.1 & 53.0 & 53.3 & 53.9 & 53.6 & 143.5 & 311 & 17.4 & 82.6 & 0.0 & 574 & 567 & 6  & 29 & 6.7 & 83.0 & 7.2 \\
\hline
Overall & 69 & 18.5 & 18.3 & 18.8 & 19.5 & 19.0 & 144.7 & 1043 & 7.7 & 55.8 & 36.5 & 9986 & 9678 & 30  & 161 & -63.8 & 77.7 & -63.5 \\
\hline
\hline

\rowcolor{gray!50}
\multicolumn{19}{|c|}{Stacked-DCFNet} \\
AA\_Crossing\_02 & 13 & 41.9 & 42.4 & 41.3 & 42.7 & 43.9 & 47.8 & 94 & 12.8 & 58.5 & 28.7 & 621 & 650 & 15 & 71 & -13.3 & 74.7 & -12.1 \\
AA\_Walking\_02 & 17 & 31.4 & 31.6 & 31.2 & 32.3 & 32.7 & 104.3 & 188 & 5.9 & 45.7 & 48.4 & 1773 & 1809 & 23 & 184 & -35.0 & 74.1 & -34.2 \\
Munich02 & 31 & 21.2 & 20.6 & 21.9 & 25.0 & 23.6 & 160.4 & 230 & 1.7 & 50.0 & 48.3 & 4974 & 4591 & 97 & 322 & -57.7 & 60.5 & -56.2 \\
RaR\_Snack\_Zone\_02 & 4 & 51.8 & 52.3 & 51.3 & 52.4 & 53.4 & 99.0 & 220 & 22.3 & 74.5 & 3.2 & 396 & 412 & 4 & 35 & 6.1 & 84.0 & 6.5 \\
RaR\_Snack\_Zone\_04 & 4 & 51.8 & 52.6 & 51.0 & 52.1 & 53.7 & 138.0 & 311 & 21.9 & 74.9 & 3.2 & 552 & 589 & 0 & 39 & 7.2 & 83.6 & 7.2 \\
\hline
Overall & 69 & 30.0 & 30.2 & 30.9 & 33.1 & 32.3 & 120.5 & 1043 & 13.8 & 62.6 & 23.6 & 8316 & 8051 & 139 & 651 & -37.3 & 71.6 & -36.1 \\ 
\hline

\end{tabular} }
\label{tab:emprExp}
\end{table*}
}

\newcommand{\preliminaryExperimentsMOT}{%

\begin{table*}
\caption{Results of DeepSORT, SORT, Tracktor++, and SMSOT-CNN on the KIT AIS pedestrian dataset.}
\resizebox{\textwidth}{!}{%
\rowcolors{2}{gray!20}{white}
\begin{tabular}{|c||c|ccc|ccc|cccc|cccc|ccc|}
\hline
Sequences & \# Images & IDF1$\uparrow$ & IDP$\uparrow$ & IDR$\uparrow$ & Rcll$\uparrow$ & Prcn$\uparrow$ & FAR$\downarrow$ & GT & MT\%$\uparrow$ & PT\%$\uparrow$ & ML\%$\downarrow$ & FP$\downarrow$ & FN$\downarrow$ & IDS$\downarrow$ & FM$\downarrow$ & MOTA$\uparrow$ & MOTP$\uparrow$ & MOTAL$\uparrow$ \\
\hline
\hline

\rowcolor{gray!50}
\multicolumn{19}{|c|}{DeepSORT with original settings} \\
\hline
AA\_Crossing\_02 & 13 & 3.1 & 3.1 & 3.1 & 100.0 & 100.0 & 0.0 & 94 & 100.0 & 0.0 & 0.0 & 0 & 0 & 940 & 1 & 17.2 & 99.7 & 99.7 \\
AA\_Walking\_02 & 17 & 7.7 & 7.7 & 7.8 & 100.0 & 98.9 & 1.7 & 188 & 100.0 & 0.0 & 0.0 & 29 & 0 & 2145  & 5 & 18.6 & 99.0 & 98.8 \\
Munich02 & 31 & 9.1 & 8.8 & 9.4 & 100.0 & 92.8 & 15.4 & 230 & 100.0 & 0.0 & 0.0 & 478 & 0 & 4681 & 1 & 15.8 & 64.0 & 92.1 \\
RaR\_Snack\_Zone\_02 & 4 & 21.0 & 20.9 & 21.2 & 100.0 & 98.7 & 2.7 & 220 & 100.0 & 0.0 & 0.0 & 11 & 0 & 351 &  2 & 58.2 & 98.1 & 98.4 \\
RaR\_Snack\_Zone\_04 & 4 & 17.9 & 17.9 & 18.0 & 100.0 & 99.6 & 1.2 & 311 & 100.0 & 0.0 & 0.0 & 5 & 0 & 510 & 0 & 58.1 & 98.6 & 99.4 \\
\hline
Overall & 69 & 10.0 & 9.8 & 10.2 & 100.0 & 95.8 & 7.6 & 1043 & 100.0 & 0.0 & 0.0 & 523 & 0 & 8627  & 9 & 23.9 & 81.1 & 98.6 \\ 
\hline
\hline

\rowcolor{gray!50}
\multicolumn{19}{|c|}{DeepSORT with original settings and doubled bounding box size.} \\
\hline
AA\_Crossing\_02 & 13 & 34.8 & 34.5 & 35.1 & 100.0 & 98.4 & 1.4 & 94 & 100.0 & 0.0 & 0.0 & 18 & 0 & 566  & 1 & 48.5 & 94.3 & 98.2 \\
AA\_Walking\_02 & 17 & 46.6 & 46.0 & 47.1 & 100.0 & 98.8 & 3.6 & 188 & 100.0 & 0.0 & 0.0 & 61 & 0 & 1073 & 5 & 57.5 & 93.1 & 97.6 \\
Munich02 & 31 & 29.5 & 27.6 & 31.5 & 100.0 & 87.7 & 27.7 & 230 & 100.0 & 0.0 & 0.0 & 859 & 0 & 2989  & 1 & 37.2 & 63.9 & 85.9 \\
RaR\_Snack\_Zone\_02 & 4 & 52.2 & 51.9 & 52.5 & 100.0 & 98.9 & 2.5 & 220 & 100.0 & 0.0 & 0.0 & 10 & 0 & 203  & 2 & 75.4 & 95.7 & 98.6 \\
RaR\_Snack\_Zone\_04 & 4 & 61.2 & 61.0 & 61.5 & 100.0 & 99.2 & 2.5 & 311 & 100.0 & 0.0 & 0.0 & 10 & 0 & 242 & 0 & 79.5 & 94.4 & 99.0 \\
\hline
Overall & 69 & 38.4 & 36.9 & 39.9 & 100.0 & 92.6 & 13.9 & 1043 & 100.0 & 0.0 & 0.0 & 958 & 0 & 5073  & 9 & 49.9 & 78.7 & 92.0 \\ 
\hline
\hline

\rowcolor{gray!50}
\multicolumn{19}{|c|}{DeepSORT with IoU threshold of 0.99 and original bounding box size.} \\
\hline
AA\_Crossing\_02 & 13 & 55.0 & 54.4 & 55.6 & 99.0 & 96.9 & 2.8 & 94 & 100.0 & 0.0 & 0.0 & 36 & 11 & 347  & 10 & 65.3 & 83.6 & 95.6 \\
AA\_Walking\_02 & 17 & 63.4 & 62.5 & 64.3 & 99.1 & 96.3 & 6.1 & 188 & 100.0 & 0.0 & 0.0 & 103 & 23 & 557  & 25 & 74.4 & 82.0 & 95.2 \\
Munich02 & 31 & 24.2 & 22.8 & 25.8 & 97.2 & 85.8 & 31.8 & 230 & 99.6 & 0.4 & 0.0 & 985 & 170 & 2737  & 151 & 36.5 & 62.9 & 81.1 \\
RaR\_Snack\_Zone\_02 & 4 & 57.7 & 57.3 & 58.2 & 100.0 & 98.5 & 3.2 & 220 & 100.0 & 0.0 & 0.0 & 13 & 0 & 177  & 2 & 78.0 & 90.4 & 98.2 \\
RaR\_Snack\_Zone\_04 & 4 & 69.1 & 68.7 & 69.5 & 99.9 & 98.8 & 3.7 & 311 & 99.7 & 0.3 & 0.0 & 15 & 1 & 191  & 1 & 83.2 & 87.2 & 98.5 \\
\hline
Overall & 69 & 43.3 & 40.8 & 44.0 & 98.3 & 91.1 & 16.7 & 1043 & 99.8 & 0.2 & 0.0 & 1152 & 205 & 4009  & 189 & 55.4 & 73.7 & 88.7 \\
\hline
\hline

\rowcolor{gray!50}
\multicolumn{19}{|c|}{DeepSORT with IoU threshold of 0.99 and doubled bounding box size.} \\
\hline
AA\_Crossing\_02 & 13 & 93.8 & 92.5 & 95.2 & 99.8 & 96.9 & 2.8 & 94 & 100.0 & 0.0 & 0.0 & 36 & 2 & 45  & 2 & 93.8 & 85.0 & 96.5 \\
AA\_Walking\_02 & 17 & 88.7 & 84.4 & 93.4 & 99.7 & 90.0 & 17.3 & 188 & 100.0 & 0.0 & 0.0 & 295 & 8 & 42  & 12 & 87.0 & 86.4 & 88.6 \\
Munich02 & 31 & 73.1 & 70.9 & 75.3 & 98.9 & 93.2 & 14.2 & 230 & 100.0 & 0.0 & 0.0 & 441 & 67 & 565 & 56 & 82.5 & 62.9 & 91.7 \\
RaR\_Snack\_Zone\_02 & 4 & 90.1 & 89.9 & 90.4 & 99.8 & 99.2 & 1.7 & 220 & 99.1 & 0.9 & 0.0 & 7 & 2 & 37 & 4 & 94.7 & 87.9 & 98.8 \\
RaR\_Snack\_Zone\_04 & 4 & 90.2 & 90.1 & 90.3 & 100.0 & 99.8 & 0.7 & 311 & 100.0 & 0.0 & 0.0 & 3 & 0 & 49  & 0 & 95.8 & 88.4 & 99.6 \\
\hline
Overall & 69 & 82.1 & 80.7 & 83.6 & 99.4 & 96.0 & 7.3 & 1043 & 99.8 & 0.2 & 0.0 & 502 & 75 & 738  & 70 & 89.1 & 74.7 & 95.2 \\ 
\hline
\hline

\rowcolor{gray!50}
\multicolumn{19}{|c|}{DeepSORT with IoU threshold of 0.99, doubled bounding box size, and fine-tuned network on the KIT AIS pedestrian dataset.} \\
\hline
AA\_Crossing\_02 & 13 & 93.1 & 92.7 & 93.4 & 100.0 & 99.3 & 0.6 & 94 & 100.0 & 0.0 & 0.0 & 8 & 0 & 43  & 1 & 95.5 & 85.1 & 99.2 \\
AA\_Walking\_02 & 17 & 93.1 & 92.4 & 93.7 & 99.8 & 98.4 & 2.5 & 188 & 100.0 & 0.0 & 0.0 & 43 & 6 & 42  & 9 & 96.6 & 86.5 & 98.1 \\
Munich02 & 31 & 73.3 & 71.2 & 75.5 & 99.0 & 93.3 & 13.9 & 230 & 100.0 & 0.0 & 0.0 & 432 & 63 & 563  & 54 & 82.7 & 62.9 & 91.9 \\
RaR\_Snack\_Zone\_02 & 4 & 90.1 & 89.9 & 90.4 & 99.8 & 99.2 & 1.7 & 220 & 99.1 & 0.9 & 0.0 & 7 & 2 & 37  & 4 & 94.7 & 87.9 & 98.8 \\
RaR\_Snack\_Zone\_04 & 4 & 90.2 & 90.1 & 90.3 & 100.0 & 99.8 & 0.7 & 311 & 100.0 & 0.0 & 0.0 & 3 & 0 & 49  & 0 & 95.8 & 88.4 & 99.6 \\
\hline
Overall & 69 & 82.4 & 81.0 & 83.8 & 99.4 & 96.0 & 7.1 & 1043 & 99.8 & 0.2 & 0.0 & 493 & 71 & 734  & 68 & 89.2 & 74.7 & 95.3 \\ 
\hline
\hline

\rowcolor{gray!50}
\multicolumn{19}{|c|}{SORT with IoU threshold of 0.99 and original bounding box size.} \\
\hline
AA\_Crossing\_02 & 13 & 55.9 & 55.4 & 56.5 & 99.1 & 97.2 & 5.5 & 94 & 100.0 & 0.0 & 0.0 & 33 & 10 & 343  & 9 & 66.0 & 83.5 & 96.0 \\
AA\_Walking\_02 & 17 & 64.0 & 63.2 & 64.9 & 99.3 & 96.7 & 5.3 & 188 & 100.0 & 0.0 & 0.0 & 90 & 19 & 550 &  21 & 75.3 & 82.0 & 95.8 \\
Munich02 & 31 & 24.6 & 23.6 & 25.8 & 98.0 & 89.7 & 22.2 & 230 & 99.6 & 0.4 & 0.0 & 689 & 122 & 2544 &  108 & 45.2 & 62.8 & 86.7 \\
RaR\_Snack\_Zone\_02 & 4 & 57.7 & 57.3 & 58.2 & 100.0 & 98.5 & 3.2 & 220 & 100.0 & 0.0 & 0.0 & 13 & 0 & 177  & 2 & 78.0 & 90.4 & 98.2 \\
RaR\_Snack\_Zone\_04 & 4 & 69.1 & 68.7 & 69.5 & 99.9 & 98.8 & 3.7 & 311 & 99.7 & 0.3 & 0.0 & 15 & 1 & 191  & 1 & 83.2 & 87.2 & 98.5 \\

\hline
Overall & 69 & 42.9 & 41.8 & 44.2 & 98.7 & 93.4 & 12.2 & 1043 & 99.8 & 0.2 & 0.0 & 840 & 151 & 3805  & 141 & 60.1 & 73.6 & 91.7 \\ 
\hline
\hline

\rowcolor{gray!50}
\multicolumn{19}{|c|}{SORT with IoU threshold of 0.99 and doubled bounding box size.} \\
\hline
AA\_Crossing\_02 & 13 & 93.1 & 92.7 & 93.4 & 100.0 & 99.3 & 0.6 & 94 & 100.0 & 0.0 & 0.0 & 8 & 0 & 45 & 1 & 95.3 & 85.0 & 99.1 \\
AA\_Walking\_02 & 17 & 94.5 & 93.9 & 95.1 & 99.3 & 98.6 & 2.2 & 188 & 100.0 & 0.0 & 0.0 & 37 & 2 & 30 & 6 & 97.4 & 86.5 & 98.5 \\
Munich02 & 31 & 80.4 & 79.6 & 81.3 & 99.3 & 97.2 & 5.7 & 230 & 100.0 & 0.0 & 0.0 & 176 & 42 & 284  & 37 & 91.8 & 63.0 & 96.4 \\
RaR\_Snack\_Zone\_02 & 4 & 90.5 & 90.2 & 90.8 & 99.8 & 99.2 & 1.7 & 220 & 99.1 & 0.9 & 0.0 & 7 & 2 & 34  & 4 & 95.0 & 87.9 & 98.8 \\
RaR\_Snack\_Zone\_04 & 4 & 90.5 & 90.4 & 90.7 & 100.0 & 99.8 & 0.7 & 311 & 100.0 & 0.0 & 0.0 & 3 & 0 & 45  & 0 & 96.1 & 88.4 & 99.6 \\
\hline
Overall & 69 & 86.5 & 85.5 & 87.2 & 99.6 & 98.1 & 3.3 & 1043 & 99.8 & 0.2 & 0.0 & 231 & 46 & 438  & 48 & 94.1 & 74.7 & 97.7 \\ 
\hline
\hline

\rowcolor{gray!50}
\multicolumn{19}{|c|}{Tracktor++} \\
\hline
AA\_Crossing\_02 & 13 & 12.7 & 19.6 & 9.4 & 48.2 & 100.0 & -- & 94 & 20.1 & 51.1 & 28.8 & 0 & 588 & 432  & 107 & 10.1 & 0.13 & --\\
AA\_Walking\_02 & 17 & 10.7 & 27.5 & 6.7 & 23.2 & 95.8 & -- & 188 & 3.2 & 43.1 & 53.7 & 27 & 2050 & 426  & 154 & 6.3 & 0.13 & --\\
Munich02 & 31 & 7.8 & 16.7 & 5.1 & 22.7 & 74.5 & -- & 230 & 2.2 & 41.3 & 56.6 & 746 & 4736 & 965 & 412 & -0.8 & 0.078 & --\\
RaR\_Snack\_Zone\_02 & 4 & 33.8 & 54.5 & 24.5 & 40.2 & 89.5 & -- & 220 & 17.7 & 45.5 & 36.8 & 41 & 517 & 134  & 27 & 20. & 0.091 & --\\
RaR\_Snack\_Zone\_04 & 4 & 32.5 & 50.2 & 24.0 & 42.9 & 89.8 & -- & 311 & 22.2 & 44.1 & 33.7 & 60 & 702 & 231 & 25 & 19.3 & 0.064 & --\\
\hline
Overall & 69 & 13.7 & 27.3 & 9.2 & 28.5 & 85.0 & -- & 1043 & 13.2 & 44.2 & 42.6 & 604 & 8593 & 2188  & 725 & 5.3 & 0.095 & --\\ 
\hline
\hline

\rowcolor{gray!50}
\multicolumn{19}{|c|}{SMSOT-CNN} \\
\hline
AA\_Crossing\_02 & 13 & 49.9 & 49.7 & 50.1 & 52.1 & 51.6 & 42.6 & 94 & 24.5 & 52.1 & 23.4 & 554 & 544 & 11  & 71 & 2.3 & 68.8 & 3.2 \\
AA\_Walking\_02 & 17 & 30.7 & 30.2 & 31.3 & 33.8 & 32.7 & 109.6 & 188 & 15.5 & 38.9 & 45.6 & 1864 & 1767 & 34  & 140 & -32.7 & 68.0 & -36.0 \\
Munich02 & 31 & 23.6 & 22.7 & 24.5 & 28.8 & 26.7 & 156.3 & 230 & 8.6 & 38.3 & 53.1 & 4846 & 4363 & 105  & 316 & -52.1 & 68.4 & -50.4 \\
RaR\_Snack\_Zone\_02 & 4 & 61.6 & 61.4 & 61.8 & 64.4 & 63.9 & 78.5 & 220 & 37.3 & 62.3 & 0.4 & 314 & 308 & 2  & 39 & 27.9 & 77.9 & 28.0 \\
RaR\_Snack\_Zone\_04 & 4 & 61.2 & 61.1 & 61.3 & 63.8 & 63.6 & 112.5 & 311 & 34.4 & 64.6 & 1.0 & 450 & 445 & 5  & 48 & 26.8 & 76.7 & 27.2 \\
\hline
Overall & 69 & 34.0 & 33.2 & 34.9 & 38.2 & 36.4 & 116.4 & 1043 & 25.0 & 52.5 & 22.5 & 8028 & 7427 & 157  & 614 & -29.8 & 71.0 & -28.5 \\
\hline
\hline

\rowcolor{gray!50}
\multicolumn{19}{|c|}{EOT with Euclidean distance of 17 pixels and original bounding box sizes} \\
\hline
AA\_Crossing\_02 & 13 & 94.4 & 94.4 & 94.4 & 95.3 & 95.2 & 4.1 & 94 & 91.5 & 8.5 & 0.0 & 54 & 53 & 4  & 34 & 90.2 & 73.8 & 90.5 \\
AA\_Walking\_02 & 17 & 94.6 & 94.0 & 95.1 & 96.9 & 95.8 & 6.7 & 188 & 96.8 & 2.7 & 0.5 & 114 & 82 & 10  & 63 & 92.3 & 76.6 & 92.6 \\
Munich02 & 31 & 76.0 & 75.8 & 76.2 & 77.0 & 76.5 & 46.6 & 230 & 44.3 & 54.8 & 0.9 & 1446 & 1409 & 15  & 930 & 53.1 & 60.4 & 53.4 \\
RaR\_Snack\_Zone\_02 & 4 & 95.0 & 94.9 & 95.1 & 96.5 & 96.3 & 8.0 & 220 & 87.7 & 12.3 & 0.0 & 32 & 30 & 3  & 16 & 92.5 & 77.6 & 92.8 \\
RaR\_Snack\_Zone\_04 & 4 & 95.2 & 95.1 & 95.2 & 96.3 & 96.3 & 11.5 & 311 & 76.2 & 23.8 & 0.0 & 46 & 45 & 5  & 31 & 92.2 & 78.6 & 92.5 \\
\hline
Overall & 69 & 85.2 & 84.9 & 85.5 & 86.5 & 86.0 & 24.5 & 1043 & 80.2 & 19.6 & 0.2 & 1692 & 1619 & 37  & 1074 & 72.2 & 69.3 & 72.5 \\
\hline

\end{tabular} }
\label{tab:emprExpMOT}
\end{table*}

}

\newcommand{\SMSOTCNNALL}{%

\begin{table*}
\caption{SMSOT-CNN on the KIT~AIS and AerialMPT datasets.}
\resizebox{\textwidth}{!}{%
\rowcolors{2}{gray!20}{white}
\begin{tabular}{|c|c|ccc|ccc|cccc|cccc|ccc|}
\hline
Sequences & \# Images & IDF1$\uparrow$ & IDP$\uparrow$ & IDR$\uparrow$ & Rcll$\uparrow$ & Prcn$\uparrow$ & FAR$\downarrow$ & GT & MT\%$\uparrow$ & PT\%$\uparrow$ & ML\%$\downarrow$ & FP$\downarrow$ & FN$\downarrow$ & IDS$\downarrow$ & FM$\downarrow$ & MOTA$\uparrow$ & MOTP$\uparrow$ & MOTAL$\uparrow$ \\
\hline
\hline
\rowcolor{gray!50}
\multicolumn{19}{|c|}{KIT~AIS Pedestrian Dataset} \\
\hline
AA\_Crossing\_02 & 13 & 49.4 & 49.2 & 49.6 & 51.7 & 51.3 & 42.92 & 94 & 22.4 & 60.6 & 17.0 & 558 & 548 & 15 & 88 & 1.2 & 66.8 & 2.4 \\
AA\_Walking\_02 & 17 & 29.6 & 29.0 & 30.2 & 31.9 & 30.6 & 113.76 & 188 & 9.1 & 45.7 & 45.2 & 1934 & 1820 & 25 & 139 & -41.5 & 65.7 & -40.6 \\
Munich02 & 31 & 20.7 & 19.9 & 21.5 & 24.5 & 22.6 & 165.45 & 230 & 3.5 & 44.3 & 52.2 & 5129 & 4625 & 91 & 271 & -60.7 & 67.1 & -59.3 \\
RaR\_Snack\_Zone\_02 & 4 & 63.1 & 62.9 & 63.4 & 64.2 & 63.7 & 79.0 & 220 & 35.0 & 63.6 & 1.4 & 316 & 310 & 1 & 39 & 27.5 & 78.2 & 27.6 \\
RaR\_Snack\_Zone\_04 & 4 & 63.5 & 63.3 & 63.7 & 65.3 & 64.9 & 108.5 & 311 & 35.0 & 64.0 & 1.0 & 434 & 427 & 3 & 48 & 29.8 & 76.7 & 30.0 \\
\hline
Overall & 69 & 32.5 & 31.7 & 33.4 & 35.7 & 33.9 & 121.32 & 1043 & 22.2 & 56.0 & 21.8 & 8371 & 7730 & 135 & 585 & -35.0 & 70.0 & -33.9 \\ 
\hline
\hline

\rowcolor{gray!50}
\multicolumn{19}{|c|}{AerialMPT Dataset} \\
\hline
Bauma3 & 16 & 29.3 & 28.6 & 30.0 & 34.6 & 33.0 & 385.69 & 609 & 9.9 & 47.1 & 43.0 & 6171 & 5748 & 200 & 458 & -37.9 & 69.1 & -35.7 \\
Bauma6 & 26 & 30.8 & 28.6 & 33.3 & 37.7 & 32.3 & 161.23 & 270 & 12.2 & 57.4 & 30.4 & 4192 & 3311 & 115 & 302 & -43.4 & 67.7 & -41.2 \\
Karlsplatz & 27 & 30.7 & 29.4 & 32.2 & 33.8 & 30.8 & 94.93 & 146 & 6.9 & 58.2 & 34.9 & 2563 & 2233 & 26 & 95 & -42.9 & 67.9 & -42.2 \\
Pasing7 & 24 & 57.7 & 54.5 & 61.3 & 61.9 & 55.1 & 43.42 & 103 & 35.9 & 54.4 & 9.7 & 1042 & 786 & 7 & 136 & 11.1 & 67.6 & 11.4 \\
Pasing8 & 27 & 33.5 & 32.6 & 34.4 & 35.1 & 33.3 & 50.30 & 83 & 8.4 & 54.2 & 37.4 & 1358 & 1253 & 10 & 82 & -35.7 & 67.0 & -35.2 \\
Witt & 8 & 15.8 & 15.7 & 15.9 & 16.4 & 16.2 & 150.38 & 185 & 1.1 & 20.5 & 78.4 & 1203 & 1184 & 1 & 9 & -68.6 & 61.5 & -68.6 \\
\hline
Overall & 128 & 32.0 & 30.7 & 33.4 & 36.6 & 33.6 & 129.13 & 1396 & 10.7 & 47.7 & 41.6 & 16529 & 14515 & 359 & 1082 & -37.2 & 68.0 & -35.6 \\ 
\hline
\hline

\rowcolor{gray!50}
\multicolumn{19}{|c|}{KIT AIS Vehicle Dataset} \\
\hline
MunichStreet02 & 20 & 87.4 & 85.0 & 90.1 & 90.5 & 85.3 & 5.80 & 47 & 87.2 & 8.5 & 4.3 & 116 & 71 & 1 & 7 & 74.8 & 80.6 & 74.9 \\
StuttgartCrossroad01 & 14 & 67.3 & 63.6 & 71.5 & 74.9 & 66.6 & 14.86 & 49 & 57.1 & 30.6 & 12.3 & 208 & 139 & 3 & 17 & 36.8 & 75.3 & 37.3 \\
MunichCrossroad02 & 45 & 50.6 & 49.5 & 51.7 & 53.5 & 51.3 & 24.38 & 66 & 45.5 & 27.3 & 27.2 & 1097 & 1001 & 17 & 41 & 1.9 & 69.4 & 2.6 \\
MunichStreet04 & 29 & 83.5 & 82.4 & 84.7 & 85.8 & 83.6 & 8.83 & 68 & 76.5 & 14.7 & 8.8 & 256 & 215 & 6 & 15 & 68.6 & 79.7 & 68.9 \\
\hline
Overall & 108 & 68.0 & 66.4 & 69.7 & 71.3 & 67.9 & 15.53 & 230 & 65.7 & 20.4 & 13.9 & 1677 & 1426 & 27 & 80 & 37.1 & 75.8 & 37.6 \\ 
\hline

\end{tabular} }

\label{tab:smsotAll}
\end{table*}
}

\newcommand{\aerialmptnetLSTM}{%

\begin{table*}
\caption{AerialMPTNet$_{LSTM}$ on the KIT~AIS and AerialMPT datasets.}
\resizebox{\textwidth}{!}{%
\rowcolors{2}{gray!20}{white}
\begin{tabular}{|c|c|ccc|ccc|cccc|cccc|ccc|}
\hline
Sequences & \# Images & IDF1$\uparrow$ & IDP$\uparrow$ & IDR$\uparrow$ & Rcll$\uparrow$ & Prcn$\uparrow$ & FAR$\downarrow$ & GT & MT\%$\uparrow$ & PT\%$\uparrow$ & ML\%$\downarrow$ & FP$\downarrow$ & FN$\downarrow$ & IDS$\downarrow$ & FM$\downarrow$ & MOTA$\uparrow$ & MOTP$\uparrow$ & MOTAL$\uparrow$ \\
\hline
\hline
\rowcolor{gray!50}
\multicolumn{19}{|c|}{KIT~AIS Pedestrian Dataset - Frozen Weights} \\
\hline
AA\_Crossing\_02 & 13 & 42.0 & 41.8 & 42.2 & 44.8 & 44.5 & 48.92 & 94 & 13.8 & 59.6 & 26.6 & 636 & 626 & 13 & 99 & -12.3 & 68.4 & -11.3 \\
AA\_Walking\_02 & 17 & 34.7 & 34.0 & 35.4 & 37.2 & 35.8 & 104.94 & 188 & 8.0 & 55.3 & 36.7 & 1784 & 1678 & 22 & 227 & -30.4 & 67.4 & -29.7 \\
Munich02 & 31 & 26.0 & 25.1 & 26.9 & 33.1 & 30.8 & 146.81 & 230 & 6.1 & 57.8 & 36.1 & 4551 & 4098 & 191 & 463 & --44.3 & 67.8 & -41.2 \\
RaR\_Snack\_Zone\_02 & 4 & 57.1 & 56.9 & 57.3 & 59.0 & 58.6 & 90.25 & 220 & 29.1 & 69.5 & 1.4 & 361 & 355 & 1 & 42 & 17.1 & 72.9 & 17.2 \\
RaR\_Snack\_Zone\_04 & 4 & 64.7 & 64.4 & 64.9 & 66.3 & 65.9 & 105.25 & 311 & 39.6 & 58.8 & 1.6 & 421 & 415 & 4 & 52 & 31.7 & 73.8 & 32.0 \\
\hline
Overall & 69 & 35.5 & 34.6 & 36.3 & 40.4 & 38.5 & 112.36 & 1043 & 22.0 & 60.3 & 17.7 & 7753 & 7172 & 231 & 883 & -26.0 & 69.3 & -24.1 \\ 
\hline
\hline

\rowcolor{gray!50}
\multicolumn{19}{|c|}{KIT~AIS Pedestrian Dataset - Trainable Weights} \\
\hline
AA\_Crossing\_02 & 13 & 47.1 & 49.9 & 47.3 & 49.6 & 49.2 & 44.77 & 94 & 23.4 & 48.9 & 27.7 & 582 & 572 & 11 & 91 & -2.6 & 68.2 & -1.8 \\
AA\_Walking\_02 & 17 & 39.8 & 39.2 & 40.5 & 41.9 & 40.5 & 96.47 & 188 & 18.6 & 46.8 & 34.6 & 1640 & 1553 & 31 & 215 & -20.7 & 67.2 & -19.6 \\
Munich02 & 31 & 29.6 & 28.6 & 30.8 & 37.1 & 34.5 & 139.10 & 230 & 8.3 & 59.6 & 32.1 & 4312 & 3852 & 221 & 506 & -36.9 & 67.1 & -33.3 \\
RaR\_Snack\_Zone\_02 & 4 & 63.0 & 62.8 & 63.2 & 64.9 & 64.4 & 77.50 & 220 & 37.3 & 60.0 & 2.7 & 310 & 304 & 4 & 31 & 28.6 & 72.2 & 28.9 \\
RaR\_Snack\_Zone\_04 & 4 & 67.6 & 67.5 & 67.8 & 69.1 & 68.8 & 96.50 & 311 & 46.0 & 50.8 & 3.2 & 386 & 380 & 3 & 43 & 37.5 & 73.3 & 37.7 \\
\hline
Overall & 69 & 39.7 & 38.8 & 40.6 & 44.6 & 42.6 & 104.78 & 1043 & 28.9 & 53.8 & 17.3 & 7230 & 6661 & 270 & 886 & -17.8 & 68.8 & -15.5 \\ 
\hline
\hline

\rowcolor{gray!50}
\multicolumn{19}{|c|}{AerialMPT Dataset} \\
\hline
Bauma3 & 16 & 28.3 & 27.7 & 29.0 & 34.6 & 33.0 & 386.00 & 609 & 8.4 & 51.2 & 40.4 & 6176 & 5745 & 246 & 608 & -38.5 & 71.0 & -35.7 \\
Bauma6 & 26 & 33.2 & 31.2 & 35.5 & 39.3 & 34.5 & 152.35 & 270 & 13.0 & 58.5 & 28.5 & 3961 & 3225 & 135 & 387 & -37.8 & 70.1 & -35.3 \\
Karlsplatz & 27 & 48.4 & 47.0 & 50.0 & 51.4 & 48.2 & 68.89 & 146 & 24.7 & 55.5 & 19.8 & 1860 & 1641 & 16 & 140 & -4.2 & 69.7 & -3.8 \\
Pasing7 & 24 & 61.0 & 58.5 & 63.6 & 64.3 & 59.2 & 38.08 & 103 & 35.9 & 56.3 & 7.8 & 914 & 737 & 5 & 127 & 19.8 & 70.5 & 20.0 \\
Pasing8 & 27 & 41.3 & 40.6 & 42.1 & 42.7 & 41.4 & 43.78 & 83 & 18.1 & 50.6 & 31.3 & 1182 & 1108 & 4 & 90 & -18.7 & 69.4 & -18.6 \\
Witt & 8 & 15.6 & 15.5 & 15.7 & 17.3 & 17.1 & 148.75 & 185 & 2.7 & 23.8 & 73.5 & 1190 & 1171 & 3 & 24 & -66.9 & 61.1 & -66.8 \\
\hline
Overall & 128 & 35.7 & 34.5 & 37.0 & 40.5 & 37.7 & 119.40 & 1396 & 12.8 & 49.8 & 37.4 & 15283 & 13627 & 409 & 1376 & -28.1 & 70.1 & -26.3 \\ 
\hline
\hline
\rowcolor{gray!50}
\multicolumn{19}{|c|}{KIT~AIS Vehicle Dataset} \\
\hline
MunichStreet02 & 20 & 81.9 & 79.9 & 84.0 & 84.9 & 80.6 & 7.60 & 47 & 74.5 & 10.6 & 14.9 & 152 & 113 & 4 & 3 & 63.9 & 79.6 & 64.4 \\
StuttgartCrossroad01 & 14 & 65.9 & 62.4 & 69.9 & 72.7 & 65.0 & 15.50 & 49 & 59.2 & 26.5 & 14.3 & 217 & 151 & 2 & 11 & 33.2 & 76.2 & 33.5 \\
MunichCrossroad02 & 45 & 57.7 & 56.0 & 59.5 & 60.6 & 56.9 & 21.93 & 66 & 48.5 & 33.3 & 18.2 & 987 & 850 & 22 & 43 & 13.7 & 69.4 & 14.7 \\
MunichStreet04 & 29 & 88.7 & 88.3 & 89.1 & 89.9 & 89.0 & 5.79 & 68 & 86.8 & 7.4 & 5.8 & 168 & 153 & 2 & 3 & 78.7 & 79.8 & 78.8 \\
\hline
Overall & 108 & 71.6 & 69.8 & 73.4 & 74.5 & 70.9 & 14.11 & 230 & 67.4 & 19.6 & 13.0 & 1524 & 1267 & 30 & 60 & 43.3 & 75.7 & 43.9 \\ 
\hline

\end{tabular} }
\label{tab:arialmptnetLSTM}
\end{table*}
}

\newcommand{\TotalResults}{%
\begin{table*}
\centering
\caption{Overall Performances of Different Tracking Methods on the KIT~AIS and AerialMPT Datasets.}
\resizebox{\textwidth}{!}{%
\rowcolors{2}{gray!25}{white}
\begin{tabular}{|c|ccc|ccc|cccc|cccc|ccc|}
\hline
Methods  &IDF1$\uparrow$ & IDP$\uparrow$ & IDR$\uparrow$ & Rcll$\uparrow$ & Prcn$\uparrow$ & FAR$\downarrow$ & GT & MT\%$\uparrow$ & PT\%$\uparrow$ & ML\%$\downarrow$ & FP$\downarrow$ & FN$\downarrow$ & IDS$\downarrow$ & FM$\downarrow$ & MOTA$\uparrow$ & MOTP$\uparrow$ & MOTAL$\uparrow$ \\  
\hline
\hline
\rowcolor{gray!50}
\multicolumn{18}{|c|}{KIT~AIS Pedestrian Dataset} \\
\hline
KCF  & 9.0 & 8.8 & 9.3 & 10.3 & 9.8 & 165.6 & 1043 & 1.1 & 53.8 & 45.1 & 11426 & 10782 & 32 & \textbf{116} & -84.9 & \textbf{87.2} & -84.7  \\ 
Median Flow  &  18.5 & 18.3 & 18.8 & 19.5 & 19.0 & 144.7 & 1043 & 7.7 & 55.8 & 36.5 & 9986 & 9678 & \textbf{30} & 161 & -63.8 & 77.7 & -63.5 \\ 
CSRT  &  16.0 & 16.9 & 15.2 & 17.5 & 19.4 & 126.5 & 1043 & 9.6 & 51.0 & 39.4 & 8732 & 9924 & 91 & 254 & -55.9 & 78.4 & -55.1 \\ 
MOSSE  &  9.1 & 8.9 & 9.3 & 10.5 & 10.0 & 163.8 & 1043 & 0.8 & 54.0 & 45.2 & 11303 & 10765 & 31 & 133 & -85.8 & 86.7 & -83.5 \\
Tracktor++  &  6.6 & 9.0 & 5.2 & 10.8 & 18.7 & \textbf{81.7} & 1043 & 1.1 & 28.4 & 70.5 & \textbf{5648} & 10723 & 648 & 367 & -41.5 & 40.5 & --   \\ 
Stacked-DCFNet  &  30.0 & 30.2 & 30.9 & 33.1 & 32.3 & 120.5 & 1043 & 13.8 & 62.6 & 23.6 & 8316 & 8051 & 139 & 651 & -37.3 & 71.6 & -36.1  \\ 
SMSOT-CNN  &  32.5 & 31.7 & 33.4 & 35.7 & 33.9 & 121.3 & 1043 & 22.2 & 56.0 & 21.8 & 8371 & 7730 & 135 & 585 & -35.0 & 70.0 & -33.9  \\ \hline
AerialMPTNet$_{LSTM}$ (Ours) &  39.7 & 38.8 & 40.6 & 44.6 & 42.6 & 104.8 & 1043 & \textbf{28.9} & 53.8 & 17.3 & 7230 & 6661 & 270 & 886 & -17.8 & 68.8 & -15.5  \\ 
AerialMPTNet$_{GCNN}$ (Ours) &  37.5 & 36.7 & 38.4 & 42.0 & 40.0 & 109.5 & 1043 & 25.3 & 55.3 & 19.4 & 7555 & 6980 & 259 & 814 & -23.0 & 69.6 & -20.9   \\ 
AerialMPTNet (Ours)  &  \textbf{40.6} & \textbf{39.7} & \textbf{41.5} & \textbf{45.1} & \textbf{43.2} & 103.4 & 1043 & 28.1 & 55.3 & \textbf{16.6} & 7138 & \textbf{6597} & 236 & 897 & \textbf{-16.2} & 69.6 & -\textbf{14.2} \\
AerialMPTNet$_{SE}$ (Ours) & 38.3 & 37.5 & 39.1 & 42.8 & 41.1 & 107.2 & 1043 & 27.4 & 54.5 & 18.1 & 7395 & 6876 & 250 & 818 & -20.7 & 69.9 & -18.7  \\ 
AerialMPTNet$_{OHEM}$ (Ours) & 38.6 & 37.7 & 39.4 & 42.7 & 40.9 & 107.7 & 1043 & 26.1 & \textbf{55.8} & 18.1 & 7435 & 6889 & 254 & 854 & -21.2 & 69.5 & -19.1  \\
\hline
\hline
\rowcolor{gray!50}
\multicolumn{18}{|c|}{AerialMPT Dataset} \\
\hline
KCF  &  11.9 & 11.5 & 12.3 & 13.4 & 12.5 & 167.2 & 1396 & 3.7 & 17.0 & 79.3 & 21407 & 19820 & 86  & 212 & -80.5 & 77.2 & -80.1   \\ 
Median Flow  &  12.2 & 12.0 & 12.4 & 13.1 & 12.7 & 162.0 & 1396 & 1.7 & 20.2 & 78.1 & 20732 & 19883 & \textbf{46} & \textbf{144} & -77.7 & 77.8 & -77.5   \\ 
CSRT  &  16.9 & 16.6 & 17.1 & 20.3 & 19.7 & 148.5 & 1396 & 2.9 & 37.8 & 59.3 & 19011 & 18235 & 426  & 668 & -64.6 & 74.6 & -62.7   \\ 
MOSSE  &  12.1 & 11.7 & 12.4 & 13.7 & 12.9 & 165.7 & 1396 & 3.8 & 17.9 & 78.3 & 21204 & 19749 & 85  & 194 & -79.3 & \textbf{80.0} & -78.9  \\ 
Tracktor++   &  4.0 & 8.8 & 3.1 & 5.0 & 8.7 & \textbf{93.0} & 1396 & 0.1 & 7.6 & 92.3 & \textbf{11907} & 21752 & 399 & 345 & -48.8 & 40.3 & --   \\ 
Stacked-DCFNet  &  28.0 & 27.6 & 28.5 & 31.4 & 30.4 & 128.3 & 1396 & 9.4 & 44.2 & 46.4 & 16422 & 15712 & 322 & 944 & -41.8 & 72.3 & -40.4  \\ 
SMSOT-CNN  &  32.0 & 30.7 & 33.4 & 36.6 & 33.6 & 129.1 & 1396 & 10.7 & 47.7 & 41.6 & 16529 & 14515 & 359 & 1082 & -37.2 & 68.0 & -35.6  \\ \hline
AerialMPTNet$_{LSTM}$ (Ours)  &  35.7 & 34.5 & 37.0 & 40.5 & 37.7 & 119.4 & 1396 & 12.8 & 49.8 & 37.4 & 15283 & 13627 & 409 & 1376 & -28.1 & 70.1 & -26.3 \\
AerialMPTNet$_{GCNN}$(Ours)  & 37.0 & 35.7 & 38.3 & 42.0 & 39.1 & 117.0 & 1396 & 15.6 & 46.0 & 38.4 & 14983 & 13279 & 433 & 1229 & -25.4 & 69.7 & -23.5 \\
AerialMPTNet (Ours)  &  37.8 & 36.5 & 39.3 & 43.1 & 40.0 & 115.5 & 1396 & 15.3 & \textbf{49.9} & \textbf{34.8} & 14782 & 13022 & 436 & 1269 & -23.4 & 69.7 & -21.5  \\ 
AerialMPTNet$_{SE}$ (Ours) & \textbf{38.9} & \textbf{37.5} & \textbf{40.4} & \textbf{44.1} & \textbf{40.9} & 113.8 & 1396 & \textbf{17.0} & 48.1 & 34.9 & 14568 & \textbf{12799} & 430 & 1212 & \textbf{-21.4} & 69.8 & \textbf{-19.6}\\
AerialMPTNet$_{OHEM}$ (Ours) & 37.2 & 35.8 & 38.7 & 42.4 & 39.3 & 117.3 & 1396 & 16.0 & 46.8 & 37.2 & 15016 & 13181 & 430 & 1284 & -25.1 & 69.8 & -23.2 \\
\hline
\hline
\rowcolor{gray!50}
\multicolumn{18}{|c|}{KIT~AIS Vehicle Dataset} \\
\hline
KCF   & 41.3 & 39.0 & 43.9 & 45.6 & 40.4 & 30.9 & 230 & 27.0 & 33.5 & 39.5 & 3339 & 2708 & 53 & 96 & -22.6 & 72.3 & -21.6 \\ 
Median Flow  & 42.0 & 39.5 & 44.9 & 46.3 & 40.8 & 31.0 & 230 & 32.2 & 40.0 & 27.8 & 3348 & 2669 & 23 & 47 & -21.4 & \textbf{82.0} & -21.0 \\
CSRT  & \textbf{76.7} & \textbf{72.1} & \textbf{81.9} & \textbf{83.1} & 73.1 & 14.1 & 230 & \textbf{72.6} & 21.7 & \textbf{5.7} & 1520 & \textbf{841} & 21 & 46 & \textbf{52.1} & 80.7 & \textbf{52.5} \\ 
MOSSE  & 29.0 & 27.4 & 30.8 & 32.4 & 28.8 & 36.8 & 230 & 19.6 & 30.0 & 50.4 & 3977 & 3364 & 56 & 81 & -48.7 & 75.0 & -47.6  \\
Tracktor++  & 55.3 & 66.6 & 47.2 & 57.3 & \textbf{80.7} & \textbf{6.3} & 230 & 30.0 & \textbf{47.4} & 22.6 & \textbf{681} & 2125 & 323 & 204 & 37.1 & 77.4 & -- \\ 
Stacked-DCFNet  & 73.8 & 71.2 & 76.6 & 77.2 & 71.8 & 14.0 & 230 & 69.1 & 15.2 & 15.7 & 1512 & 1133 & \textbf{9} & \textbf{39} & 46.6 & \textbf{82.0} & 46.8 \\ 
SMSOT-CNN  & 68.0 & 66.4 & 69.7 & 71.3 & 67.9 & 15.5 & 230 & 65.7 & 20.4 & 13.9 & 1677 & 1426 & 27 & 80 & 37.1 & 75.8 & 37.6  \\  \hline
AerialMPTNet$_{LSTM}$ (Ours) & 71.6 & 69.8 & 73.4 & 74.5 & 70.9 & 14.1 & 230 & 67.4 & 19.6 & 13.0 & 1524 & 1267 & 30 & 60 & 43.3 & 75.7 & 43.9 \\
AerialMPTNet$_{GCNN}$ (Ours) & 71.1 & 69.4 & 72.9 & 74.1 & 70.6 & 14.2 & 230 & 67.0 & 18.7 & 14.3 & 1536 & 1289 & 22 & 58 & 42.8 & 75.9 & 43.2 \\ 
AerialMPTNet (Ours) &  70.0 & 68.3 & 71.8 & 73.9 & 70.3 & 14.4 & 230 & 66.5 & 20.9 & 12.6 & 1556 & 1299 & 29 & 67 & 42.0 & 76.3 & 42.6  \\
AerialMPTNet$_{SE}$ (Ours) & 70.0 & 68.4 & 71.7 & 73.2 & 69.8 & 14.6 & 230 & 63.5 & 24.8 & 11.7 & 1574 & 1334 & 23 & 84 & 41.1 & 75.6 & 41.5 \\
AerialMPTNet$_{OHEM}$ (Ours) & 71.7 & 70.0 & 73.4 & 74.6 & 71.2 & 13.9 & 230 & 67.0 & 19.6 & 13.4 & 1505 & 1262 & 27 & 66 & 43.8 & 75.5 & 44.3 \\
\hline

\end{tabular} }
\label{tab:overallperformance}

\end{table*}
}

\newcommand{\aerialmptnetGCNN}{%

\begin{table*}
\caption{AerialMPTNet$_{GCNN}$ on the KIT~AIS and AerialMPT datasets.}
\resizebox{\textwidth}{!}{%
\rowcolors{2}{gray!20}{white}
\begin{tabular}{|c|c|ccc|ccc|cccc|cccc|ccc|}
\hline
Sequences & \# Images & IDF1$\uparrow$ & IDP$\uparrow$ & IDR$\uparrow$ & Rcll$\uparrow$ & Prcn$\uparrow$ & FAR$\downarrow$ & GT & MT\%$\uparrow$ & PT\%$\uparrow$ & ML\%$\downarrow$ & FP$\downarrow$ & FN$\downarrow$ & IDS$\downarrow$ & FM$\downarrow$ & MOTA$\uparrow$ & MOTP$\uparrow$ & MOTAL$\uparrow$ \\
\hline
\hline
\rowcolor{gray!50}
\multicolumn{19}{|c|}{KIT~AIS Pedestrian Dataset} \\
\hline
AA\_Crossing\_02 & 13 & 43.5 & 43.3 & 43.7 & 45.5 & 45.1 & 48.4 & 94 & 18.1 & 51.1 & 30.8 & 629 & 619 & 11 & 90 & -10.9 & 68.5 & -10.1 \\
AA\_Walking\_02 & 17 & 35.8 & 35.3 & 36.2 & 38.2 & 37.2 & 101.3 & 188 & 14.9 & 47.9 & 37.2 & 1723 & 1650 & 35 & 204 & -27.6 & 68.1 & -26.3 \\
Munich02 & 31 & 29.1 & 28 & 30.2 & 35.5 & 32.9 & 142.9 & 230 & 8.3 & 53.9 & 37.8 & 4431 & 3951 & 204 & 434 & -40.2 & 68.1 & -36.9 \\
RaR\_Snack\_Zone\_02 & 4 & 55.2 & 55.0 & 55.4 & 56.9 & 56.5 & 94.7 & 220 & 28.2 & 69.5 & 2.3 & 379 & 373 & 3 & 41 & 12.7 & 73.3 & 13.0 \\
RaR\_Snack\_Zone\_04 & 4 & 67.2 & 67 & 67.3 & 68.5 & 68.2 & 98.2 & 311 & 44.4 & 52.1 & 3.5 & 393 & 387 & 6 & 45 & 36.1 & 73.9 & 36.5 \\
\hline
Overall & 69 & 37.5 & 36.7 & 38.4 & 42.0 & 40.0 & 109.5 & 1043 & 25.3 & 55.3 & 19.4 & 7555 & 6980 & 259 & 814 & -23.0 & 69.6 & -20.9 \\ 
\hline
\hline
\rowcolor{gray!50}
\multicolumn{19}{|c|}{AerialMPT Dataset} \\
\hline
Bauma3 & 16 & 29.6 & 28.9 & 30.4 & 36.5 & 34.7 & 376.7 & 609 & 11.3 & 48.3 & 40.4 & 6028 & 5581 & 276 & 550 & -35.2 & 70.0 & -32.1 \\
Bauma6 & 26 & 36.7 & 34.4 & 39.3 & 43.7 & 38.2 & 144.2 & 270 & 20.4 & 50.4 & 29.2 & 3750 & 2994 & 126 & 329 & -29.3 & 70.6 & -26.9 \\
Karlsplatz & 27 & 43.7 & 72.3 & 45.2 & 46.4 & 43.4 & 75.6 & 146 & 15.8 & 63.0 & 21.2 & 2042 & 1809 & 25 & 145 & -14.9 & 68.5 & -14.2 \\
Pasing7 & 24 & 68.6 & 66.0 & 71.4 & 71.6 & 66.1 & 31.5 & 103 & 51.5 & 39.8 & 8.7 & 756 & 857 & 4 & 96 & 34.7 & 71.0 & 34.9 \\
Pasing8 & 27 & 41.2 & 40.4 & 42.1 & 42.7 & 41.0 & 44.0 & 83 & 18.1 & 51.8 & 30.1 & 1188 & 1108 & 2 & 94 & -18.9 & 68.2 & -18.9 \\
Witt & 8 & 14.1 & 14.0 & 14.2 & 15.3 & 15.1 & 152.4 & 185 & 1.6 & 19.5 & 78.9 & 1219 & 1200 & 0 & 15 & -70.8 & 60.8 & -70.8 \\
\hline
Overall & 128 & 37.0 & 35.7 & 38.3 & 42.0 & 39.1 & 117.1 & 1396 & 15.6 & 46.0 & 38.4 & 14983 & 13279 & 433 & 1229 & -25.4 & 69.7 & -23.5 \\  
\hline
\hline
\rowcolor{gray!50}
\multicolumn{19}{|c|}{KIT~AIS Vehicle Dataset} \\
\hline
MunichStreet02 & 20 & 82.6 & 80.5 & 84.7 & 85.4 & 81.1 & 7.4 & 47 & 76.6 & 6.4 & 17.0 & 148 & 109 & 4 & 3 & 65.0 & 79.5 & 65.5 \\
StuttgartCrossroad01 & 14 & 70.0 & 66.5 & 73.8 & 76.7 & 69.1 & 13.6 & 49 & 65.3 & 22.4 & 12.3 & 190 & 129 & 2 & 11 & 42.1 & 75.7 & 42.3 \\
MunichCrossroad02 & 45 & 56.3 & 54.7 & 58.0 & 59.4 & 56.0 & 22.3 & 66 & 44.0 & 34.8 & 21.2 & 1005 & 876 & 14 & 41 & 12.1 & 70.0 & 12.7 \\
MunichStreet04 & 29 & 87.3 & 86.8 & 87.8 & 88.5 & 87.4 & 6.7 & 68 & 83.8 & 8.8 & 7.4 & 193 & 175 & 2 & 3 & 75.6 & 79.7 & 75.7 \\
\hline
Overall & 108 & 71.1 & 69.4 & 72.9 & 74.1 & 70.6 & 14.2 & 230 & 67.0 & 18.7 & 14.3 & 1536 & 1289 & 22 & 58 & 42.8 & 75.9 & 43.2 \\ 

\hline
\end{tabular} }
\label{tab:gcnn}
\end{table*}
}

\newcommand{\aerialMPTNet}{%
\begin{table*}

\centering
\caption{AerialMPTNet on the KIT AIS and AerialMPT datasets.}
\resizebox{\textwidth}{!}{%
\rowcolors{2}{gray!20}{white}
\begin{tabular}{|c|c|ccc|ccc|cccc|cccc|ccc|}
\hline
Sequences & \# Images & IDF1$\uparrow$ & IDP$\uparrow$ & IDR$\uparrow$ & Rcll$\uparrow$ & Prcn$\uparrow$ & FAR$\downarrow$ & GT & MT\%$\uparrow$ & PT\%$\uparrow$ & ML\%$\downarrow$ & FP$\downarrow$ & FN$\downarrow$ & IDS$\downarrow$ & FM$\downarrow$ & MOTA$\uparrow$ & MOTP$\uparrow$ & MOTAL$\uparrow$ \\
\hline
\hline
\rowcolor{gray!50}
\multicolumn{19}{|c|}{KIT~AIS Pedestrian Dataset} \\
\hline
AA\_Crossing\_02 & 13 & 46.7 & 45.6 & 46.9 & 49.3 & 48.8 & 45.1 & 94 & 23.4 & 51.1 & 25.5 & 586 & 576 & 12 & 92 & -3.4 & 69.7 & -2.5 \\
AA\_Walking\_02 & 17 & 41.4 & 40.8 & 42.1 & 43.7 & 42.3 & 93.6 & 188 & 17.0 & 51.6 & 31.4 & 1591 & 1504 & 25 & 231 & -16.8 & 68.5 & -15.9 \\
Munich02 & 31 & 31.2 & 30.2 & 32.3 & 37.8 & 35.3 & 136.8 & 230 & 10.4 & 55.7 & 33.9 & 4240 & 3808 & 192 & 498 & -34.5 & 67.6 & -31.4 \\
RaR\_Snack\_Zone\_02 & 4 & 59.0 & 58.8 & 59.2 & 60.9 & 60.5 & 86.0 & 220 & 33.2 & 65.0 & 1.8 & 344 & 3338 & 4 & 34 & 20.7 & 73.4 & 21.1 \\
RaR\_Snack\_Zone\_04 & 4 & 68.5 & 68.3 & 68.6 & 69.8 & 69.5 & 94.2 & 311 & 45.7 & 51.8 & 2.5 & 377 & 371 & 3 & 42 & 38.9 & 74.2 & 39.1 \\
\hline
Overall & 69 & 40.6 & 39.7 & 41.5 & 45.1 & 43.2 & 103.4 & 1043 & 28.1 & 55.3 & 16.6 & 7138 & 6597 & 236 & 897 & -16.2 & 69.6 & -14.2 \\ 
\hline
\hline
\rowcolor{gray!50}
\multicolumn{19}{|c|}{AerialMPT Dataset} \\
\hline
Bauma3 & 16 & 31.2 & 30.4 & 32.0 & 38.2 & 36.3 & 368.1 & 606 & 11.6 & 51.7 & 36.7 & 5890 & 5435 & 277 & 582 & -32.0 & 70.8 & -28.9 \\
Bauma6 & 26 & 37.2 & 34.8 & 39.9 & 44.2 & 38.6 & 143.7 & 270 & 17.0 & 58.1 & 24.9 & 3736 & 2964 & 123 & 333 & -28.4 & 70.2 & -26.1 \\
Karlsplatz & 27 & 45.6 & 44.2 & 47.1 & 48.6 & 45.6 & 72.4 & 146 & 19.9 & 61.6 & 18.5 & 1954 & 1733 & 25 & 153 & -10.0 & 67.4 & -9.3 \\
Pasing7 & 24 & 67.6 & 64.8 & 70.7 & 71.3 & 65.3 & 32.6 & 103 & 49.5 & 43.7 & 6.8 & 782 & 593 & 5 & 93 & 33.1 & 70.7 & 33.3 \\
Pasing8 & 27 & 39.7 & 38.7 & 40.8 & 41.3 & 39.2 & 45.8 & 83 & 15.7 & 55.4 & 28.9 & 1238 & 1134 & 2 & 83 & -22.9 & 68.9 & -22.8 \\
Witt & 8 & 16.0 & 15.9 & 16.1 & 17.9 & 17.6 & 147.7 & 185 & 2.7 & 24.3 & 73.0 & 1182 & 1163 & 4 & 25 & -65.9 & 60.1 & -65.7 \\
\hline
Overall  & 128 & 37.8 & 36.5 & 39.3 & 43.1 & 40.0 & 115.5 & 1396 & 15.3 & 49.9 & 34.8 & 14782 & 13022 & 436 & 1269 & -23.4 & 69.7 & -21.5 \\   
\hline
\hline
\rowcolor{gray!50}
\multicolumn{19}{|c|}{KIT~AIS Vehicle Dataset} \\
\hline
MunichStreet02 & 20 & 83.2 & 81.1 & 85.4 & 86.3 & 82.0 & 07.1 & 47 & 76.6 & 10.6 & 12.7 & 141 & 102 & 4 & 3 & 66.9 & 80.1 & 67.3 \\
StuttgartCrossroad01 & 14 & 68.4 & 65.0 & 72.2 & 75.3 & 67.8 & 14.14 & 49 & 61.2 & 26.5 & 12.3 & 198 & 137 & 1 & 16 & 39.4 & 76.3 & 39.5 \\
MunichCrossroad02 & 45 & 54.5 & 52.9 & 56.3 & 58.5 & 54.9 & 22.9 & 66 & 43.9 & 37.9 & 18.2 & 1033 & 895 & 20 & 45 & 9.6 & 70.1 & 10.5 \\
MunichStreet04 & 29 & 86.5 & 86.0 & 87.0 & 89.1 & 88.0 & 6.3 & 68 & 85.3 & 7.4 & 7.3 & 184 & 165 & 4 & 3 & 76.8 & 80.2 & 77.0 \\
\hline
Overall & 108 & 70.0 & 68.3 & 71.8 & 73.9 & 70.3 & 14.4 & 230 & 66.5 & 20.9 & 12.6 & 1556 & 1299 & 29 & 67 & 42.0 & 76.3 & 42.6 \\ 
\hline

\end{tabular} 
}
\label{tab:aerialMPTNetresults}

\end{table*}
}

\newcommand{\lossresults}{%
\begin{table*}

\centering
\caption{Comparison of AerialMPTNet trained with the L1 and Huber Losses.
}

\resizebox{\textwidth}{!}{%
\rowcolors{2}{gray!25}{white}
\begin{tabular}{|c|ccc|ccc|cccc|cccc|ccc|}

\hline
Loss & IDF1$\uparrow$ & IDP$\uparrow$ & IDR$\uparrow$ & Rcll$\uparrow$ & Prcn$\uparrow$ & FAR$\downarrow$ & GT & MT\%$\uparrow$ & PT\%$\uparrow$ & ML\%$\downarrow$ & FP$\downarrow$ & FN$\downarrow$ & IDS$\downarrow$ & FM$\downarrow$ & MOTA$\uparrow$ & MOTP$\uparrow$ & MOTAL$\uparrow$ \\  
\hline
\hline
\rowcolor{gray!50}
\multicolumn{18}{|c|}{KIT~AIS Pedestrian Dataset} \\
\hline
L1 & \textbf{40.6} & \textbf{39.7} & \textbf{41.5} & \textbf{45.1} & \textbf{43.2} & \textbf{103.45} & 1043 & \textbf{28.1} & 55.3 & \textbf{16.6} & \textbf{7138} & \textbf{6597} & 236 & 897 & \textbf{-16.2} & \textbf{69.6} & \textbf{-14.2} \\
Huber & 38.8 & 37.9 & 39.7 & 43.1 & 41.1 & 107.42 & 1043 & 25.0 & \textbf{56.5} & 18.5 & 7412 & 6845 & \textbf{212} & \textbf{866} & -20.3 & 69.4 & -18.6  \\
\hline
\hline
\rowcolor{gray!50}
\multicolumn{18}{|c|}{AerialMPT Dataset} \\
\hline
L1 & 37.8 & 36.5 & 39.3 & \textbf{43.1} & \textbf{40.0} & \textbf{115.48} & 1396 & 15.3 & \textbf{49.9} & \textbf{34.8} & \textbf{14782} & \textbf{13022} & 436 & 1269 & \textbf{-23.4} & 69.7 & \textbf{-21.5}  \\
Huber & \textbf{38.0} & \textbf{36.7} & \textbf{39.5} & 43.0 & 39.9 & 115.70 & 1396 & \textbf{15.6} & 48.4 & 36.0 & 14809 & 13051 & \textbf{415} & \textbf{1196} & -23.5 & \textbf{69.9} & -21.7 \\
\hline
\hline
\rowcolor{gray!50}
\multicolumn{18}{|c|}{KIT~AIS Vehicle Dataset} \\
\hline
L1 & \textbf{70.0} & \textbf{68.3} & \textbf{71.8} & \textbf{73.9} & \textbf{70.3} & \textbf{14.41} & 230 & 66.5 & \textbf{20.9} & \textbf{12.6} & \textbf{1556} & \textbf{1299} & \textbf{29} & 67 & \textbf{42.0} & \textbf{76.3} & \textbf{42.6} \\
Huber & 67.2 & 65.5 & 69.0 & 70.6 & 67.1 & 15.98 & 230 & \textbf{67.0} & 17.4 & 15.6 & 1726 & 1461 & 34 & \textbf{65} & 35.2 & 76.1 & 35.9 \\  

\hline
\end{tabular} 
}
\label{tab:huber}
\end{table*}
}

\begin{document}
%
\title{Multiple Pedestrians and Vehicles Tracking in Aerial Imagery: A Comprehensive Study}

\author{Seyed~Majid~Azimi, 
        Maximilian~Kraus, 
        Reza~Bahmanyar, 
        and~Peter~Reinartz~\IEEEmembership{Member,~IEEE,}
\thanks{S.M.~Azimi$^{+,*}$, M.~Kraus$^{\dagger}$, R.~Bahmanyar, and P.~Reinartz are with the Remote Sensing Technology Institute~(IMF), German Aerospace Center (DLR),
	    Wessling, Germany (e-mails: seyedmajid.azimi@dlr.de; maximilian.kraus@dlr.de; reza.bahmanyar@dlr.de; peter.reinartz@dlr.de).}
\thanks{$^{*}$S.M.~Azimi is also affiliated with the Department of Aerospace, Aeronautics and Geodesy, Technical University of Munich, Munich, Germany (e-mail: seyedmajid.azimi@tum.de).}
\thanks{$^{\dagger}$M.~Kraus is also affiliated with the Department of Informatics, Technical University of Munich, Munich, Germany (e-mail: maximilian.kraus@tum.de).}
\thanks{$^{+}$The corresponding author.}
}

\maketitle

\begin{abstract}
In this paper, we address various challenges in multi-pedestrian and vehicle tracking in high-resolution aerial imagery by intensive evaluation of a number of traditional and Deep Learning based Single- and Multi-Object Tracking methods.
We also describe our proposed Deep Learning based Multi-Object Tracking method AerialMPTNet that fuses appearance, temporal, and graphical information using a Siamese Neural Network, a Long Short-Term Memory, and a Graph Convolutional Neural Network module for a more accurate and stable tracking.
Moreover, we investigate the influence of the Squeeze-and-Excitation layers and Online Hard Example Mining on the performance of AerialMPTNet.
To the best of our knowledge, we are the first in using these two for a regression-based Multi-Object Tracking.  
Additionally, we studied and compared the $L1$ and Huber loss functions.
In our experiments, we extensively evaluate AerialMPTNet 
on three aerial Multi-Object Tracking datasets, namely AerialMPT and KIT~AIS pedestrian and vehicle datasets.
Qualitative and quantitative results show that AerialMPTNet outperforms all previous methods for the pedestrian datasets and achieves competitive results for the vehicle dataset. In addition, Long Short-Term Memory and Graph Convolutional Neural Network modules enhance the tracking performance. Moreover, using Squeeze-and-Excitation and Online Hard Example Mining significantly helps for some cases while degrades the results for other cases. In addition, according to the results, $L1$ yields better results with respect to Huber loss for most of the scenarios.
The presented results provide a deep insight into challenges and opportunities of the aerial Multi-Object Tracking domain, paving the way for future research.
\end{abstract}


%
\IEEEpeerreviewmaketitle

\section{Introduction}
%
%
%
%

\IEEEPARstart{V}{isual} Object Tracking, i.e., locating objects in video frames over time, is a dynamic field of research with a wide variety of practical applications such as in autonomous driving, robot aided surgery, security, and safety. 
\begin{figure}[!ht]
    \centering
    \subfloat[]{\includegraphics[width=\columnwidth]{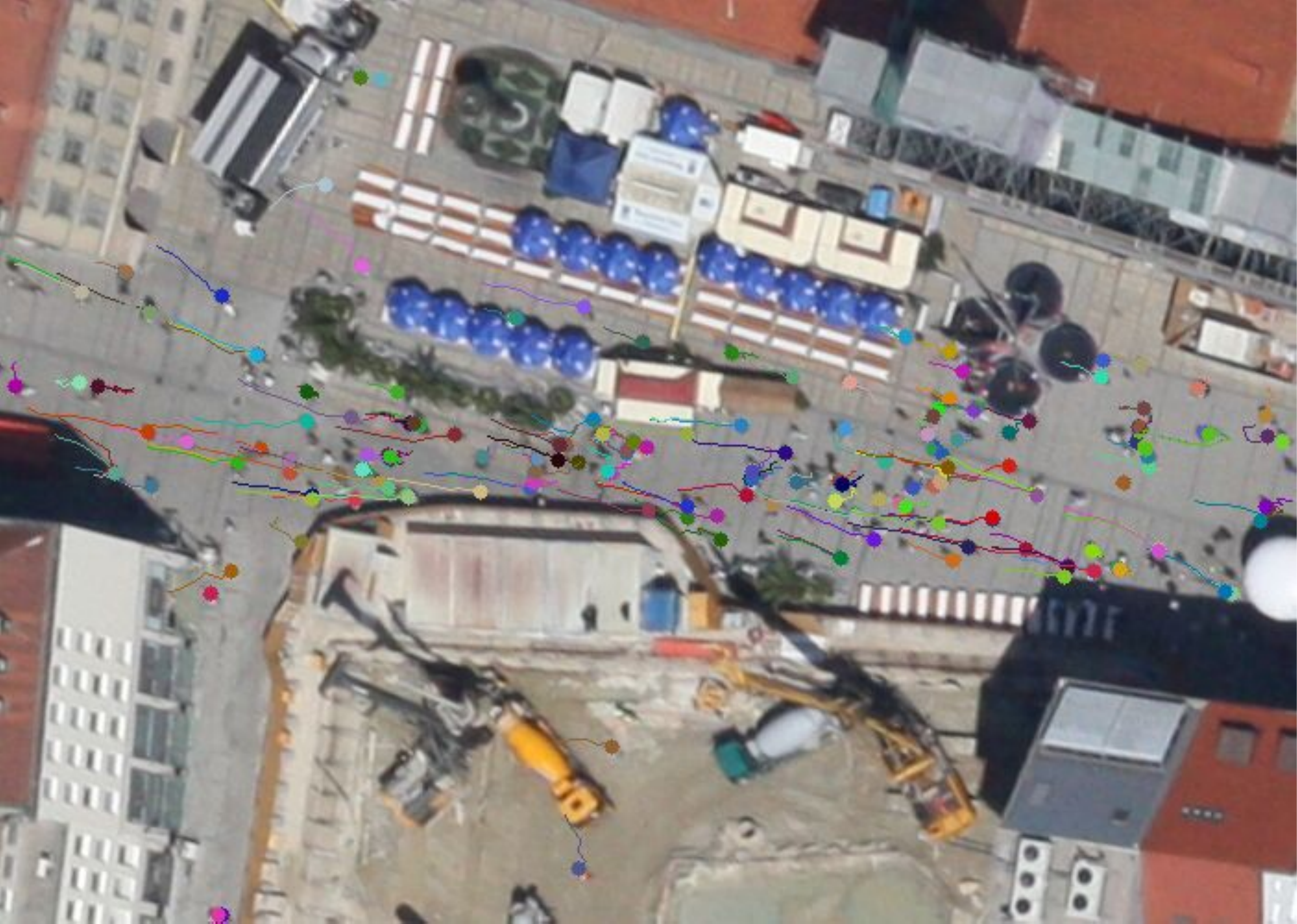}}\\
    \subfloat[]{\includegraphics[width=\columnwidth]{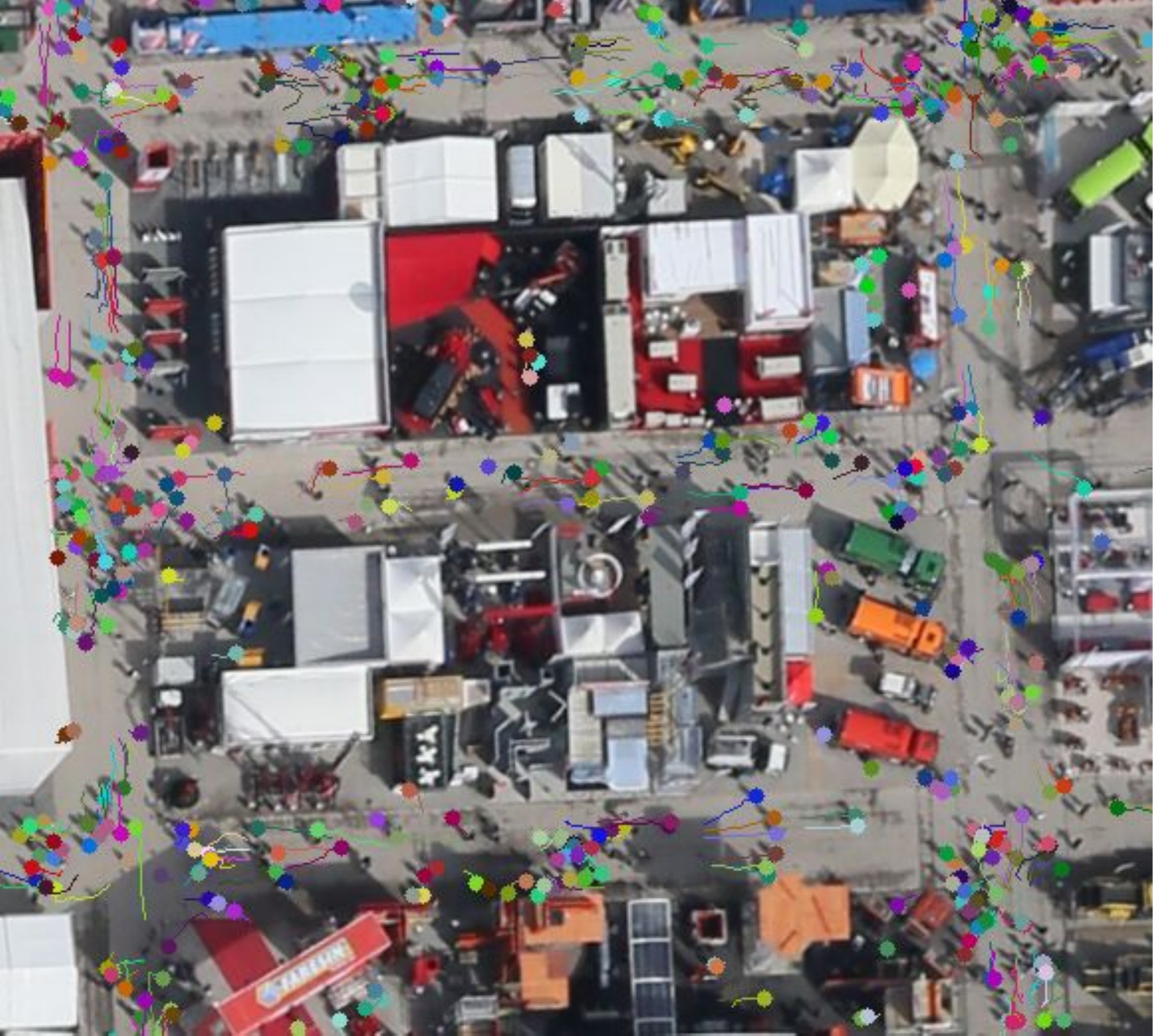}}
    \caption{Multi-Pedestrian tracking results of AerialMPTNet on the frame 18 of the ``Munich02" (top) and frame 10 of the ``Bauma3" (bottom) sequences of the AerialMPT dataset. Different pedestrians are depicted in different colors with the corresponding trajectories.}
    \label{fig:overview}
    \vspace{-5pt}
\end{figure}
The recent advances in machine and deep learning techniques have drastically boosted the performance of \gls{vot} methods by solving long-standing issues such as modeling appearance feature changes and relocating the lost objects~\cite{wojke2017simple, bergmann2019tracking, xiang2015learning, bertinetto2016fully}.
Nevertheless, the performance of the existing \gls{vot} methods is not always satisfactory due to hindrances such as heavy occlusions, difference in scales, background clutter or high-density in the crowded scenes. Thus, developing more sophisticated \gls{vot} methods overcoming these challenges is highly demanded. 
%

The \gls{vot} methods can be categorized into \gls{sot} and \gls{mot} methods, which track single and multiple objects throughout subsequent video frames, respectively. The \gls{mot} scenarios are often more complex than the \gls{mot} because the trackers must handle a larger number of objects in a reasonable time (e.g., ideally real-time).
Most of previous \gls{vot} works using traditional approaches such as Kalman and particle filters~\cite{cuevas2005kalman, cuevas2007particle}, \gls{dcf}~\cite{bolme2010visual}, or silhouette tracking~\cite{boudoukh2009visual}, simplify the tracking procedure by constraining the tracking scenarios with, for example, stationary cameras, limited number of objects, limited occlusions, or absence of sudden background or object appearance changes. These methods usually use handcrafted feature representations (e.g., \gls{hog}~\cite{dalal2005histograms}, color, position) and their target modeling is not dynamic~\cite{marvasti2019deep}. 
In real-world scenarios, however, such constraints are often not applicable and \gls{vot} methods based on these traditional approaches perform poorly.
The rise of \gls{dl} offered several advantages in object detection, segmentation, and classification \cite{he2016deep, szegedy2016rethinking, ren2015faster}. Approaches based on \gls{dl} have also been successfully applied to \gls{vot} problems, and significantly enhancing the performance, especially in unconstrained scenarios. Examples include the \gls{cnn}~\cite{wang2015visual, zhang2016robust}, \gls{rnn}~\cite{kim2018residual}, \gls{snn}~\cite{li2018high, held_learning_2016,bahmanyar2019multiple,kraus2020aerialmptnet}, \gls{gan}~\cite{song2018vital} and several customized architectures~\cite{zhang2017deep}.

Despite the many progress made for \gls{vot} in ground imagery, in the remote sensing domain, \gls{vot} has not been fully exploited, due to the limited available volume of images with high enough resolution and level of details. In recent years, the development of more advanced camera systems and the availability of very high-resolution aerial images have opened new opportunities for research and applications in the aerial \gls{vot} domain ranging from the analysis of ecological systems to aerial surveillance~\cite{remoteSensing2008, everaerts2008use}.

Aerial imagery allows collecting very high-resolution data from wide open areas in a cost- and time-efficient manner. Performing \gls{mot} based on such images (e.g., with \gls{gsd} $<$ 20 cm/pixel) allows us to track and monitor the movement behaviours of multiple small objects such as pedestrians and vehicles for numerous applications such as disaster management and predictive traffic and event monitoring.
However, few works have addressed aerial \gls{mot}~\cite{reilly2010detection, meng2012object, bahmanyar2019multiple}, and the aerial \gls{mot} datasets are rare. 
The large number and the small sizes of moving objects compared to the ground imagery scenarios together with large image sizes, moving cameras, multiple image scale, low frame rates as well as various visibility levels and weather conditions makes \gls{mot} in aerial imagery especially complicate. 
Existing drone or ground surveillance datasets frequently used as \gls{mot} benchmarks, such as MOT16 and MOT17~\cite{milan2016mot16}, are very different from aerial \gls{mot} scenarios with respect to their image and object characteristics. For example, the objects are bigger and the scenes are less crowded, with the objects appearance features usually being discriminative enough to distinguish the objects. Moreover, the videos have higher frame rates and better qualities and contrasts.

In this paper, we aim at investigating various existing challenges in the tracking of multiple pedestrian and vehicles in aerial imagery through intensive experiments with a number of traditional and DL-based \gls{sot} and \gls{mot} methods. This paper extends our recent work~\cite{kraus2020aerialmptnet}, in which we introduced a new \gls{mot} dataset, the so-called \gls{aerialmpt}, as well as a novel DL-based \gls{mot} method, the so-called AerialMPTNet, that fuses appearance, temporal, and graphical information for a more accurate \gls{mot}. In this paper, we also extensively evaluate the effectiveness of different parts of AerialMPTNet and compare it to traditional and state-of-the-art DL-based \gls{mot} methods.
We believe that our paper can promote research on aerial \gls{mot} (esp. for pedestrians and vehicles) by providing a deep insight into its challenges and opportunities.

\begin{figure*}
    \centering
    \subfloat[]{\includegraphics[width=4.45cm, height=4.45cm]{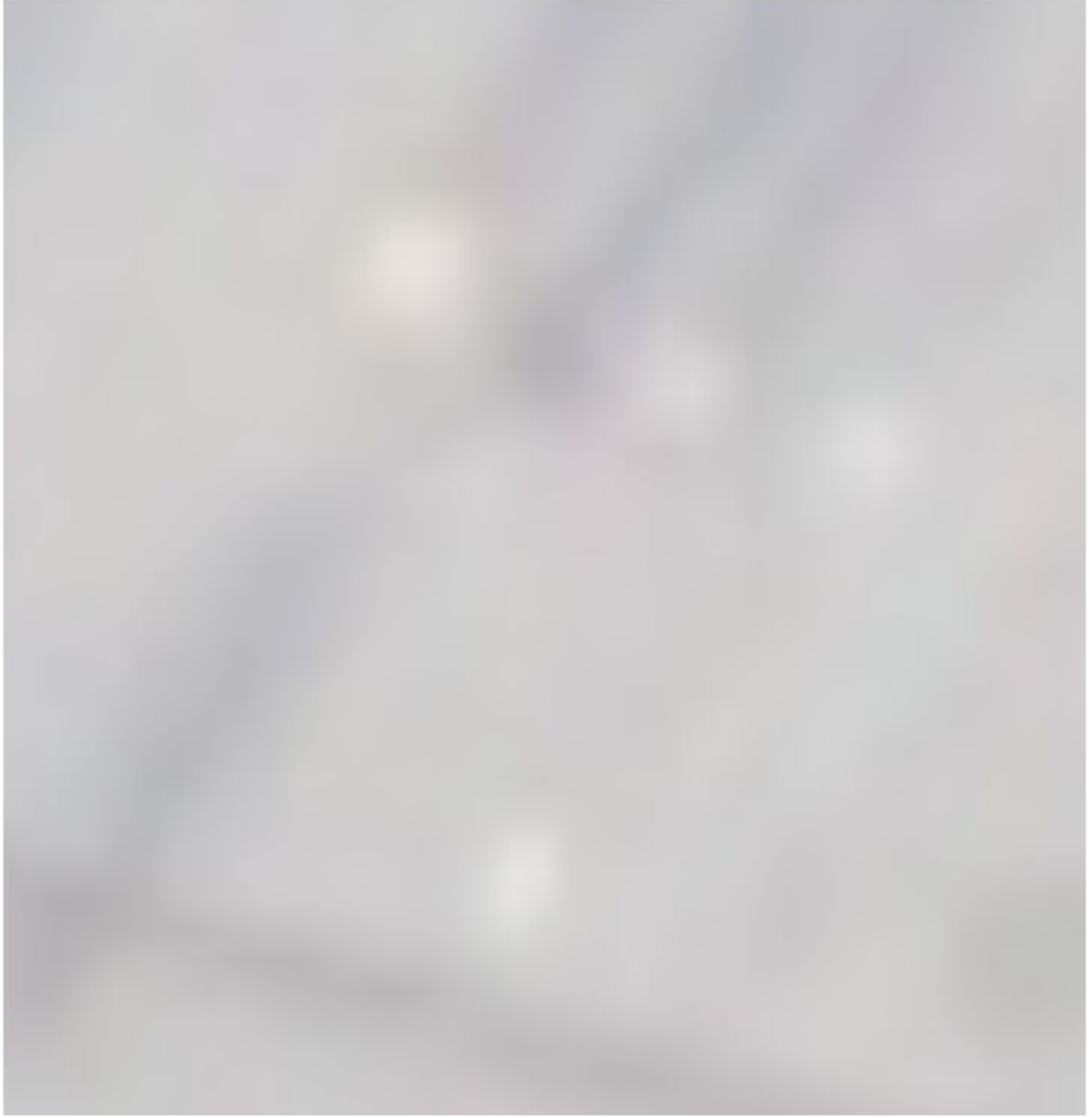}}
    \subfloat[]{\includegraphics[width=4.45cm, height=4.45cm]{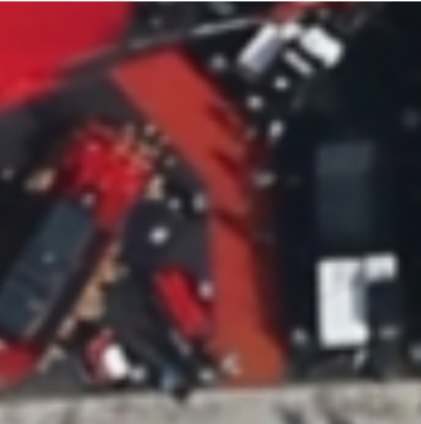}}
    \subfloat[]{\includegraphics[width=4.45cm, height=4.45cm]{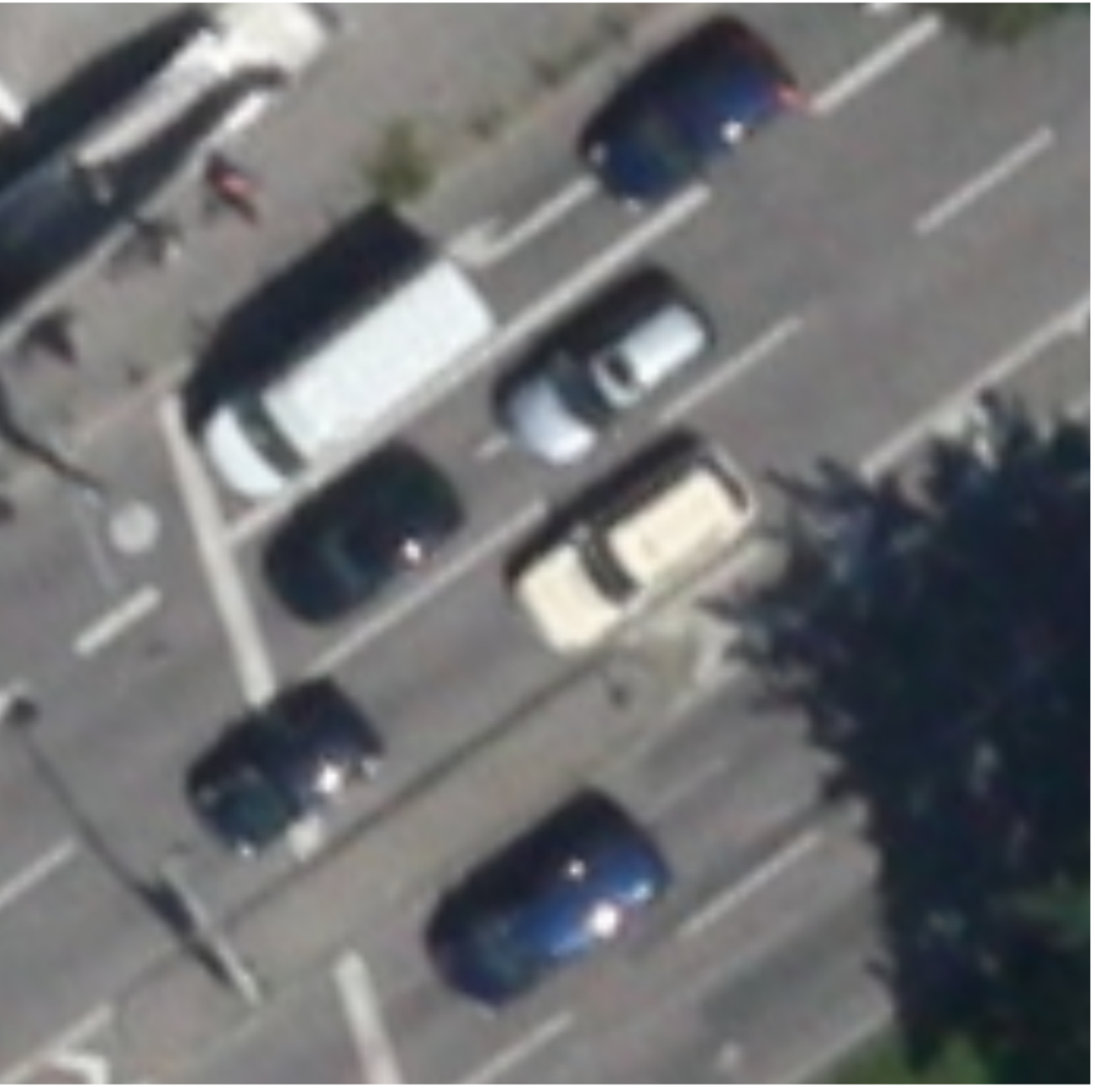}}
    \subfloat[]{\includegraphics[width=4.45cm, height=4.45cm]{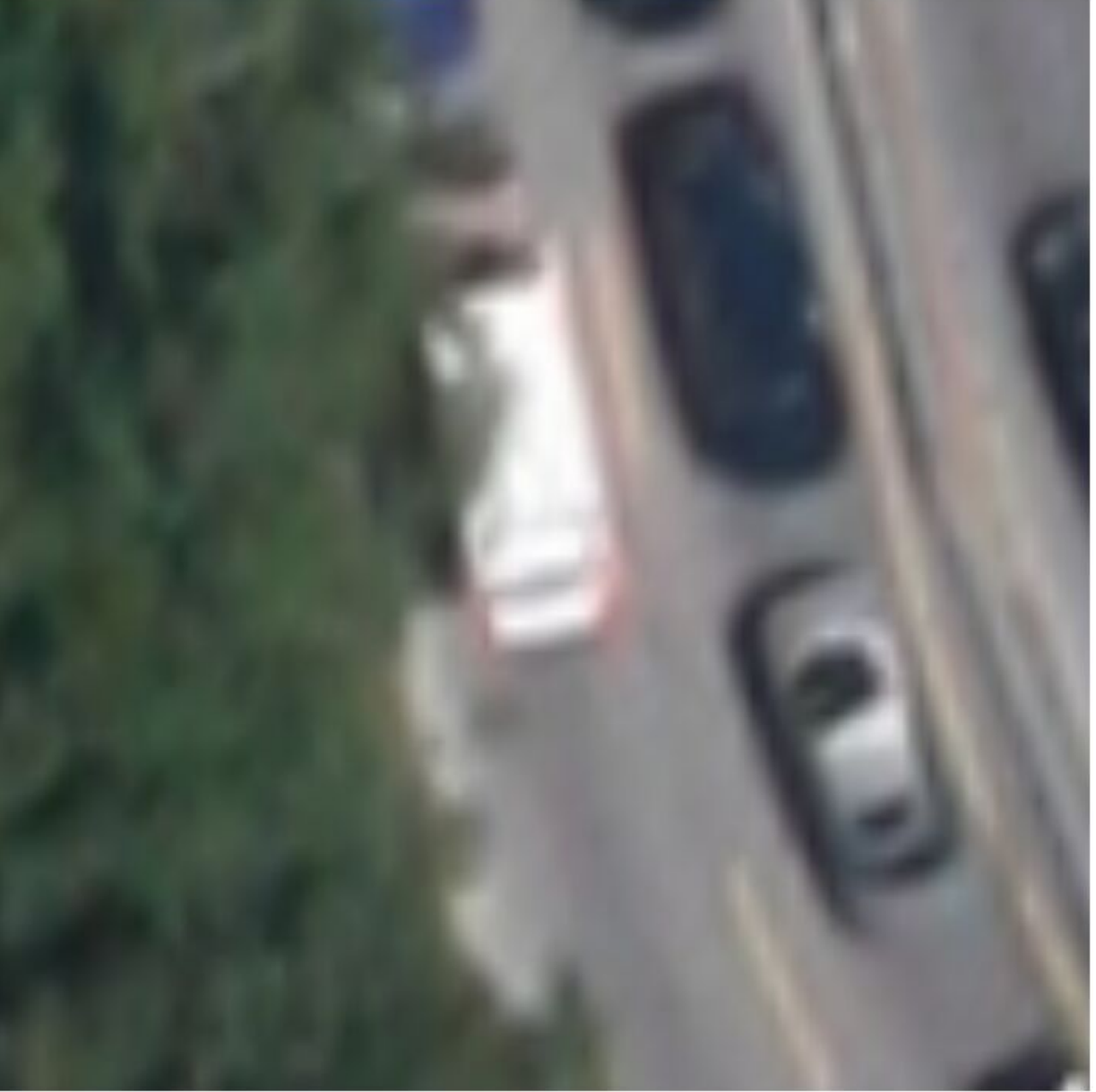}}
    \caption{Illustrations of some challenges in aerial \gls{mot} datasets. The examples are from the KIT~AIS pedestrian (a), AerialMPT (b), and KIT AIS vehicle datasets (c,d). Multiple pedestrians which are hard to distinguish due to their similar appearance features and low image contrast (a). Multiple pedestrians at a trade fair walking closely together with occlusions, shadows, and strong background colors (b). Multiple vehicles at a stop light where the shadow on the right hand side can be problematic (c). Multiple vehicles with some of them occluded by trees (d).}
    \label{fig:zoom}
\end{figure*}

We conduct our experiments on three aerial \gls{mot} datasets, namely \gls{aerialmpt} and KIT~AIS\footnote{https://www.ipf.kit.edu/code.php} pedestrian and vehicle datasets. All image sequences were captured by an airborne platform during different flight campaigns of the \gls{dlr}\footnote{https://www.dlr.de} and vary significantly in object density, movement patterns, and image size and quality. \autoref{fig:overview} shows sample images from the \gls{aerialmpt} dataset with the tracking results of our AerialMPTNet. 
The images were captured at different flight altitudes and their \gls{gsd} (reflecting the spatial size of a pixel) varies between 8~cm and 13~cm. 
The total number of objects per sequence ranges up to 609. Pedestrians in these datasets appear as small points, hardly exceeding an area of 4$\times$4 pixels. Even for human experts, distinguishing multiple pedestrians based on their appearance is laborious and challenging. Vehicles appear as bigger objects and are easier to distinguish based on their appearance features. However, different vehicle sizes, fast movements together with the low frame rates (e.g., 2 fps) and occlusions by bridges, trees, or other vehicles presents challenges to the vehicle tracking algorithm, illustrated in~\autoref{fig:zoom}.

AerialMPTNet is an end-to-end trainable regression-based neural network comprising a \gls{snn} module which takes two image patches as inputs, a target and a search patch, cropped from a previous and a current frame, respectively. The object location is known in the target patch and should be predicted for the search patch. In order to overcome the tracking challenges of the aerial \gls{mot} such as the objects with similar appearance features and densely moving together, AerialMPTNet incorporates temporal and graphical information in addition to the appearance information provided by the \gls{snn} module.
Our AerialMPTNet employs a \gls{lstm} for temporal information extraction and movement prediction, and a \gls{gcnn} for modeling the spatial and temporal relationships between adjacent objects (graphical information). 
AerialMPTNet outputs four values indicating the coordinates of the top-left and bottom-right corners of each object's bounding box in the search patch.
In this paper, we also investigate the influence of \gls{se} and \gls{ohem}~\cite{Shrivastava2016ohem} on the tracking performance of AerialMPTNet. To the best of our knowledge, we are the first work applying adaptive weighting of convolutional channels by \gls{se} and employ \gls{ohem} for the training of a DL-based tracking-by-regression method.

According to the results, our AerialMPTNet outperforms all previous methods for the pedestrian datasets and achieves competitive results for the vehicle dataset. Furthermore, \gls{lstm} and \gls{gcnn} modules adds value to the tracking performance. Moreover, while using \gls{se} and \gls{ohem} can significantly help in some scenarios, in other cases they may degrade the tracking results.

The rest of the paper is organized as follows. Section~\ref{sec:related_work} presents an overviews on related works; \autoref{sec:datasets} introduces the datasets used in our experiments; \autoref{sec:metrics} represents the metrics used for our quantitative evaluations; \autoref{sec:preExperiments} provides a comprehensive study on previous traditional and DL-based tracking methods on the aerial \gls{mot} datasets, with \autoref{sec:aerialMPTNet} explaining our AerialMPTNet with all its configurations; \autoref{sec:exp_setup} represents our experimental setups; \autoref{sec:evaluation} provides an extensive evaluation of our AerialMPTNet and compares it to the other methods; and \autoref{sec:conclusion} concludes our paper and gives ideas for future works. 

\section{Related Works} \label{sec:related_work}

This section introduces various categorizations of \gls{vot} as well as related previous works.

\subsection{Visual Object Tracking} \label{section:tracking}
Visual object tracking is defined as locating one or multiple objects in videos or image sequences over time. The traditional tracking process comprises four phases including initialization, appearance modeling, motion modeling, and object finding. During initialization, the targets are detected manually or by an object detector. In the appearance modeling step, visual features of the region of interest are extracted by various learning-based methods for detecting the target objects. The variety of scales, rotations, shifts, and occlusions makes this step challenging. 
Image features play a key role in the tracking algorithms. They can be mainly categorized into handcrafted and deep features. In recent years, research studies and applications have focused on developing and using deep features based on DNNs which have shown to be able to incorporate multi-level information and more robustness against appearance variations~\cite{fiaz2019handcrafted}. 
Nevertheless, DNNs require large enough training datasets which are not always available. Thus, for many applications, the handcrafted features are still preferable.
The motion modeling step aims at predicting the object movement in time and estimate the object locations in the next frames. This procedure effectively reduces the search space and consequently the computation cost. Widely used methods for motion modeling include Kalman filter~\cite{kalman1960new}, Sequential Monte Carlo methods~\cite{montecarlo2014} and RNNs. 
In the last step, object locations are found as the ones close to the estimated locations by the motion model.

\subsubsection{SOT and MOT}
Visual object tracking methods can be divided into \gls{sot}~\cite{wang_dcfnet_2017, ma2015hierarchical} and \gls{mot}~\cite{bahmanyar2019multiple, wojke_simple_2017} methods. While \gls{sot}s only track a single predetermined object throughout a video, even if there are multiple objects, \gls{mot}s can track multiple objects at the same time. Thus, \gls{mot}s can face exponential complexity and runtime increase based on the number of objects as compared to \gls{sot}s. 

\subsubsection{Detection-Based and Detection-Free}
Object tracking methods also can be categorized into detection-based~\cite{huang2008robust} and detection-free methods~\cite{lu_deep_nodate}. While the detection-based methods utilize object detectors to detect objects in each frame, the detection-free methods only need the initial object detection. Therefore, detection-free methods are usually faster than the detection-based ones; however, they are not able to detect new objects entering the scene and require manual initialization. 

\subsubsection{Online and Offline Learning}
Object tracking methods can be further divided based on their training strategies using either online or offline learning strategy. The methods with an online learning strategy can learn about the tracked objects during runtime. Thus, they can track generic objects~\cite{wang2016stct}. The methods with offline learning strategy are trained beforehand and are therefore faster during runtime~\cite{huang2017learning}.

\subsubsection{Online and Offline Tracking}
Tracking methods can be categorized into online and offline. Offline trackers take advantage of past and futures frames, while online ones can only infer from past frames. Although having all frames by offline tracking methods can increase the performance, in real-world scenarios future frames are not available. 

\subsubsection{One- and Two-Stage Tracking}
Most existing tracking approaches are based on a two-stage tracking-by-detection paradigm~\cite{chahyati2017tracking, zhang2017real}. In the first stage, a set of target samples is generated around the previously estimated position using region proposal, random sampling, or similar methods. In the second stage, each target sample is either classified as background or as the target object. In one-stage-tracking, however, the model receives a search sample together with a target sample as two inputs and directly predicts a response map or object coordinates by a previously trained regressor~\cite{held_learning_2016,bahmanyar2019multiple}.

\subsubsection{Traditional and DL-Based Trackers}
Traditional tracking methods mostly rely on the Kalman and particle filters to estimate object locations. They use velocity and location information to perform tracking~\cite{cuevas2005kalman, cuevas2007particle, okuma2004boosted}. Tracking methods only relying on such approaches have shown poor performance in unconstrained environments. Nevertheless, such filters can be advantageous in limiting the search space (decreasing the complexity and computational cost) by predicting and propagating object movements to the following frames.

A number of traditional tracking methods follow a tracking-by-detection paradigm based on template matching~\cite{brunelli2009template}. A given target patch models the appearance of the region of interest in the first frame. Matched regions are then found in the next frame using correlation, normalized cross-correlation, or the sum of squared distances methods~\cite{hager1996real, briechle2001template}. Scale, illumination, and rotation changes can cause difficulties with these methods.

More advanced tracking-by-detection-based methods rely on discriminative modeling, separating targets from their backgrounds within a specific search space. Various methods have been proposed for discriminative modeling, such as boosting methods and Support Vector Machines (SVMs)~\cite{avidan2007ensemble, hare2015struck}. A series of traditional tracking algorithms, such as MOSSE and KCF~\cite{bolme2010visual, henriques2014high}, utilizes correlation filters, which model the target's appearance by a set of filters trained on the images. In these methods, the target object is initially selected by cropping a small patch from the first frame centered at the object. For the tracking, the filters are convolved with a search window in the next frame. The output response map assumes to have a peak at the target's next location. As the correlation can be computed in the Fourier domain, such trackers achieve high frame rates. 

Recently, many research works and applications have focused on using  DL-based tracking methods. The great advantage of DL-based features over handcrafted ones such as HOG, raw pixels values or grey-scale templates have been presented previously for a variety of computer vision applications. These features are robust against appearance changes, occlusions, and dynamic environments. Examples of DL-based tracking methods include re-identification with appearance modeling and deep features \cite{wojke_simple_2017}, position regression mainly based on SNNs~\cite{ held_learning_2016, li2018high}, path prediction based on RNN-like networks~\cite{sadeghian2017tracking}, and object detection with DNNs such as YOLO~\cite{redmon2016you}.

\subsection{SOTs and MOTs}

In this section, we present a few \gls{sot} and \gls{mot} methods.

\subsubsection{\gls{sot} Methods} \label{subsec:sotm}
Kalal \etal proposed Median Flow~\cite{kalal2010forward}, which utilizes point and optical flow tracking. The inputs to the tracker are two consecutive images together with the initial bounding box of the target object. The tracker calculates a set of points from a rectangular grid within the bounding box. Each of these points is tracked by a Lucas-Kanade tracker generating a sparse motion flow. Afterwards, the framework evaluates the quality of the predictions and filters out the worst 50\%. The remaining point predictions are used to calculate the new bounding box positions considering the displacement. 

MOSSE~\cite{bolme2010visual}, KFC~\cite{henriques2014high} and CSRT~\cite{lukezic2017discriminative} are based upon \gls{dcf}s. Bolme~\etal\cite{bolme2010visual} proposed MOSSE which uses a new type of correlation filter called \gls{mosse}, which aims at producing stable filters when initialized using only one frame and grey-scale templates. MOSSE is trained with a set of training images \(f_i\) and training outputs \(g_i\), where \(g_i\) is generated from the ground truth as a 2D Gaussian centered on the target. This method can achieve state-of-the-art performances while running with high frame rates. Henriques~\etal\cite{henriques2014high} replaced the grey-scale templates with HOG features and proposed the idea of \gls{kcf}. \gls{kcf} works with multiple channel-like correlation filters. Additionally, the authors proposed using non-linear regression functions which are stronger than linear functions and provide non-linear filters that can be trained and evaluated as efficiently as linear correlation filters. Similar to \gls{kcf}, dual correlation filters use multiple channels. However, they are based on linear kernels to reduce the computational complexity while maintaining almost the same performance as the non-linear kernels. 
Recently, Lukezic~\etal~\cite{lukezic2017discriminative} proposed to use channel and reliability concepts to improve tracking based on \gls{dcf}s. In this method, the channel-wise reliability scores weight the influence of the learned filters based on their quality to improve the localization performance. Furthermore, a spatial reliability map concentrates the filters to the relevant part of the object for tacking. This makes it possible to widen the search space and improves the tracking performance for non-rectangular objects.

As we stated before, the choice of appearance features plays a crucial role in object tracking. Most previous DCF-based works utilize handcrafted features such as HOG, grey-scale features, raw pixels, and color names or the deep features trained independently for other tasks. Wang~\etal\cite{wang_dcfnet_2017} proposed an end-to-end trainable network architecture able to learn convolutional features and perform the correlation-based tracking simultaneously. The authors encode a \gls{dcf} as a correlation filter layer into the network, making it possible to backpropagate the weights through it. Since the calculations remain in the Fourier domain, the runtime complexity of the filter is not increased. The convolutional layers in front of the \gls{dcf} encode the prior tracking knowledge learned during an offline training process. The \gls{dcf} defines the network output as the probability heatmaps of object locations.

In the case of generic object tracking, the learning strategy is typically entirely online. However, online training of neural networks is slow due to backpropagation leading to a high run time complexity. However, Held~\etal\cite{held_learning_2016} developed a regression-based tracking method, called GOTURN, based on a \gls{snn}, which uses an offline training approach helping the network to learn the relationship between appearance and motion. This makes the tracking process significantly faster. 
This method utilizes the knowledge gained during the offline training to track new unknown objects online. The authors showed that without online backpropagation, GOTURN can track generic objects at 100 fps. The inputs to the network are two image patches cropped from the previous and current frames, centered at the known object position in the previous frame. The size of the patches depends on the object bounding box sizes and can be controlled by a hyperparameter. This determines the amount of contextual information given to the network. The network output is the coordinates of the object in the current image patch, which is then transformed to the image coordinates. GOTURN achieves state-of-the-art performance on common SOT benchmarks such as VOT~2014\footnote{https://www.votchallenge.net/vot2014/}.

\subsubsection{\gls{mot} Methods}

Bewley~\etal~\cite{bewley_simple_2016} proposed a simple multi-object tracking approach, called SORT, based on the Jaccard distance, the Kalman filter, and the Hungarian algorithm~\cite{kuhn1955hungarian}. Bounding box position and size are the only values used for motion estimation and assigning the objects to their new positions in the next frame. In the first step, objects are detected using Faster R-CNN~\cite{ren2015faster}. Subsequently, a linear constant velocity model approximates the movements of each object individually in consecutive frames. Afterwards, the algorithm compares the detected bounding boxes to the predicted ones based on~\gls{iou}, resulting in a distance matrix. The Hungarian algorithm then assigns each detected bounding box to a predicted (target) bounding box. Finally, the states of the assigned targets are updated using a Kalman filter. SORT runs with more than 250 \gls{fps} with almost state-of-the-art accuracy. Nevertheless, occlusion scenarios and re-identification issues are not considered for this method, which makes it inappropriate for long-term tracking. 

Wojke~\etal\cite{wojke_simple_2017} extended SORT to DeepSORT and tackled the occlusion and re-identification challenges, keeping the track handling and Kalman filtering modules almost unaltered. The main improvement takes place into the assignment process, in which two additional metrics are used: 1) motion information provided based on the Mahalanobis distance between the detected and predicted bounding boxes, 2) appearance information by calculating the cosine distance between the appearance features of a detected object and the already tracked object. The appearance features are computed by a deep neural network trained on a large person re-identification dataset~\cite{zheng2016mars}. A cascade strategy then determines object-to-track assignments. This strategy effectively encodes the probability spread in the association likelihood. DeepSORT performs poorly if the cascade strategy cannot match the detected and predicted bounding boxes.

Recently, Bergmann~\etal\cite{bergmann2019tracking} introduced Tracktor++ which is based on the Faster R-CNN object detection method. Faster R-CNN classifies region proposals to target and background and fits the selected bounding boxes to object contours by a regression head. The authors trained Faster R-CNN on the MOT17Det pedestrian dataset~\cite{milan2016mot16}. The first step is an object detection by Faster R-CNN. The detected objects in the first frame are then initialized as tracks. Afterwards, the tracks are tracked in the next frame by regressing their bounding boxes using the regression head. In this method, the lost or deactivated tracks can be re-identified in the following frames using a~\gls{snn} and a constant velocity motion model.

\subsection{Tracking in Satellite and Aerial Imagery} \label{subsec:trackingSat}

Visual object tracking for targets such as pedestrians and vehicles in satellite and aerial imagery is a challenging task that has been addressed by only few works, compared to the huge number addressing pedestrian and vehicle tracking in ground imagery~\cite{wang2015visual, yokoyama2005contour}.
Tracking in satellite and aerial imagery is much more complex. This is due to the moving cameras, large image sizes, different scales, large number of moving objects, tiny size of the objects (e.g., 4$\times$4 pixels for pedestrians, 30$\times$15 for vehicles), low frame rates, different visibility levels, and different atmospheric and weather conditions~\cite{jadhav_aerial_2019, milan2016mot16}.

\subsubsection{Tracking by Moving Object Detection}
Most of the previous works in satellite and aerial object tracking are based on moving object detection~\cite{reilly2010detection, meng2012object, benedek2009detection}.
Reilly~\etal\cite{reilly2010detection} proposed one of the earliest aerial object tracking approaches focusing on vehicle tracking mainly in highways. They compensate camera motion by a correction method based on point correspondence. A median background image is then modeled from ten frames and subtracted from the original frame for motion detection, resulting in the moving object positions. All images are split into overlapping grids, with each one defining an independent tracking problem. Objects are tracked using bipartite graph, matching a set of label nodes and a set of target nodes. The Hungarian algorithm solves the cost matrix afterwards to determine the assignments. The usage of the grids allows tracking large number of objects with the \(O(n^3)\) runtime complexity for the Hungarian algorithm. 

Meng~\etal\cite{meng2012object} followed the same direction. They addressed the tracking of ships and grounded aircrafts. Their method detects moving objects by calculating an \gls{adi} from frame to frame. Pixels with high values in the \gls{adi} are likely to be moving objects. Each target is afterwards modeled by extracting its spectral and spatial features, where spectral features refer to the target probability density functions and the spatial features to the target geometric areas. Given the target model, matching candidates are found in the following frames via regional feature matching using a sliding window paradigm.

Tracking methods based on moving object detection are not applicable for our pedestrian and vehicle tracking scenarios. For instance, Reilly~\etal\cite{reilly2010detection} use a road orientation estimate to constrain the assignment problem. Such estimations which may work for vehicles moving along predetermined paths (e.g., highways and streets), do not work for pedestrian tracking with much more diverse and complex movement behaviors (e.g., crowded situations and multiple crossings). In general, such methods perform poorly in unconstrained environments, are sensitive to illumination change and atmospheric conditions (e.g., clouds, shadows, or fog), suffer from the parallax effect, and cannot handle small or static objects. Additionally, since finding the moving objects requires considering multiple frames, these methods cannot be used for the real-time object tracking.

\subsubsection{Tracking by Appearance Features}

The methods based on appearance-like features overcome the issues of the tracking by moving object detection approaches~\cite{butenuth2011integrating, Schmidt2011, liu2015fast, qi2015unsupervised, bahmanyar2019multiple}, making it possible to detect small and static objects on single images. 
Butenuth~\etal\cite{butenuth2011integrating} deal with pedestrian tracking in aerial image sequences. They employ an iterative Bayesian tracking approach to track numerous pedestrians, where each pedestrian is described by its position, appearance features, and direction. A linear dynamic model then predicts futures states. Each link between a prediction and a detection is weighted by evaluating the state similarity and associated with the direct link method described in~\cite{huang2008robust}. 
Schmidt~\etal\cite{Schmidt2011} developed a tracking-by-detection framework based on Haar-like features. They use a Gentle AdaBoost classifier for object detection and an iterative Bayesian tracking approach, similar to~\cite{butenuth2011integrating}. Additionally, they calculate the optical flow between consecutive frames to extract motion information. However, due to the difficulties of detecting small objects in aerial imagery, the performance of the method is degraded by a large number of false positives and negatives.
%

Bahmanyar~\etal\cite{bahmanyar2019multiple} proposed \gls{smsot-cnn} and extended the GOTURN method, a SOT method developed by Held~\etal\cite{held_learning_2016}, by stacking the architecture of GOTURN to track multiple pedestrians and vehicles in aerial image sequences. SMSOT-CNN is the only previous DL-based work dealing with \gls{mot}. SMSOT-CNN expands the GOTURN network by three additional convolutional layers to improve the tracker's performance in locating the object in the search area. In their architecture, each SOT-CNN is responsible for tracking one object individually leading to a linear increase in the tracking complexity by the number of objects. They evaluate their approach on the vehicle and pedestrian sets of the KIT~AIS aerial image sequence dataset. Experimental results shows that SMSOT-CNN significantly outperforms GOTURN. Nevertheless, SMSOT-CNN performs poorly in crowded situations and when objects share similar appearance features.

In Section~\ref{sec:preExperiments}, we experimentally investigate a set of the reviewed visual object tracking methods on three aerial object tracking datasets.

\section{Datasets}\label{sec:datasets}

In this section, we introduce the datasets used in our experiments, namely the KIT~AIS (pedestrian and vehicle sets), the Aerial Multi-Pedestrian Tracking (AerialMPT)~\cite{kraus2020aerialmptnet}, and DLR's Aerial Crowd Dataset (DLR-ACD)~\cite{bahmanyar2019mrcnet}.
All these datasets are the first of their kind and aim at promoting pedestrian and vehicle detection and tracking based on aerial imagery.
The images of all these datasetes have been acquired by the German Aerospace Center~(DLR) using the 3K camera system, comprising a nadir-looking and two side-looking DSLR cameras, mounted on an airborne platform flying at different altitudes.
The different flight altitudes and camera configurations allow capturing images with multiple spatial resolutions (ground sampling distances - GSDs) and viewing angles.

For the tracking datasets, since the camera is continuously moving, in a post-processing step, all images were orthorectified with a digital elevation model, co-registered, and geo-referenced with a GPS/IMU system. Afterwards, images taken at the same time were fused into a single image and cropped to the region of interest. 
This process caused small errors visible in the frame alignments. Moreover, the frame rate of all sequences is 2~Hz. 
The image sequences were captured during different flight campaigns and differ significantly in object density, movement patterns, qualities, image sizes, viewing angles, and terrains. Furthermore, different sequences are composed by a varying number of frames ranging from 4 to 47. The number of frames per sequence depends on the image overlap in flight direction and the camera configuration. 

\subsection{KIT AIS}
The KIT~AIS dataset is generated for two tasks, vehicle and pedestrian tracking. The data have been annotated manually by human experts and suffer from a few human errors. Vehicles are annotated by the smallest enclosing rectangle (i.e., bounding box) oriented in the direction of their travel, while individual pedestrians are marked by point annotations on their heads. In our experiments, we used bounding boxes of sizes $4 \times 4$ and $5 \times 5$ pixels for the pedestrians according to the GSDs of the images, ranging from 12 to 17 cm. As objects may leave the scene or be occluded by other objects, the tracks are not labeled continuously for all cases. For the vehicle set cars, trucks, and buses are annotated if they lie entirely within the image region with more than \(\frac{2}{3}\) of their bodies visible. 
In the pedestrian set only pedestrians are labeled. Due to crowded scenarios or adverse atmospheric conditions in some frames, pedestrians can be hardly visible. In these cases, the tracks have been estimated by the annotators as precisely as possible. \autoref{tab:KITAISPED} and \autoref{tab:KITAISVEH} represent the statistics of the pedestrian and vehicle sets of the KIT~AIS dataset, respectively.  

The KIT~AIS pedestrian is composed of 13 sequences with 2,649 pedestrians (Pedest.), annotated by 32,760 annotation points (Anno.) throughout the frames \autoref{tab:KITAISPED}. The dataset is split into 7 training and 6 testing sequences with 104 and 85 frames (Fr.), respectively. The sequences are characterized by different lengths ranging from 4 to 31 frames.
The image sequences come from different flight campaigns over Allianz Arena (Munich, Germany), Rock am Ring concert (Nuremberg, Germany), and Karlsplatz (Munich, Germany).
\KITAISPedestrian

\begin{figure*}
    \centering
    \includegraphics[width=\textwidth]{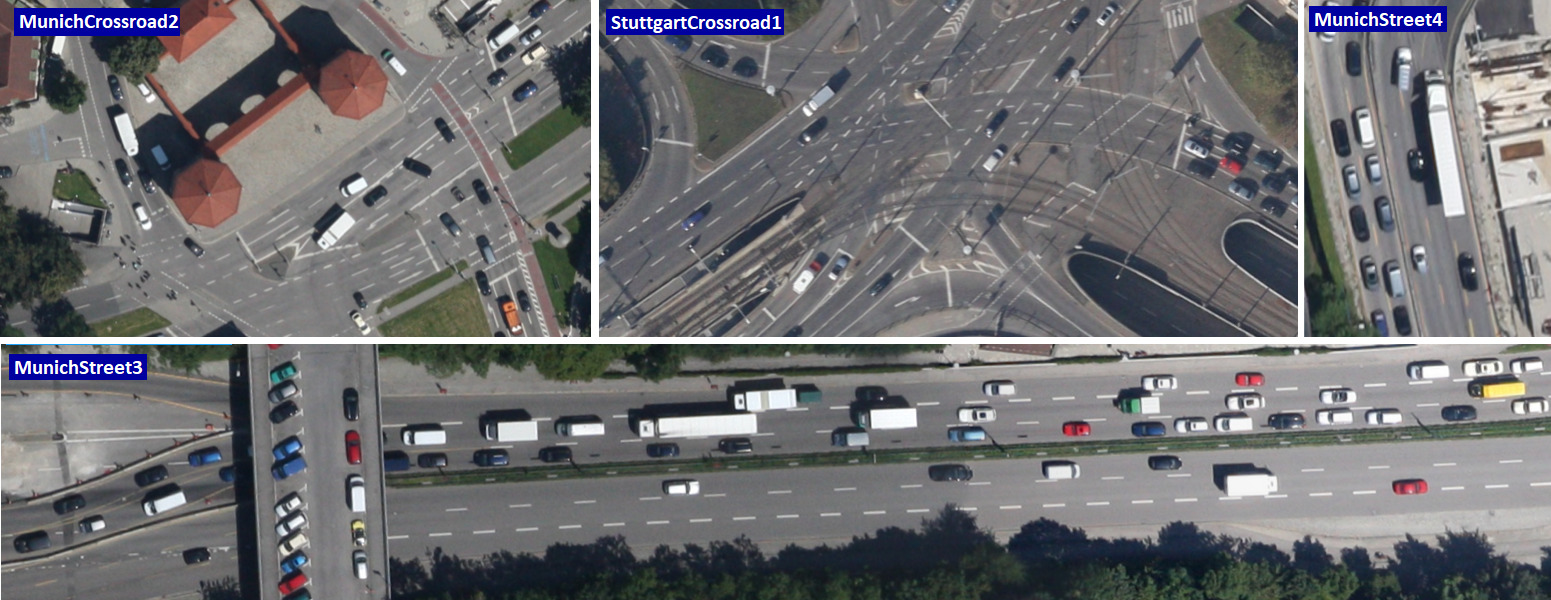}
    \caption{Sample images from the KIT~AIS vehicle dataset acquired at different locations in Munich and Stuttgart, Germany.}
    \label{fig:vehicleSamples}
\end{figure*}

KIT~AIS vehicle comprises 9 sequences with 464 vehicles annotated by 10,817 bounding boxes throughout 239 frames. It has no pre-defined train/test split. For our experiments, we split the dataset into 5 training and 4 testing sequences with 131 and 108 frames, respectively, similarly to~\cite{bahmanyar2019multiple}. According to~\autoref{tab:KITAISVEH}, the lengths of the sequences vary between 14 and 47 frames.
The image sequences have been acquired from a few highways, crossroads, and streets in Munich and Stuttgart, Germany. The dataset presents several tracking challenges such as lane change, overtaking, and turning maneuvers as well as partial and total occlusions by big objects (e.g., bridges). \autoref{fig:vehicleSamples} demonstrates sample images from the KIT~AIS vehicle dataset.
\KIAISVehicle

\subsection{AerialMPT}

\begin{figure*}
    \centering
    \includegraphics[width=\textwidth]{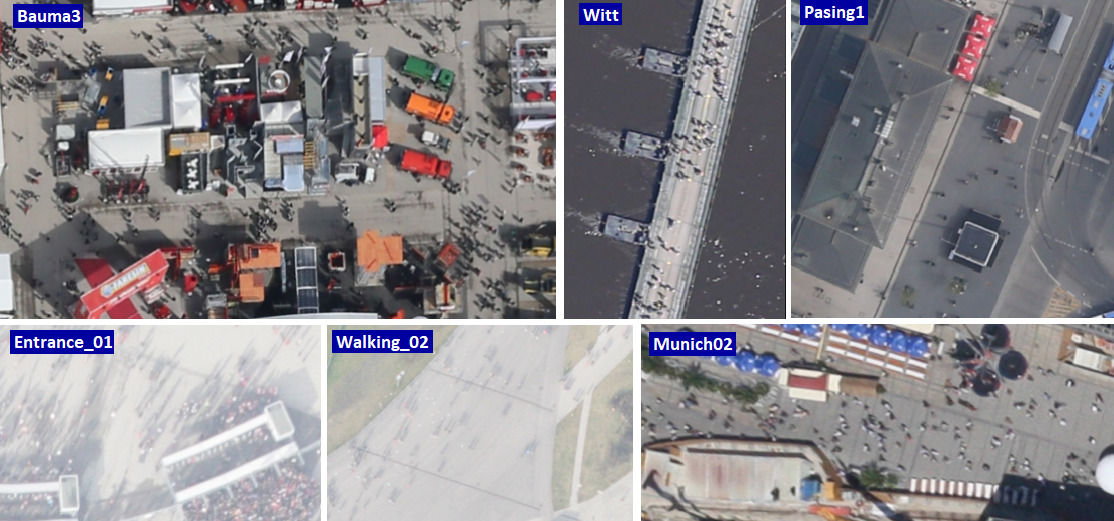}
    \caption{Sample images from the AerialMPT and KIT AIS datasets. ``Bauma3", ``Witt", ``Pasing1" are from AerialMPT. ``Entrance\_01", ``Walking\_02", and ``Munich02" are from KIT~AIS.}
    \label{fig:KITAErialMPTCompare}
\end{figure*}

The Aerial Multi-Pedestrian Tracking (AerialMPT) dataset~\cite{kraus2020aerialmptnet} is newly introduced to the community, and deals with the shortcomings of the KIT~AIS dataset such as the poor image quality and limited diversity. 
AerialMPT consists of 14 sequences with 2,528 pedestrians annotated by 44,740 annotation points throughout 307 frames ~\autoref{tab:MPTDataset}. 
Since the images have been acquired by a newer version of the DLR's 3K camera system, their quality and contrast are much better than the images of KIT~AIS dataset. \autoref{fig:KITAErialMPTCompare} compares a few sample images from the AerialMPT and KIT~AIS datasets.
\AerialMOTPedestrian

AerialMPT is split into 8 training and 6 testing sequences with 179 and 128 frames, respectively. The lengths of the sequences vary between 8 and 30 frames. The image sequences were selected from different crowd scenarios, e.g., from moving pedestrians on mass events and fairs to sparser crowds in the city centers.
\autoref{fig:overview} demonstrates an image from the AerialMPT dataset with the overlaid annotations.
%

\subsubsection{AerialMPT vs. KIT~AIS}
The AerialMPT has been generated in order to mitigate the limitations of the KIT~AIS pedestrian dataset. In addition to the higher quality of the images, the numbers of minimum annotations per frame and the total annotations of AerialMPT are significantly larger than those of the KIT~AIS dataset. All sequences in AerialMPT contain at least 50 pedestrian, while more than 20\% of the sequences of KIT~AIS include 
less than ten pedestrians.
Based on our visual inspection, not only the pedestrian movements in AerialMPT are more complex and realistic, but also the diversity of the crowd densities are greater than those of KIT~AIS.
The sequences in AerialMPT differ in weather conditions and visibility, incorporating more diverse kinds of shadows as compared to KIT~AIS. 
Furthermore, the sequences of AerialMPT are longer in average, with 60\% longer than 20 frames (less than 20\% in KIT~AIS).
Further details on these datasets can be found in~\cite{kraus2020aerialmptnet}.

\subsection{DLR-ACD}

\begin{figure*}
    \centering
    \subfloat[]{\includegraphics[width=.48\textwidth]{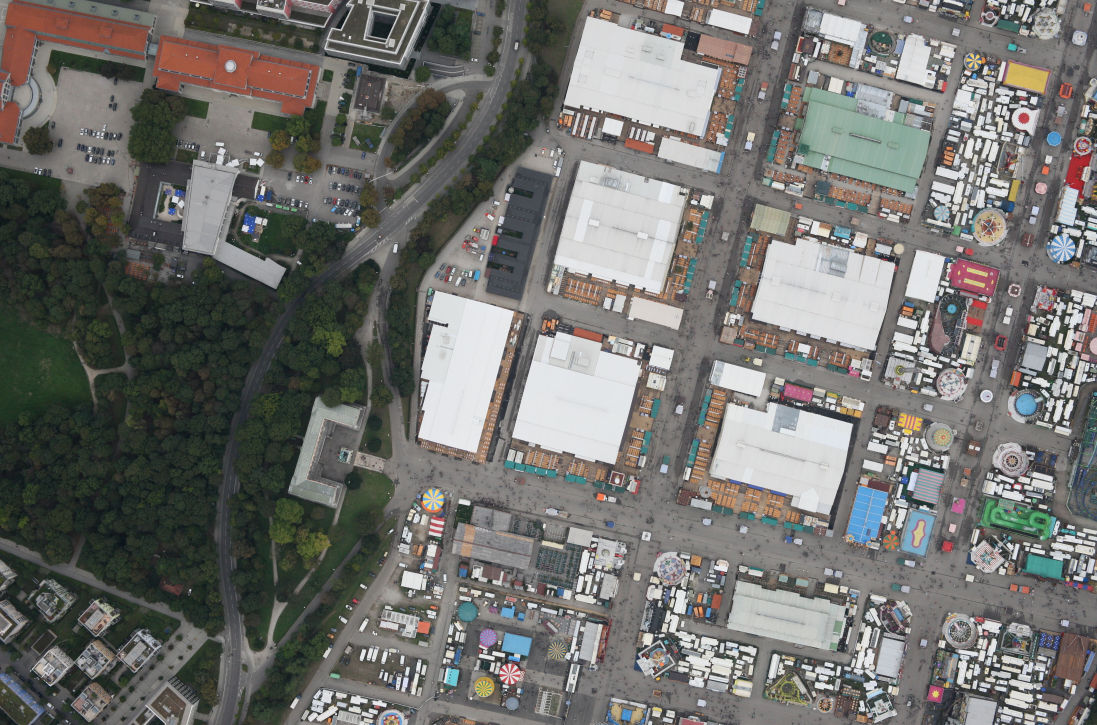}}
    \subfloat[]{\includegraphics[width=.48\textwidth]{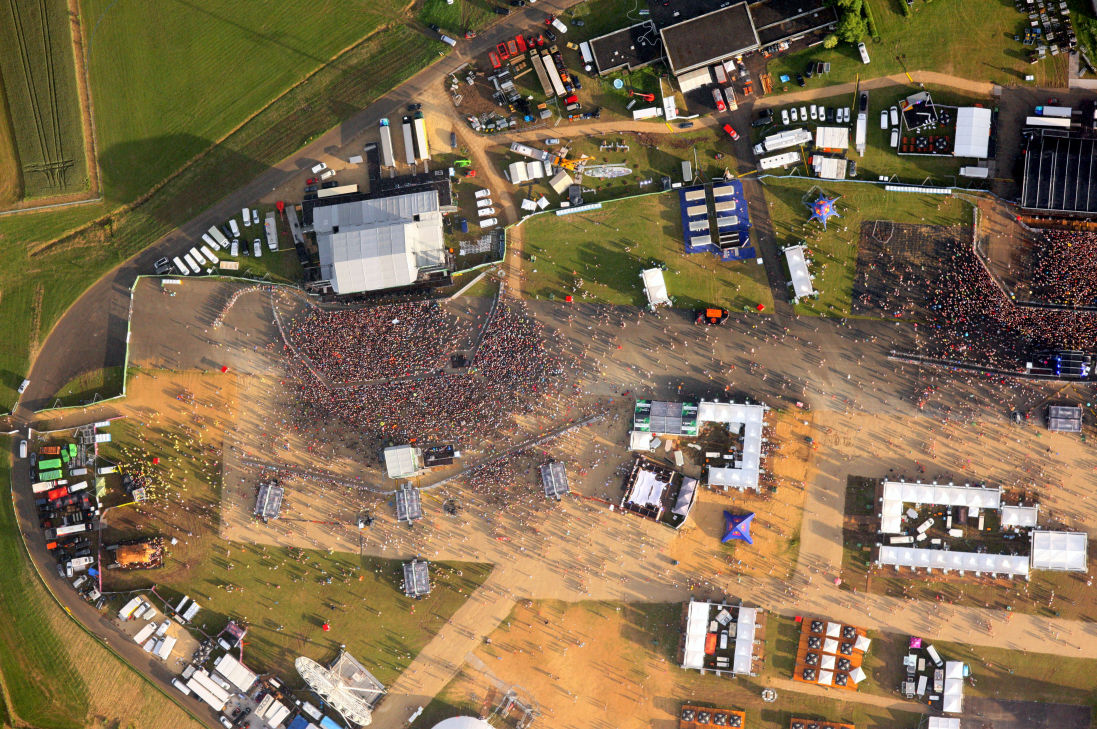}}
    \caption{Example images of the DLR-ACD dataset. The images are from an open-air (a) festival (b) and music concert.}
    \label{fig:dlr_acd_samples}
\end{figure*}

DLR-ACD is the first aerial crowd image dataset~\cite{bahmanyar2019mrcnet} comprises 33 large aerial RGB images with average size of $3619\times 5226$ pixels from different mass events and urban scenes containing crowds such as sports events, city centers, open-air fairs, and festivals. The GSDs of the images vary between 4.5 and 15 cm/pixel.
In DLR-ACD 226,291 pedestrians have been manually labeled by point annotations, with the number of pedestrians ranging from 285 to 24,368 per image. In addition to its unique viewing angle, the large number of pedestrians in most of the images ($>$2K) makes DLR-ACD stand out among the existing crowd datasets. Moreover, the crowd density can vary significantly within each image due to the large field of view of the images. \autoref{fig:dlr_acd_samples} demonstrates example images from the DLR-ACD dataset. For further details on this dataset, the interested reader is remanded to~\cite{bahmanyar2019mrcnet}.

\section{Evaluation Metrics}\label{sec:metrics}

In this section we introduce the most important metrics we use for our quantitative evaluations. We adopted widely-used metrics in the MOT domain based on~\cite{milan2016mot16} which are listed in~\autoref{tab:metrics}. In this table, $\uparrow$ and $\downarrow$ denote higher or lower values being better, respectively.  
The objective of MOT is finding the spatial positions of \(p\) objects as bounding boxes throughout an image sequence (object trajectories). Each bounding box is defined by the \(x\) and \(y\) coordinates of its top-left and bottom-right corners in each frame.
Tracking performances are evaluated based on true positives (TP), correctly predicting the object positions, false positives (FP), predicting the position of another object instead of the target object's position, and false negatives (FN), where an object position is totally missed. In our experiments, a prediction (tracklet) is considered as TP if the intersection over union (IoU) of the predicted and the corresponding ground truth bounding boxes is greater than \(0.5\). Moreover, an identity switch (IDS) occurs if an annotated object \(a\) is associated with a tracklet \(t\), and the assignment in the previous frame was \(a \neq t\). The fragmentation metric shows the total number of times a trajectory is interrupted during tracking.
\begin{table}
    \centering
    \caption{Description of the metrics used for quantitative evaluations.}
    \rowcolors{2}{gray!20}{white}
    \begin{tabular}{cc|c}
        Metric& & Description \\
        \hline
        IDF1 &$\uparrow$ & ID F1-Score\\
        IDP  &$\uparrow$ &  ID Global Min-Cost Precision\\
        IDR  &$\uparrow$ & ID Global Min-Cost Recall\\
        Rcll &$\uparrow$ & Recall\\
        Prcn &$\uparrow$ & Precision\\
        FAR  &$\downarrow$ & False Acceptance Rate\\
        MT   &$\uparrow$ & Ratio of Mostly Tracked Objects\\
        PT   &$\uparrow$ & Ratio of Partially Tracked Objects\\
        ML   &$\downarrow$ & Ratio of Mostly Lost Objects\\
        FP   &$\downarrow$ & False Positives \\
        FN   &$\downarrow$ & False Negatives\\
        IDS  &$\downarrow$ & Number of Identity Switches\\
        FM   &$\downarrow$ & Number of Fragmented Tracks\\
        MOTA &$\uparrow$ & Multiple Object Tracker Accuracy \\
        MOTP &$\uparrow$ & Multiple Object Tracker Precision\\
        MOTAL&$\uparrow$ & Multiple Object Tracker Accuracy Log\\
        \hline
    \end{tabular}
    \label{tab:metrics}
\end{table}

Among these metrics, the crucial ones are the multiple object tracker accuracy (MOTA) and the multiple object tracker precision (MOTP). 
MOTA represents the ability of trackers in following the trajectories throughout the frames \(t\), independently from the precision of the predictions: 
\begin{equation}
    MOTA = 1- \frac{\sum_t (FN_t + FP_t + ID_t)}{\sum_t GT_t}.
\end{equation}
The multiple object tracker accuracy log (MOTAL) is similar to MOTA; however, ID switches are considered on a logarithmic scale.
\begin{equation}
    MOTAL = 1 - \frac{\sum FN_T + FP_t + log_{10}(ID_t+1)}{\sum GT_t}.
\end{equation}
MOTP measures the performance of the trackers in precisely estimating object locations: 
\begin{equation}
    MOTP = \frac{\sum_{t,i}d_{t,i}}{\sum_t c_t},
\end{equation}
where \(d_{t,i}\) is the distance between a matched object \(i\) and the ground truth annotation in frame \(t\), and \(c\) is the total number of matched objects. 

Each tracklet can be considered as mostly tracked (MT), partially tracked (PT), or mostly lost (ML), based on how successful an object is tracked during its whole lifetime. A tracklet is mostly lost if it is only tracked less than 20\% of its lifetime and mostly tracked if it is tracked more than 80\% of its lifetime. Partially tracked applies to all remaining tracklets. We report MT, PT, and ML as percentages of the total amount of tracks. The false acceptance rate (FAR) for an image sequence with \(f\) frames describes the average amount of FPs per frame:
\begin{equation}
    FAR = \frac{\sum FP_t}{f}.
\end{equation}

In addition, we use recall and precision measures, defined as follows:
\begin{equation}
    Rcll = \frac{\sum TP_t}{\sum (TP_t + FN_t)},
\end{equation}
\begin{equation}
    Prcn = \frac{\sum TP_t}{\sum (TP_t + FP_t)}.
\end{equation}

Identification precision (IDP), identification recall (IDR), and IDF1 are similar to precision and recall; however, they take into account how long the tracker correctly identifies the targets. IDP and IDR are the ratios of computed and ground-truth detections that are correctly identified, respectively.
IDF1 is calculated as the ratio of correctly identified detections over the average number of computed and ground-truth detections. IDF1 allows ranking different trackers based on a single scalar value. For any further information on these metrics, the interested reader is remanded to to~\cite{ristani2016performance}.

\section{Preliminary Experiments}\label{sec:preExperiments}
This section empirically shows the existing challenges in aerial pedestrian tracking. 
We study the performance of a number of existing tracking methods including KCF~\cite{henriques2014high}, MOSSE~\cite{bolme2010visual}, CSRT~\cite{lukezic2017discriminative}, Median Flow~\cite{kalal2010forward}, SORT, DeepSORT~\cite{wojke_simple_2017}, Stacked-DCFNet~\cite{wang2017dcfnet}, Tracktor++~\cite{bergmann2019tracking}, SMSOT-CNN~\cite{bahmanyar2019multiple}, and Euclidean Online Tracking on aerial data, and show their strengths and limitations. Since in the early phase of our research, only the KIT AIS pedestrian dataset was available to us, the experiments of this section have been conducted on this dataset. However, our findings also hold for the AerialMPT dataset.

The tracking performance is usually correlated to the detection accuracy for both detection-free and detection-based methods. As our main focus is at tracking performance, in most of our experiments we assume perfect detection results and use the ground truth data. While for the object locations in the first frame are given to the detection-free methods, the detection-based methods are provided with the object locations in every frame. Therefore, for the detection-based methods, the most substantial measure is the number of ID switches, while for the other methods all metrics are considered in our evaluations.

\subsection{From Single- to Multi-Object Tracking}
Many tracking methods have been initially designed to track only single objects. However, according to~\cite{bahmanyar2019multiple}, most of them can be extended to handle MOT. Tracking management is an essential function in MOT which stores and exploits multiple active tracks at the same time, in order to remove and initialize the tracks of objects leaving from and entering into the scenes. For our experiments we developed a tracking management module for extending the SOT methods to MOT. It unites memory management, including the assignment of unique track IDs and individual object position storage, with track initialization, aging, and removing functionalities.

\subsubsection{KCF, MOSSE, CSRT, and Median Flow}
OpenCV provides several built-in object tracking algorithms. Among them, we investigate the KCF, MOSSE, CSRT, and Median Flow SOT methods. We extend them to the MOT scenarios within the OpenCV framework. We initialize the trackers by the ground truth bounding box positions. Moreover, we remove the objects if they leave the scene and their track ages are greater than 3 frames. 

\autoref{tab:emprExp} shows the tracking results of these methods on the KIT AIS dataset.
Results indicate the poor performance of all of these methods with a total MOTA scores varying between -85.8 and -55.9. The results of KCF and MOSSE are very similar. However, the use of HOG features and non-linear kernels in KCF improves MOTA by 0.9 and MOTP by 0.5 points respectively, compared to MOSSE. Moreover, both methods mostly track about 1~\% of the pedestrians in average.
ACSRT (which is also DCF-based) outperforms both prior methods significantly, reaching a total MOTA and MOTP of -55.9 and 78.4. It mostly tracks about 10~\% of the pedestrians in average and proves the effectiveness of the channel and reliability scores. According to the table, Median Flow achieves comparable results to CSRT with total MOTA and MOTP scores of -63.8 and 77.7, respectively.
Comparing the results on different sequences indicates that all algorithms perform significantly better on the \textit{RaR\_Snack\_Zone\_02} and \textit{RaR\_Snack\_Zone\_04} sequences. Based on visual inspection, we argue that this is due to their short length. Additionally, we argue that the overall low performance of these methods can be caused by the use of handcrafted features.

\preliminaryExperimentsSOT

\subsubsection{Stacked-DCFNet}
DCFNet \cite{wang2017dcfnet} is also a SOT  on a DCF. However, the DCF is implemented as part of a DNN and uses the featuresbased extracted by a light-weight CNN. Therefore, DCFNet is a perfect choice to study whether deep features improve the tracking performance compared to the handcrafted ones. 
For our experiments, we took the \textit{PyTorch} implementation\footnote{https://github.com/foolwood/DCFNet\_pytorch} of DCFNet and modified its network structure to handle multi-object tracking, and we refer to it as ``Stacked-DCFNet".
From the KIT AIS pedestrian training set we crop a total of 20,666 image patches centered at every pedestrian. The patch size is the bounding box size multiplied by 10 in order to consider contextual information to some degree. Then we scale the patches to 125$\times$125 pixels to match the network input size. Using the patches, we retrain the convolutional layers of the network for 50 epochs with ADAM \cite{kingma2014adam} optimizer, MSE loss, initial learning rate of 0.01, and a batch size of 64. Moreover, we set the spatial bandwidth to 0.1 for both online tracking and offline training. Furthermore, in order to adapt it to MOT, we use our developed \textit{Python} module, and remove the objects if they leave the scene while their track ages are greater than 3 frames. Multiple targets are given to the network within one batch. For each target object, the network receives two image patches, from previous and current frames, centered on the known previous position of the object. The network output is the probability heatmap in which the highest value represents the most likely object location in the image patch of the current frame (search patch). If this value is below a certain threshold, we consider the object as lost.
Furthermore, we propose a simple linear motion model and set the center point of the search patch to the position estimate of this model instead of the position of the object in the previous frame patch (as in the original work). Based on the latest movement \(v_t(x,y)\) of a target, we estimate its position as:
\begin{equation}
    p_{est}(x,y) = p(x,y) + k \cdot v_t(x,y),
\end{equation}
where \(k\) determines the influence of the last movement.

The tracking results in~\autoref{tab:emprExp} demonstrate that Stacked-DCFNet significantly outperforms the method with handcrafted features by a MOTA score of -37.3 (18.6 points higher than that of the CSRT). The MT and ML rates are also improving with only losing 23.6~\% of all tracks while mostly tracking the 13.8~\% of the pedestrians. According to the results, Stacked-DCFNet performs better on the longer sequences (\textit{AA\_Crossing\_02}, \textit{AA\_Walking\_02} and \textit{Munich02}), which shows the ability of the method in tracking objects for a longer time.
Altogether, the results indicate that deep features outperform the handcrafted ones by a large margin.

\subsection{Multi-Object Trackers}
In this section, we study a number of MOT methods including SORT, DeepSORT, and Tracktor++. Additionally, we propose a new tracking algorithm called Euclidean Online Tracking (EOT) which uses the Euclidean distance for object matching.

\subsubsection{DeepSORT and SORT}
DeepSORT \cite{wojke_simple_2017} is a MOT method comprising deep features and an IoU-based tracking strategy. For our experiments, we use the \textit{PyTorch} implementation\footnote{https://github.com/ZQPei/deep\_sort\_pytorch} of DeepSORT and adapt it for the KIT AIS dataset by changing the bounding box size and IoU threshold, as well as fine-tuning the network on the training set of the KIT AIS dataset. As mentioned, for the object locations we use the ground truth and do not use the DeepSORT's object detector. \autoref{tab:emprExpMOT} shows the tracking results of our experiments in which Rcll, Prcn, FAR, MT, PT, ML, FN, MOTP, and FM are not important in our evaluations as the ground truth is used instead of the detection results.
\preliminaryExperimentsMOT

In the first experiment, we employ DeepSORT with its original parameter settings. As the results show, this configuration is not suitable for tracking small objects (pedestrians) in aerial imagery. DeepSORT utilizes deep appearance features to associate objects to tracklets; however, for the first few frames, it relies on IoU metric until enough appearance features are available. The original IoU threshold is \(0.5\). The standard DeepSORT uses a Kalman filter for each object to estimate its position in the next frame. However, due to small IoU overlaps between most predictions and detections, many tracks can not be associated with any detection, making it impossible to use the deep features afterwards. The main cause of minor overlaps is the small size of the bounding boxes. For example, if the Kalman filter estimates the object position only 2 pixels off the detection's position, for a bounding box of 4$\times$4 pixels, the overlap would be below the threshold and, consequently, the tracklet and the object cannot be matched. These mismatches result in a large number of falsely initiated new tracks, leading to a total amount of 8,627 ID switches, an average amount of 8.27 ID switches per person, and an average amount of 0.71 ID switches per detection.

\begin{figure}
  \centering
  \includegraphics[width=\columnwidth]{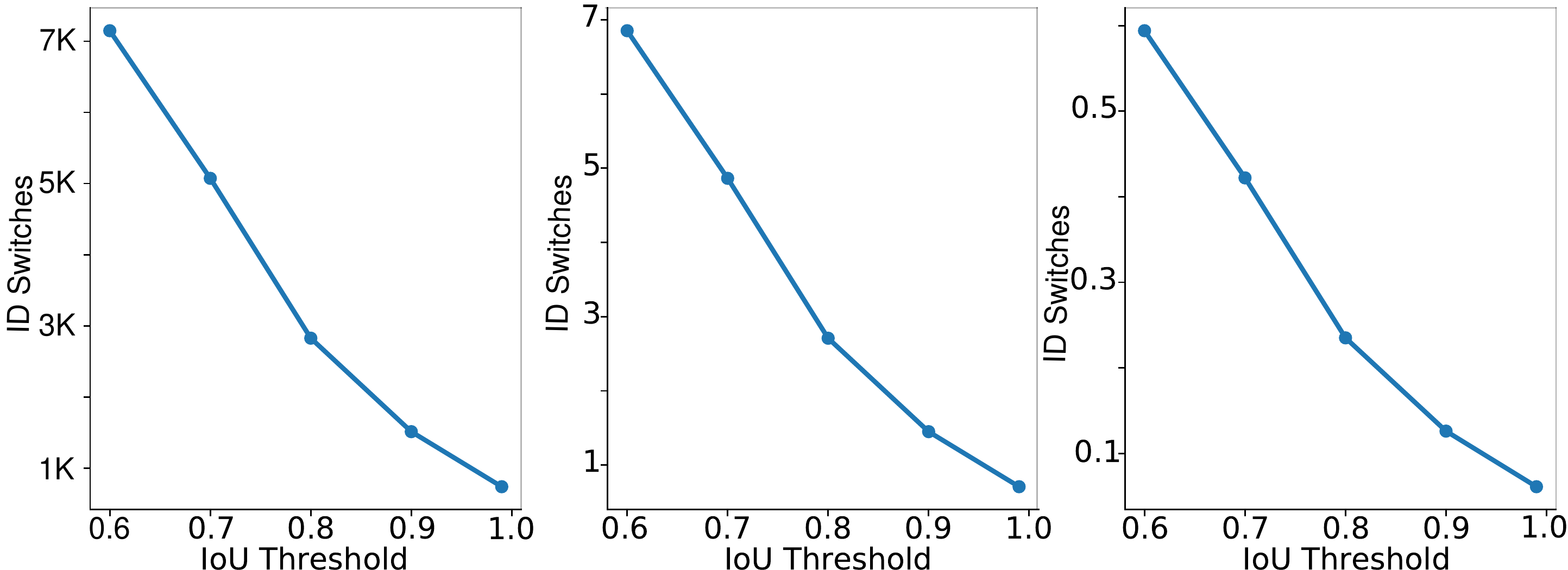}%
  \caption{ID Switches versus IoU thresholds in DeepSORT. From left to right: total, average per person, and average per detection ID Switches.} 
  \label{fig:id_switches}
\end{figure}

We tackle this problem by enlarging the bounding boxes by a factor of two in order to increase the IoU overlaps, increase the number of matched tracklets and detections, and enable the use of appearance features. According to~\autoref{tab:emprExpMOT}, it results in a 41.19\% decrease in the total number of ID switches (from 8,627 to 5,073), a 56.38\% decrease in the average number of ID switches per person (from 8.62 to 4.86), and a 59.15\% decrease in the average number of ID switches per detection (from 0.71 to 0.42). 
We further analyze the impact of using different IoU thresholds on the tracking performance. \autoref{fig:id_switches} illustrates the number of ID switches with different IoU thresholds. It can be observed that by increasing the threshold (minimizing the required overlap for object matching) the number of ID switches reduces. The least number of ID switches (738 switches) is achieved by the IoU threshold of 0.99. More details can be seen in~\autoref{tab:emprExpMOT}. Based on the results, enlarging the bounding boxes and changing the IoU threshold significantly improves the tracking results of DeepSORT as compared to its original settings (ID switches by 91.44\% and MOTA by 3.7 times). This confirms that the missing IoU overlap is the main issue with the standard DeepSORT.
After adapting the IoU object matching, the deep appearance features play a prominent role in the object tracking after the first few frames. Thus, a fine-tuning of the DeepSORT's neural network on the training set of the KIT AIS pedestrian dataset can further improve the results. Originally, the network has been trained on a large person re-identification dataset, which is very different from our scenario, especially in the looking angle and the object sizes, as the bounding boxes in aerial images are much smaller than in the person re-identification dataset (\(4 \times 4\) vs. \(128 \times 64\) pixels). Scaling the bounding boxes of our aerial dataset to fit the network input size leads to relevant interpolation errors. For our experiments we initialize the last re-identification layers from scratch, and the rest of the network using the pre-trained weights and biases. We also changed the number of classes to 610, representing the number of different pedestrians after cropping the images into the patches with the size of the bounding boxes, and ignoring the patches located at the image border. Instead of scaling the patches to \(128 \times 64\) pixels, we only scale them to \(50 \times 50\). We trained the classifier for 20 epochs with SGD optimizer, Cross-Entropy loss function, batch size of 128, and an initial learning rate of \(0.01\). Moreover, we doubled the bounding box sizes for our experiment.

The results in~\autoref{tab:emprExpMOT} show that the total number of ID switches only decreases from 738 to 734. This indicates that the deep appearance features of DeepSORT are not useful for our problem. While for a large object a small deviation of the bounding box position is tolerable (as the bounding box still mostly contains object-relevant areas), for our very small objects this can cause significant changes in object relevance. The extracted features mostly contain background information. Consequently, in the appearance matching step, the object features from its previous and currently estimated positions can differ significantly. Additionally, the appearance features of different pedestrians in aerial images are often not discriminative enough to distinguish them.

In order to better demonstrate this effect, we evaluate DeepSORT without any appearance feature, also known as SORT. \autoref{tab:emprExpMOT} shows the tracking results with original and doubled bounding box sizes and an IoU threshold of 0.99. According to the results, SORT outperforms the fine-tuned DeepSORT with 438 ID switches. Nevertheless, the number of ID switches is still high, given that we use the ground truth object positions. This could be due to the low frame rate of the dataset and the small sizes of the the objects. Although enlarging the bounding boxes improved the performance significantly, it leads to a poor localization accuracy.

\subsubsection{Tracktor++}\label{subsec:tracktor}

Tracktor++~\cite{bergmann2019tracking} is an MOT method based on deep features. It employs a Faster-RCNN to perform object detection and tracking through regression. We use its \textit{PyTorch} implementation\footnote{https://github.com/phil-bergmann/tracking\_wo\_bnw} and adapt it to our aerial dataset.
We tested Tracktor++ with the ground truth object positions instead of using its detection module; however, it totally failed the tracking task with these settings. Faster-RCNN has been trained on the datasets which are very different to our aerial dataset, for example in looking angle, number and size of the objects. Therefore, we fine-tune Faster-RCNN on the KIT AIS dataset. To this end, we had to adjust the training procedure to the specification of our dataset. 

We use Faster-RCNN with a ResNet50 backbone, pre-trained on the ImageNet dataset. We change the anchor sizes to \{2, 3, 4, 5, 6\} and the aspect ratios to \{0.7, 1.0, 1.3\}, enabling it to detect small objects. Additionally, we increase the maximum detections per image to 300, set the minimum size of an image to be rescaled to 400 pixels, the region proposal non-maximum suppression (NMS) threshold to 0.3, and the box predictor NMS threshold to 0.1. The NMS thresholds influence the amount of overlap for region proposals and box predictions. Instead of SGD, we use an ADAM optimizer with an initial learning rate of 0.0001 and a weight decay of 0.0005. Moreover, we decrease the learning rate every 40 epochs by a factor of 10 and set the number of classes to 2, corresponding to background and pedestrians. We also apply substantial online data augmentation including random flipping of every second image horizontally and vertically, color jitter, and random scaling in a range of 10\%.

The tracking results of Tracktor++ with the fine-tuned Faster-RCNN are presented in~\autoref{tab:emprExpMOT}. The detection precision and recall of Faster-RCNN are 25~\% and 31~\%, respectively, with this poor detection performance potentially propagated to the tracking part. According to the table, Tracktor++ only achieves an overall MOTA of 5.3 and 2,188 ID switches even when we use ground truth object positions.
We conclude by assuming that Tracktor++ has difficulties with the low frame rate of the dataset and the small object sizes.

\subsubsection{SMSOT-CNN}

SMSOT-CNN~\cite{bahmanyar2019multiple} is the first DL-based method for multi-object tracking in aerial imagery. It is an extension to GOTURN~\cite{held2016learning}, an SOT regression-based method using CNNs to track generic objects at high speed. SMSOT-CNN adapts GOTURN for MOT scenarios by three additional convolution layers and a tacking management module. The network receives two image patches from the previous and current frames, where both are centered at the object position in the previous frame. The size of the image patches (the amount of contextual information) is adjusted by a hyperparameter. The network regresses the object position in the coordinates of the current frame's image patch. SMSOT-CNN has been evaluated on the KIT~AIS pedestrian dataset in~\cite{bahmanyar2019multiple}, where the objects' first positions are given based on the ground truth data. The tracking results can be seen in~\autoref{tab:emprExpMOT}. Due to the use of a deep network and the local search for the next position of the objects, the number of ID switches by SMSOT-CNN is 157, which is small, relative to the other methods. Moreover, this algorithm achieves an overall MOTA and MOTP  of -29.8 and 71.0, respectively. Based on our visual inspections, SMSOT-CNN has some difficulties in densely crowded situations where the objects share similar appearance features. In these cases, multiple similarly looking objects can be present in an image patch, resulting in ID switches and losing track of the target objects.

\subsubsection{Euclidean Online Tracking}

Inspired by the tracking results of SORT besides its simplicity, we propose EOT based on the architecture of SORT. EOT uses a Kalman filter similarly to SORT. Then it calculates the euclidean distance between all predictions ($x_i,y_i$) and detections ($x_j,y_j$), and normalizes them w.r.t. the GSD of the frame to construct a cost matrix as follows: 
\begin{equation}
    D_{i,j} = GSD \cdot \sqrt{(x_i-x_j)^2 + (y_i-y_j)^2}. 
    \label{eq:formel}
\end{equation}
After that, as in SORT, we use the Hungarian algorithm to look for global minima. However, if objects enter or leave the scene, the Hungarian algorithm can propagate an error to the whole prediction-detection matching procedure: therefore, we constrain the cost matrix so that all distances greater than a certain threshold are ignored and set to an infinity cost. We set the threshold to $17\cdot GSD$ empirically. Furthermore, only objects successfully tracked in the previous frame are considered for the matching process.
According to~\autoref{tab:emprExpMOT}, while the total MOTA score is competitive with the previously studied methods, EOT achieves the least ID switches (only 37). Compared to SORT, as EOT keeps better track of the objects, the deviations in the Kalman filter predictions are smaller. Therefore, Euclidean distance is a better option as compared to IoU for our aerial image sequences.

\subsection{Conclusion of the Experiments}

\begin{figure*}%
    \centering
    \subfloat{\begin{overpic}[width=2.9cm]{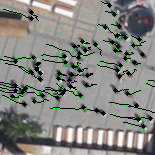}
         \put(2,70){\color{red}25}
       \end{overpic}}    
    \subfloat{\begin{overpic}[width=2.9cm]{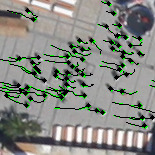}
         \put(2,70){\color{red}26}
       \end{overpic}}
    \subfloat{\begin{overpic}[width=2.9cm]{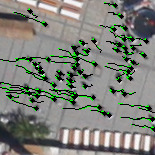}
         \put(2,70){\color{red}27}
       \end{overpic}}    
    \subfloat{\begin{overpic}[width=2.9cm]{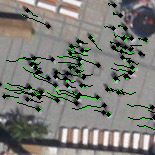}
         \put(2,70){\color{red}28}
       \end{overpic}}
    \subfloat{\begin{overpic}[width=2.9cm]{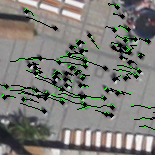}
         \put(2,70){\color{red}29}
       \end{overpic}}    
    \subfloat{\begin{overpic}[width=2.9cm]{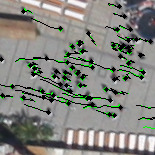}
         \put(2,70){\color{red}30}
       \end{overpic}} 
    \caption{A success case processed by Stacked-DCFNet on the sequence ``Munich02". The tracking results and ground truth are depicted in green and black, respectively.}%
    \label{fig:successfullTrackingEOT}%
\end{figure*}

In this section, we conclude our preliminary study. According to the results, our EOT is the best performing tracking method.
\autoref{fig:successfullTrackingEOT} illustrates a major case of success by our EOT method. We can observe that almost all pedestrians are tracked successfully, even though the sequence is crowded and people walk in different directions. Furthermore, the significant cases of false positives and negatives are caused by the limitation of the evaluation approach. In other words, while EOT tracks most of the objects, since the evaluation approach is constrained to the minimum 50\% overlap (4 pixels), the correctly tracked objects with smaller overlaps are not considered.

\begin{figure*}%
    \centering
    \subfloat{\begin{overpic}[width=2.9cm]{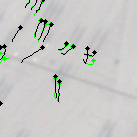}
         \put(2,5){54}
       \end{overpic}}
    \subfloat{\begin{overpic}[width=2.9cm]{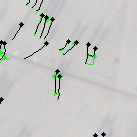}
         \put(2,5){55}
       \end{overpic}}
    \subfloat{\begin{overpic}[width=2.9cm]{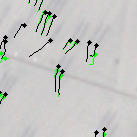}
         \put(2,5){56}
       \end{overpic}}    
    \subfloat{\begin{overpic}[width=2.9cm]{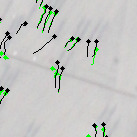}
         \put(2,5){57}
       \end{overpic}}
    \subfloat{\begin{overpic}[width=2.9cm]{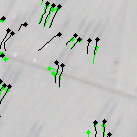}
         \put(2,5){58}
       \end{overpic}}    
    \subfloat{\begin{overpic}[width=2.9cm]{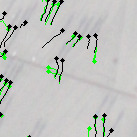}
         \put(2,5){59}
       \end{overpic}}
    \caption{A failure case by Stacked-DCFNet on the sequence ``AA\_Walking\_02". The tracking results and ground truth are depicted in green and black, respectively.}%
    \label{fig:failureCaseDCFNet1}%
\end{figure*}

\begin{figure*}%
    \centering
    \subfloat{\begin{overpic}[width=2.9cm]{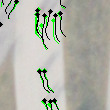}
         \put(2,5){180}
       \end{overpic}}    
    \subfloat{\begin{overpic}[width=2.9cm]{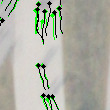}
         \put(2,5){181}
       \end{overpic}}
    \subfloat{\begin{overpic}[width=2.9cm]{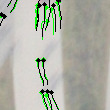}
         \put(2,5){182}
       \end{overpic}}    
    \subfloat{\begin{overpic}[width=2.9cm]{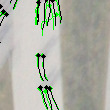}
         \put(2,5){183}
       \end{overpic}}
    \subfloat{\begin{overpic}[width=2.9cm]{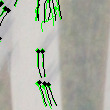}
         \put(2,5){184}
       \end{overpic}}    
    \subfloat{\begin{overpic}[width=2.9cm]{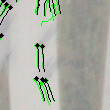}
         \put(2,5){185}
       \end{overpic}}
    \caption{A success case by Stacked-DCFNet on the sequence ``AA\_Crossing\_02". The tracking results and ground truth are depicted in green and black, respectively.}

    \label{fig:successfullTrackingDCFNet}%
\end{figure*}

\begin{figure}%
    \centering
    \subfloat{\begin{overpic}[width=.24\columnwidth]{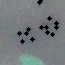}
         \put(2,80){141}
       \end{overpic}}    
    \subfloat{\begin{overpic}[width=.24\columnwidth]{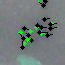}
         \put(2,80){142}
       \end{overpic}}
    \subfloat{\begin{overpic}[width=.24\columnwidth]{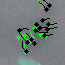}
         \put(2,80){143}
       \end{overpic}}    
    \subfloat{\begin{overpic}[width=.24\columnwidth]{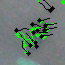}
         \put(2,80){144}
       \end{overpic}}
    \caption{A failure case by Stacked-DCFNet on the test sequence ``RaR\_Snack\_Zone\_04". The tracking results and the ground truth are depicted in green and black, respectively.}%
    \label{fig:failureCaseDCFNet2}%
\end{figure}

\begin{figure}%
    \centering
    \subfloat[]{\includegraphics[width=.48\columnwidth]{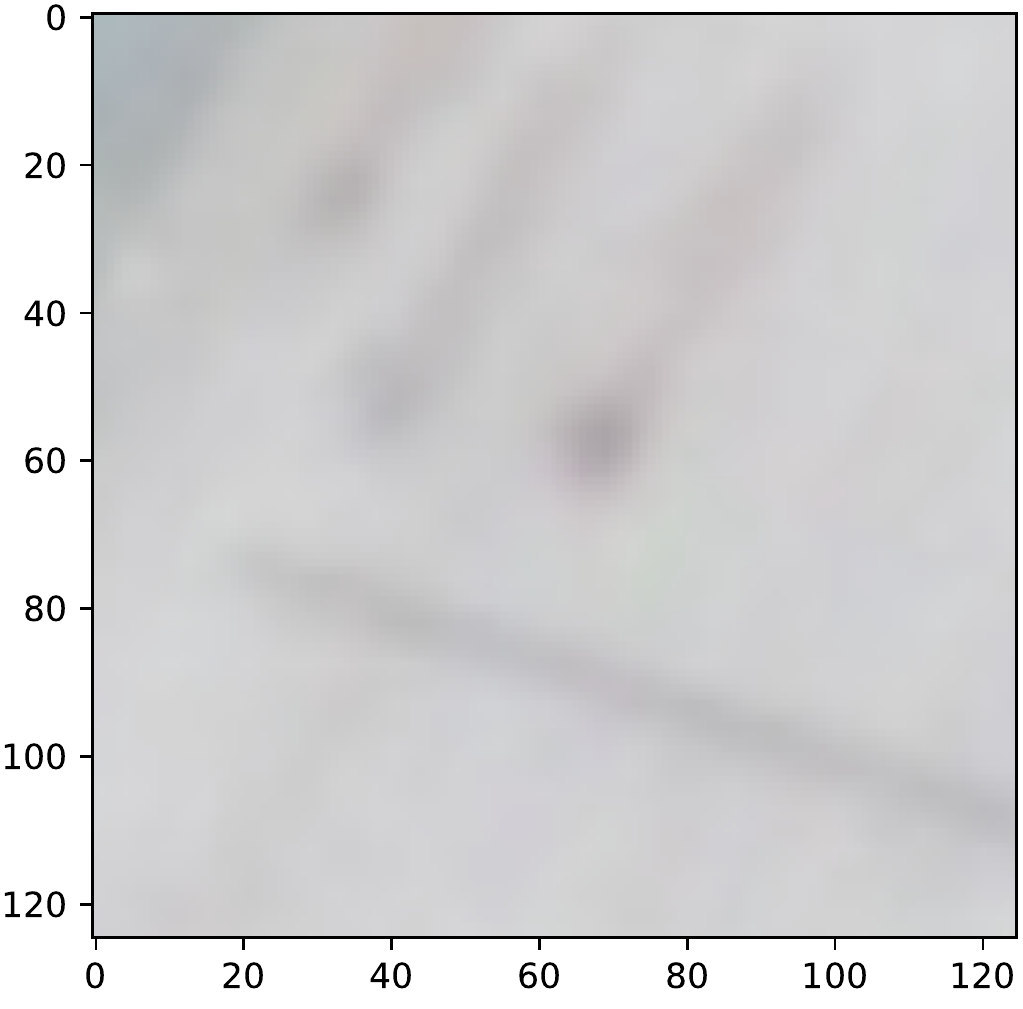} }%
    \subfloat[]{\includegraphics[width=.48\columnwidth]{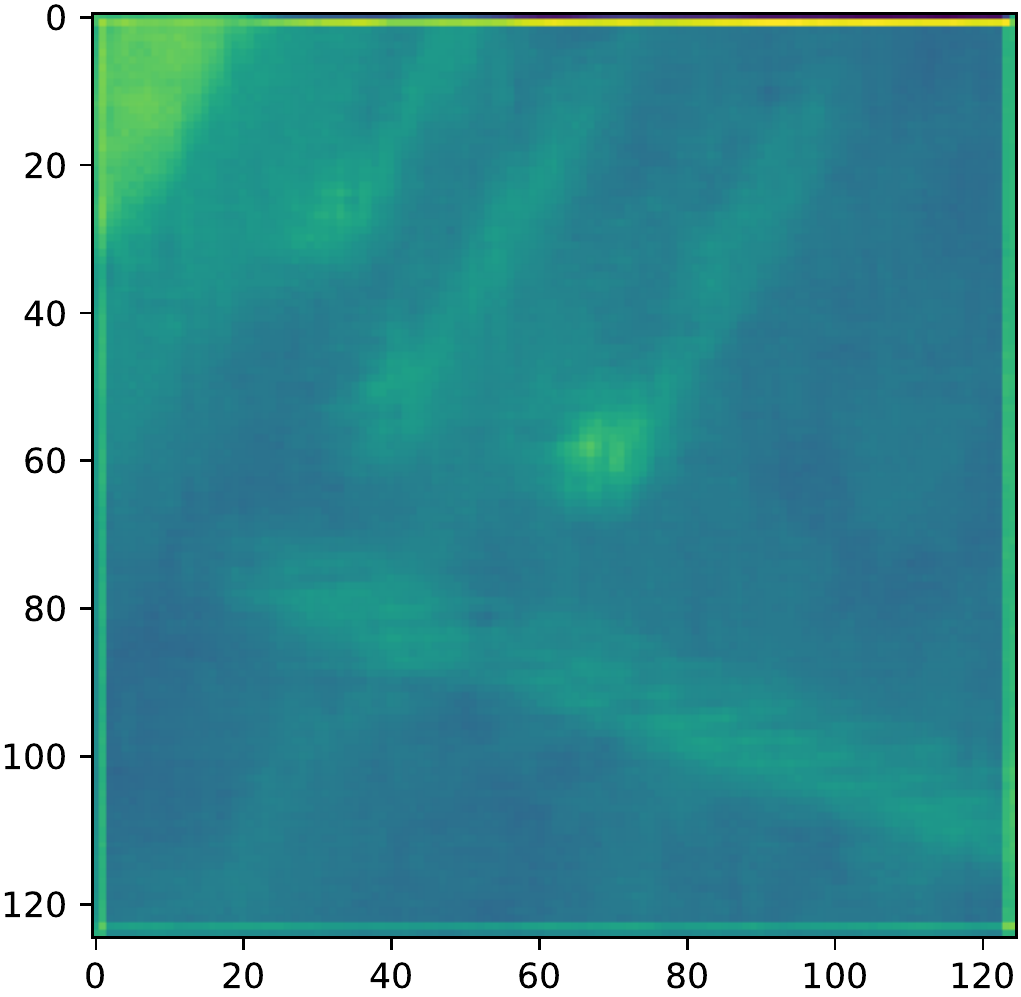} }
    
    \caption{ (a) An input image patch to the last convolutional layer of Stacked-DCFNetand and (b) its corresponding activation map.}%
    \label{fig:activationDCFNet}%
\end{figure}

\autoref{fig:failureCaseDCFNet1} shows a typical failure case of the Stacked-DCFNet method. In the first two frames, most of the objects are tracked correctly; however, after that, the diagonal line in the patch center is confused with the people walking across it. We assume that the line shares similar appearance features with the crossing people. 
\autoref{fig:failureCaseDCFNet2} illustrates another typical failure case of DCFNet. The image includes several people walking closely in different directions, introducing confusion into the tracking method due to the people's similar appearance features.
We closely investigate these failure cases in~\autoref{fig:activationDCFNet}. In this figure, we visualize the activation map of the last convolution layer of the network. Although the convolutional layers of Stacked-DCFNet are supposed to be trained only for people, the line and the people (considering their shadows) appear indistinguishable. Moreover, based on the features, different people cannot be discriminated.
\autoref{fig:successfullTrackingDCFNet} demonstrates a successful tracking case by Stacked-DCFNet. People are not walking closely together and the background is more distinguishable from the people. 
We also evaluated SMSOT-CNN and found that it shares similar failure and success cases with Stacked-DCFNet, as both take advantage of convolutional layers for extracting appearance features.

Altogether, the Euclidean distance paired with trajectory information in EOT works better than IoU for tracking in aerial imagery. However, detection-based trackers such as EOT require object detection in every frame. As shown for Tracktor++, the detection accuracy of the object detectors is very poor for pedestrians in aerial images. Thus, detection-based methods are not appropriate for our scenarios. Moreover, the approaches which employ deep appearance features for re-identification share similar problems with object detectors, features with poor discrimination abilities in the presence of similarly looking objects, leading to ID switches and loosing track of objects. The tracking methods based on regression and correlation (e.g., Stacked-DCFNet and SMSOT-CNN) show, in general, better performances than the methods based on re-identification because they track objects by local image patches that errors to be propagated to the whole image. 
Furthermore, according to our investigations, the path taken by every pedestrian is influenced by three factors: 1) the pedestrian's path history, 2) the positions and movements of the surrounding people, 3) the arrangement of the scene. 

We conclude that both regression- and correlation-based tracking methods are good choices for our scenario. They can be improved by considering trajectory information and the pedestrians movement relationships.

\section{AerialMPTNet}\label{sec:method}
In this section we explain our proposed AerialMPTNet tracking algorithm with its different configurations. Part of its architecture and configurations has been presented in~\cite{kraus2020aerialmptnet}.

As stated in~\autoref{sec:preExperiments}, a pedestrian's movement trajectory is influenced by its movement history, its motion relationships to its neighbours, and scene arrangements. The same holds for the vehicles in traffic scenarios. For the vehicles, there are other constraints such as moving along predetermined paths (e.g., streets, highways, railways) in most of the time. 
Different objects have different motion characteristics such as speed and acceleration. For example, several studies have shown that walking speed of pedestrians are strongly influenced by their age, gender, temporal variations as well as distractions (e.g., cell phone usage), whether the individual is moving in a group or not, and even the size of the city where the event takes place~\cite{rastogi2011design, finnis2006field}. Regarding road traffic, similar factors could influence driving behaviors and movement characteristics (e.g., cell phone usage, age, stress level, and fatigue)~\cite{strayer2004profiles, rakha2007characterizing}. Furthermore, similar to the pedestrians, maneuvers of a vehicle can directly affect the movements of other neighbouring vehicles: for example, if the vehicle brakes, all the following vehicles must brake, too.

The understanding of individual motion patterns is crucial for tracking algorithms, especially when only limited visual information about target objects is available. However, current regression-based tracking methods such as GOTURN and SMSOT-CNN do not incorporate movement histories or relationships between adjacent objects. These networks locate the next position of objects by monitoring a search area in their immediate proximity. Thus, the contextual information provided to the network is limited. Additionally, during the training phase, the networks do not learn how to differentiate the targets from similarly looking objects within the search area. Thus, as discussed in~\autoref{sec:preExperiments}, ID switches and losing of object tracks happen often for these networks in crowded situations or by object intersections.

\begin{figure*}
    \centering
    \includegraphics[width=\textwidth]{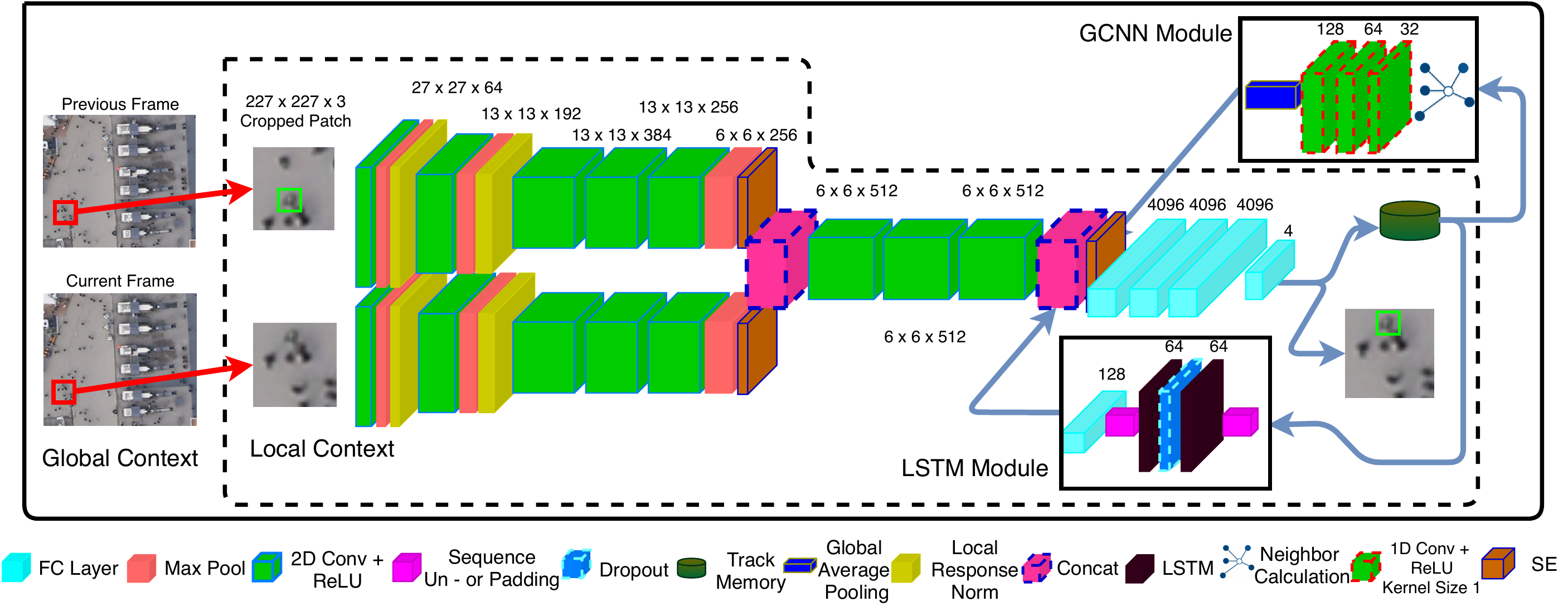}
    \caption[Overview of AerialMPTNet's architecture.]{Overview of the network's architecture composing a SNN, a LSTM and a GraphCNN module. The inputs are two consecutive images cropped and centered to a target object, while the output is the object location in search crop coordinates.}
    \label{fig:modelOverview}
\end{figure*}

In order to tackle the limitations of previous works we propose to fuse visual features, track history, and the movement relationships of adjacent objects in an end-to-end fashion within a regression-based DNN, which we refer to as AerialMPTNet. \autoref{fig:modelOverview} shows an overview of the network architecture. AerialMPTNet takes advantage of a Siamese Neural Network (SNN) for visual features, a Long Short-Term Memory (LSTM) module for movement histories, and a GraphCNN for movement relationships. 
The network takes two local image patches cropped from two consecutive images (previous and current), called target and search patch in which the object location is known and has to be predicted, respectively.
Both patches are centered at the object coordinates known from the previous frame. 
Their size (the degree of contextual information) is correlated with the size of the objects, and it is set to $227\times 227$ pixels to be compatible to the network's input. 
Both patches are then given to the SNN module (retained from~\cite{bahmanyar2019multiple}) composed of two branches of five 2D convolutional, two local response normalization, and three max-pooling layers with shared weights.
Afterwards, the two output features \(Out_{SNN}\) are concatenated and given to three 2D convolutional layers and, finally, four fully connected layers regressing the object position in the search patch coordinates. We use ReLU activations for all these convolutional layers.
The network output is a vector of four values indicating the $x$ and $y$ coordinates of the top-left and bottom-right corners of the objects' bounding boxes. These coordinates are then transformed into image coordinates.
In our network, the LSTM module and the GraphCNN module use the object coordinates in the search patch and image domain, respectively.

\subsection{Long Short-Term Memory Module}

In order to encode movement histories and predict object trajectories, recent works mainly relied on LSTM- and RNN-based structures~\cite{alahi2016social, xue2018ss, vemula2018social}. While these structures have been mostly used for individual objects, due to the large number of objects, we cannot apply these structures directly to our scenarios. Thus, we propose using a structure which treats all object by only one model and predicts the movements (movement vectors) instead of positions.

In order to test our idea, we built an LSTM comprising two bidirectional LSTM layers with 64 dimensions, a dropout layer with \(p=0.5\) in between, and a linear layer which generates two-dimensional outputs, representing the \(x\) and \(y\) values of the movement vector. 
The input of the LSTM module are two-dimensional movement vectors with dynamic lengths up to five steps of the objects' movement histories.
We applied this module to our pedestrian tracking datasets. The results of this experiment show that our LSTM module can predict the next movement vector of multiple pedestrians with about 3.6 pixels (0.43~m) precision, which is acceptable for our scenarios.
Therefore, training a single LSTM on multiple objects would be enough for predicting the objects' movement vectors.

We embed a similar LSTM module into our network as shown in~\autoref{fig:modelOverview}. For the training of the module, the network first generates a sequence of object movement vectors based on the object location predictions. In our experiments, each track has a dynamic history of up to five last predictions. As tracks are not assumed to start at the same time, the length of each track history can be different. Thus, we use zero-padding to make the lengths of track histories similar, allowing to process them together as a batch. These sequences are fed into the first LSTM layer with a hidden size of 64. A dropout with \(p = 0.5\) is then applied to the hidden state of the first LSTM layer, and passes the results to the second LSTM layer. The output features of the second  LSTM  layer are fed into a linear layer of size 128. The 128-dimensional output of the LSTM module \(Out_{LSTM}\) is then concatenated with \(Out_{SNN}\) and \(Out_{Graph}\), the output of the GCNN module. The concatenation allows the network to predict object locations more precisely based on a fusion of appearance and movement features.

\subsection{GraphCNN Module}

The GraphCNN module consists of three 1D convolution layers with $1\times1$ kernels and respectively 32, 64, and 128 channels.
We generate each object's adjacency graph based on the location prediction of all objects. To this end, the eight closest neighbors in a radius of 7.5~m from the object are considered and modeled as a directed graph by a set of vectors $v_i$ from the neighbouring objects to the target object's position $(x,y)$. The resulting graph is represented as \([x, y, x_{v_1}, y_{v_1}, ..., x_{v_8}, y_{v_8}]\).
If less than eight neighbors are existing, we zero-pad the rest of the vectors. 

The GraphCNN module also uses historical information by considering five previous graph configurations. Similarly to the LSTM module, we use zero-padding if less than five previous configurations are available. 
The resulting graph sequences are described by a $18 \times 5$ matrix which is fed into the first convolution layer. In our setup, graph sequences of multiple objects are given to the network as a batch of matrices. The output of the last convolutional layer is gone through a global average pooling in order to generate the final 128-dimensional output of the module \(Out_{Graph}\), which is concatenated to \(Out_{SNN}\) and \(Out_{LSTM}\). The features of the GraphCNN module enable the network to better understand group movements.

\subsection{Squeeze-And-Excitation Layers}\label{sec:squeeze}

During our preliminary experiments in~\autoref{sec:preExperiments}, we experienced a high deviation in the quality of activation maps produced by the convolution layers in DCFNet and SMSOT-CNN. This deviation shows the direct impact of single channels and their importance for the final result of the network. In order to consider this factor in our approach, we model the dominance of the single channels by Squeeze-And-Excitation (SE) layers~\cite{hu2018squeeze}. 

CNNs extract image information by sliding spatial filters across the inputs to different layers. While the lower layers extract detailed features such as edges and corners, the higher layers can extract more abstract structures such as object parts. In this process, each filter at each layer has a different relevance to the network output. However, all filters (channels) are usually weighted equally. Adding the SE layers to a network helps weighting each channel adaptively based on their relevance. In the SE layers, each channel is squeezed to a single value by using global average pooling~\cite{lin2013network}, resulting in a vector with \(k\) entries. This vector is given to a fully connected layer reducing the size of the output vector by a certain ratio, followed by a ReLu activation function. The result is fed into a second fully connected layer scaling the vector back to its original size and applying a sigmoid activation afterwards. In the final step, each channel of the convolution block is multiplied by the results of the SE layer. This channel weighting step adds less than 1\% to the overall computational cost.
As can bee seen in~\autoref{fig:modelOverview}, we add one SE layer after each branch of the SNN module, and one SE layer after the fusion of \(Out_{SNN}\), \(Out_{LSTM}\), and \(Out_{Graph}\).

\subsection{Online Hard Example Mining}\label{sec:ohem}

In the object detection domain, datasets usually contain a large number of easy cases with respect to cases which are challenging for the algorithms. Several strategies have been developed in order to account for this, such as sample-aware loss functions (e.g., Focal Loss \cite{lin2017focal}), where the easy and hard samples are weighted based on their frequencies, and online hard example mining (OHEM)~\cite{Shrivastava2016ohem}, which gives hard examples to the network if they are previously failed to be correctly predicted. The selection and focusing on such hard examples can make the training more effective.
However, in the multi-object tracking domain, such strategies have been rarely used although the tracking datasets suffer from the sample problem as the detection datasets. To the best of our knowledge, none of the previous works in the regression-based tracking used OHEM during their training process.

Thus, in order to deal with the sample imbalance problem of our datasets, we propose adapting and employing OHEM for our training process. To this end, if the tracker loses an object during training, we reset the object to its original starting position and the starting frame, and feed it to the network in the next iteration again. If the tracker fails again, we ignore the sample by removing it from the batch.

\section{Experimental Setup} \label{sec:exp_setup}
For all of our experiments, we used \textit{PyTorch} and \textit{Nvidia Titan XP} GPUs.
We trained all networks with an SGD optimizer and an initial learning rate of \(10^{-6}\).
For all training setups, unless indicated otherwise, we used the $L1$ loss, $L(x, \hat{x}) =|x - \hat{x}|$, where \(x\) and $\hat{x}$ represent the output of the network and ground truth, respectively.
The batch size of all our experiments is 150; however, during offline feedback training, the batch size can differ due to unsuccessful tracking cases and subsequent removal of the object from the batch.

In our experiments, we consider SMSOT-CNN as baseline network and compare different parts of our approach to it.
The original SMSOT-CNN is described in \textit{Caffe}. In order to make it completely comparable to our approach, we re-implemented it in \textit{PyTorch} and trained it.
For the training of SMSOT-CNN, we assign different fractions of the initial learning rate to each layer, as in the original \textit{Caffe} implementation, inspired by the GOTURN's implementation. 
In more detail, we assigned the initial learning rate to each convolutional layer, and assigned a learning rate 10 times larger to the fully connected layers. Weights were initialized by Gaussians with different standard deviations, while biases were initialized by constant values (zero or one), as in the \textit{Caffe} version. 
The training process of SMSOT-CNN is based on a so-called \textit{Example Generator}. Provided with one target image with known object coordinates, this creates multiple examples by creating and shifting the search crop to create different kinds of movements. It is also possible to give the true target and search images. A hyperparameter set to 10 controls the number of examples generated for each image.
For the pedestrian tracking, we also use DLR-ACD to increase the number of available training samples. SMSOT-CNN is trained completely offline and learns to regress the object location based on only the previous location of the object.

For AerialMPTNet, we train the SNN module and the fully connected layers as in SMSOT-CNN. After that, the layers are initialized with the learnt weights, and the remaining layers are initialized with the standard \textit{PyTorch} initialization. Moreover, we decay the learning rate by a factor of 0.1 for every twenty thousand iterations and train AerialMPTNet in an end-to-end fashion by using feedback loops to integrate previous movement and relationship information between adjacent objects. In contrast to the training process of SMSOT-CNN, which is based on artificial movements created by the example generator, we train our networks based on real tracks.

In the training process, a batch of 150 random tracks (i.e., objects from random sequences of the training set) is first selected starting at a random time step between 0 and the track end \(t_{end}-1\). We give the network the target and search patches for these objects. The network's goal is to regress each object position in the search patches consecutively until either the object is lost or the track ends. The target and search patches are generated based on the network predictions in consecutive frames. The object will remain in the batch as long as the network tracks it successfully. If the ground truth object position lies outside of the predicted search area or the track reaches its end frame, we remove the object from the batch and replace it with a new randomly selected object.

For each track and each time step, the network's prediction is stored and used from the LSTM and GraphCNN module. For each object in the batch, the LSTM module is given the objects' movement vectors from the latest time steps up to a maximum number of five, as explained in~\autoref{sec:method}. This process provides the network with an understanding of each object's movement characteristics by a prediction of the next movement. As a result, our network uses its predictions as feedback to improve its performance. 
Furthermore, we perform gradient clipping for the LSTM during training to prevent exploding gradients. The neighbor calculation of the GraphCNN module is also based on the network's prediction of each object's position, as mentioned in~\autoref{sec:method}. Based on the network's prediction of the object position, we search for the nearest neighbors in the ground truth annotation of that frame. However, during the testing phase, we search nearest neighbors based on the network's prediction of the object positions.

For the pedestrian dataset, we set the context factor to 4, with each object with a bounding box size of $4 \times 4$ pixel resulting in an image patch of $16\times 16$ pixels. For vehicle tracking, however, due to the larger sizes of their bounding boxes, we reduce the context factor to 3. This helps avoiding multiple vehicles in a single image patch which could cause track confusion.
\section{Evaluation and Discussion}\label{sec:evaluation}

In this section, we evaluate different parts of our proposed AerialMPTNet on the KIT~AIS and AerialMPT datasets through a set of ablation studies. Furthermore, we compare our results to the tracking methods discussed in \autoref{sec:preExperiments}. \autoref{tab:configs} reports the different network configurations for our ablation studies.
\begin{table}
    \centering
    \caption{Different network configurations.}
    \resizebox{\columnwidth}{!}{%
    \rowcolors{2}{gray!25}{white}
    \begin{tabular}{c|ccccc}
         Name & SNN & LSTM & GCNN & SE Layers & OHEM \\
         \hline
         SMSOT-CNN & \checkmark & $\times$ & $\times$ & $\times$ & $\times$ \\
         AerialMPTNet$_{LSTM}$ & \checkmark & \checkmark  & $\times$ & $\times$ & $\times$\\
         AerialMPTNet$_{GCNN}$ & \checkmark  & $\times$ & \checkmark & $\times$ & $\times$\\
         AerialMPTNet & \checkmark & \checkmark & \checkmark & $\times$ & $\times$ \\
         AerialMPTNet$_{SE}$ & \checkmark & \checkmark & \checkmark & \checkmark & $\times$ \\ 
         AerialMPTNet$_{OHEM}$ & \checkmark & \checkmark & \checkmark & $\times$ & \checkmark \\ 
    \end{tabular}}
    \label{tab:configs}
\end{table}

\subsection{SMSOT-CNN (\textit{PyTorch})}
As mentioned in Section~\ref{sec:exp_setup}, for a better comparison we re-implemented the SMSOT-CNN in the \textit{PyTorch} framework and trained it on our experimental datasets (as the pretrained weights could not be used.)
The tracking results of our \textit{PyTorch} SMOST-CNN on the ArialMPT and KIT~AIS pedestrian and vehicle datasets are presented in~\autoref{tab:smsotAll}.
Therein, SMSOT-CNN achieves a MOTA and MOTP scores of -35.0 and 70.0 for the KIT~AIS pedestrian, and 37.1 and 75.8 for the KIT AIS vehicle dataset, respectively. It achieves respectively a MOTA and MOTP of -37.2 and 68.0 on the AerialMPT dataset.
A comparison of the results to~\cite{bahmanyar2019multiple} shows that our \textit{PyTorch} implementation works rather similarly to the original \textit{Caffe} version, with only 5.2 and 4.0 points smaller MOTA for the KIT~AIS pedestrian and vehicle, respectively.
For the rest of our experiments, we consider the results of this implementation of SMOST-CNN as the baseline for our evaluations.

\SMSOTCNNALL

\subsection{AerialMPTNet (LSTM only)}

In this step, we evaluate the influence of the LSTM module on the tracking performance of our AerialMPTNet. \autoref{tab:arialmptnetLSTM} reports the tracking result of AerialMPTNet$_{LSTM}$ on our experimental datasets. We use the pre-trained weights of SMSOT-CNN to initialize the convolutional weights and biases. For the KIT~AIS pedestrian dataset, we evaluate the effects of freezing the weights during the training of LSTM. The tracking results with frozen and trainable convolutional weights in~\autoref{tab:arialmptnetLSTM} show that the latter improves MOTA and MOTP values by 8.2 and 0.5, respectively. Moreover, the network trained with trainable weights tracks 6.9\% more objects mostly during their lifetimes (MT). We can observe that this increase in performance holds for all sequences with different number of frames and objects.
We can also see that the number of ID switches for frozen weights is smaller with respect to the trainable weights (231 vs. 270). Based on our visual inspections, the smaller number of ID switches is caused by the network with frozen weights losing track of the objects. The network with the trainable weights can track objects for a longer time; however, when the objects get into crowded scenarios, it loses their track by switching their IDs, leading to an increase in the amount of ID switches.
Based on these comparisons, we can argue that the computed features in SNN need fine tuning to some degree in order to work jointly with the LSTM module. That could be the reason why the training with the trainable weights outperforms the setting employing frozen weights. 
Thus, for the rest of our experiments, we use trainable weights. Consequently,~\autoref{tab:arialmptnetLSTM} shows only the results with trainable weights for the AerialMPT and KIT~AIS vehicle datasets.

\autoref{tab:overallperformance} represents the overall performances of different tracking methods on the KIT~AIS and AerialMPT datasets. According to the table, AerialMPTNet$_{LSTM}$ outperforms SMSOT-CNN with significant larger MOTA on all experimental datasets. In particular, based on~\autoref{tab:smsotAll} and~\autoref{tab:arialmptnetLSTM}, the main improvements happen for complex sequences such as the ``AA\_Walking\_02" and ``Munich02" sequences of the KIT~AIS pedestrian dataset, with a 20.8 and 23.8 points larger MOTA, respectively.
On the AerialMPT dataset, the most complex sequences are ``Bauma3" and ``Bauma6" presenting overcrowded scenarios with many pedestrians intersecting. According to the results, using the LSTM module does not help the performance relevantly. In such complex sequences, the trajectory information of the LSTM module is not enough for distinguishing pedestrians and tracking them within the crowds.
Furthermore, the increase in the number of mostly and partially tracked objects (MT and PT) and the decrease in the number of mostly lost ones (ML) indicate that the LSTM module helps AerialMPTNet in the tracking of the objects for a longer time. This, however, causes a larger number of ID switches as discussed before. 
On the KIT~AIS vehicle dataset, although the results show a significant improvement of AerialMPTNet$_{LSTM}$ over SMSOT-CNN, the performance improvements are minor compared to the pedestrian datasets. This could be due the more distinguishable appearance features of the vehicles, leading to a good performance even when relying solely on the SNN module.
\aerialmptnetLSTM
\TotalResults
\subsection{AerialMPTNet (GCNN only)}

In this step, we focus on the modeling of the movement relationships between adjacent objects by AerialMPTNet$_{GCNN}$. As described in~\autoref{tab:configs}, we only consider the SNN and GCNN modules, and train the network on our experimental datasets. The tracking results on the test sequences of the datasets are shown in~\autoref{tab:gcnn}, and the comparisons to the other methods are provided in~\autoref{tab:overallperformance}.
AerialMPTNet$_{GCNN}$ outperforms SMSOT-CNN significantly with an improvement of 11.8, 12.0, and 5.7 points of the MOTA on the AerialMPT and KIT~AIS pedestrian and vehicle datasets, respectively. Additionally, AerialMPTNet$_{GCNN}$ enhances the MT, PT, and ML values for the pedestrian datasets, while only the MT is enhanced and the PT and ML get worse for the vehicle dataset.
Altogether, these results indicate that the relational information is more important for the pedestrians than the vehicles.
Moreover, according to~\autoref{tab:gcnn}, as in LSTM results, the use of GCNN helps more for complex sequences. For example, MOTA on the ``AA\_Walking\_02" and ``Munich02" sequences increase by 13.9 and 20.5, respectively; however, it decreases respectively by 12.1 and 14.8 on ``AA\_Crossing\_02" and ``RaR\_Snack\_Zone\_02". This could be due to the negative impact of the large number of zero paddings in the less crowded sequences with smaller number of adjacent objects.
Compared to AerialMPTNet$_{LSTM}$, for the AerialMPT, AerialMPTNet$_{GCNN}$ performs slightly better while on the other two datasets it performs worse with a narrow margin. We assume that, due to the higher crowd densities in the AerialMPT dataset, the relationships between adjacent objects are more critical with respect to their movement histories.

\aerialmptnetGCNN

\subsection{AerialMPTNet}\label{sec:aerialMPTNet}

In this step, we evaluate the complete AerialMPTNet by fusing the SNN, LSTM, and GCNN modules.
\autoref{tab:aerialMPTNetresults} represents the tracking results of AerialMPTNet on the test sets of our experimental datasets, and~\autoref{tab:overallperformance} compares its overall performance to the other tracking methods. 
According to the results, the AerialMPTNet outperforms AerialMPTnet$_{LSTM}$ and AerialMPTNet$_{GCNN}$ for both pedestrian datasets. However, this is not the case for the vehicle dataset.
This is due to the main idea behind the development of the network. Since AerialMPTNet is initially designed for pedestrian tracking, it needs to be further adapted to domain specific challenges posed by vehicle tracking. For example, the distance threshold for the modeling if the adjacent object relationships (in GCNN) which considers objects within a distance of 50 pixels from the target object might miss many neighbouring vehicles, as usually the distances between vehicles are larger than those between pedestrians.
Finally, AerialMPTNet achieves better tracking results than SMSOT-CNN on all three datasets.

\aerialMPTNet

\subsubsection{Pedestrian Tracking}
\begin{figure}%
    \centering
       \subfloat{\begin{overpic}[width=.49\columnwidth]{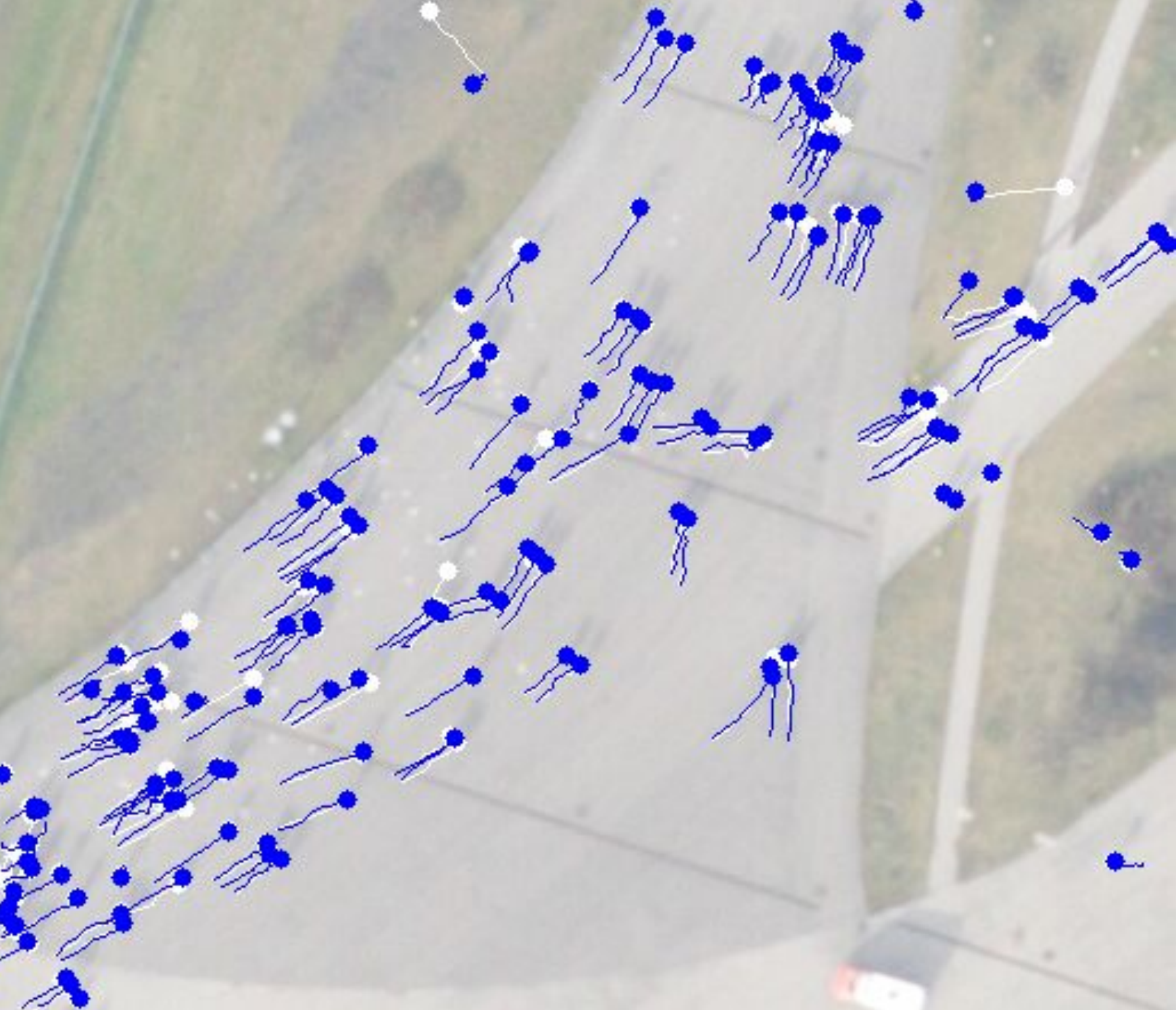}
         \put(5,75){8}
       \end{overpic}}
       \subfloat{\begin{overpic}[width=.49\columnwidth]{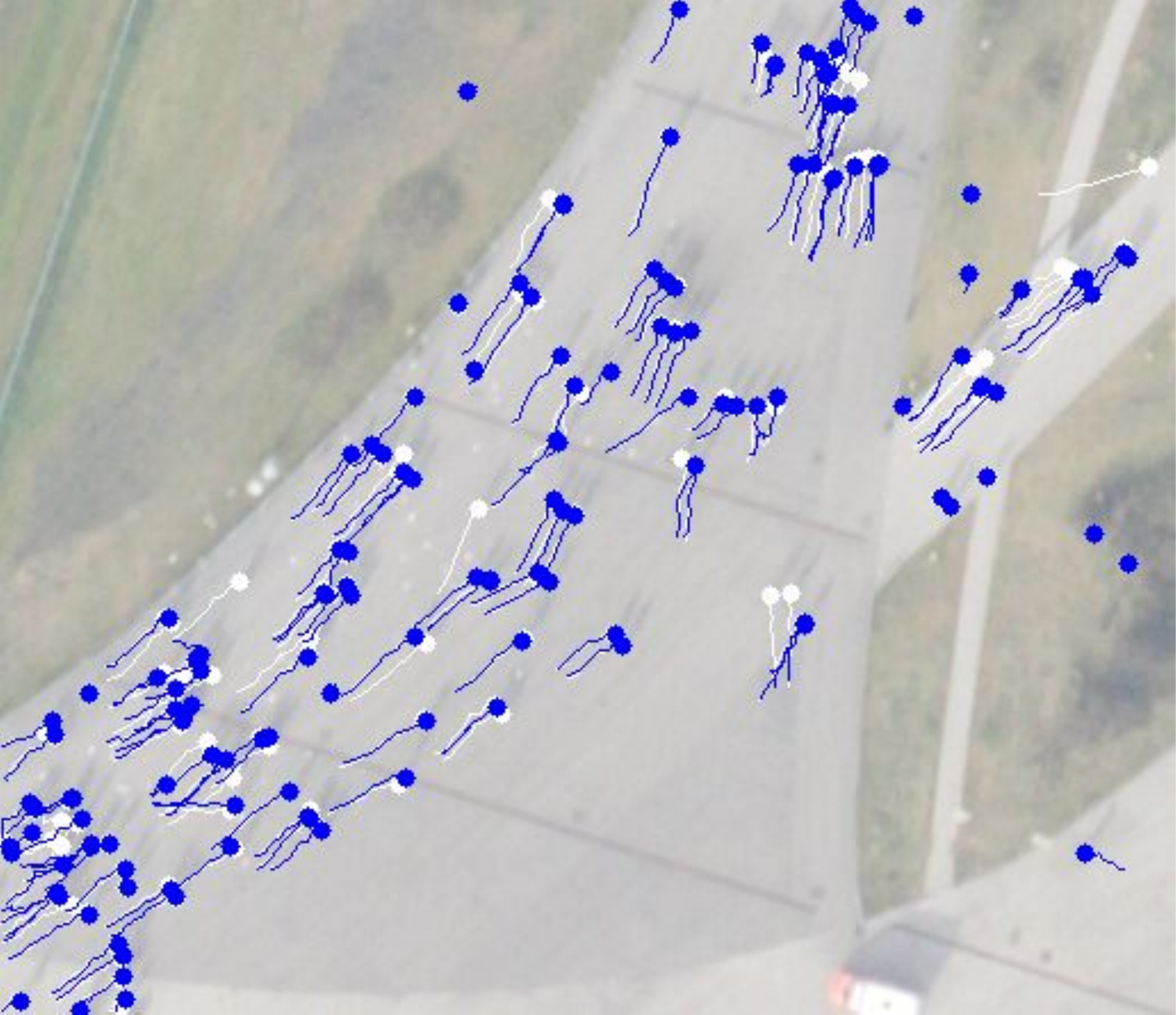}
         \put(5,75){14}
       \end{overpic}}
               \\[1ex]
      \subfloat{\begin{overpic}[width=.49\columnwidth]{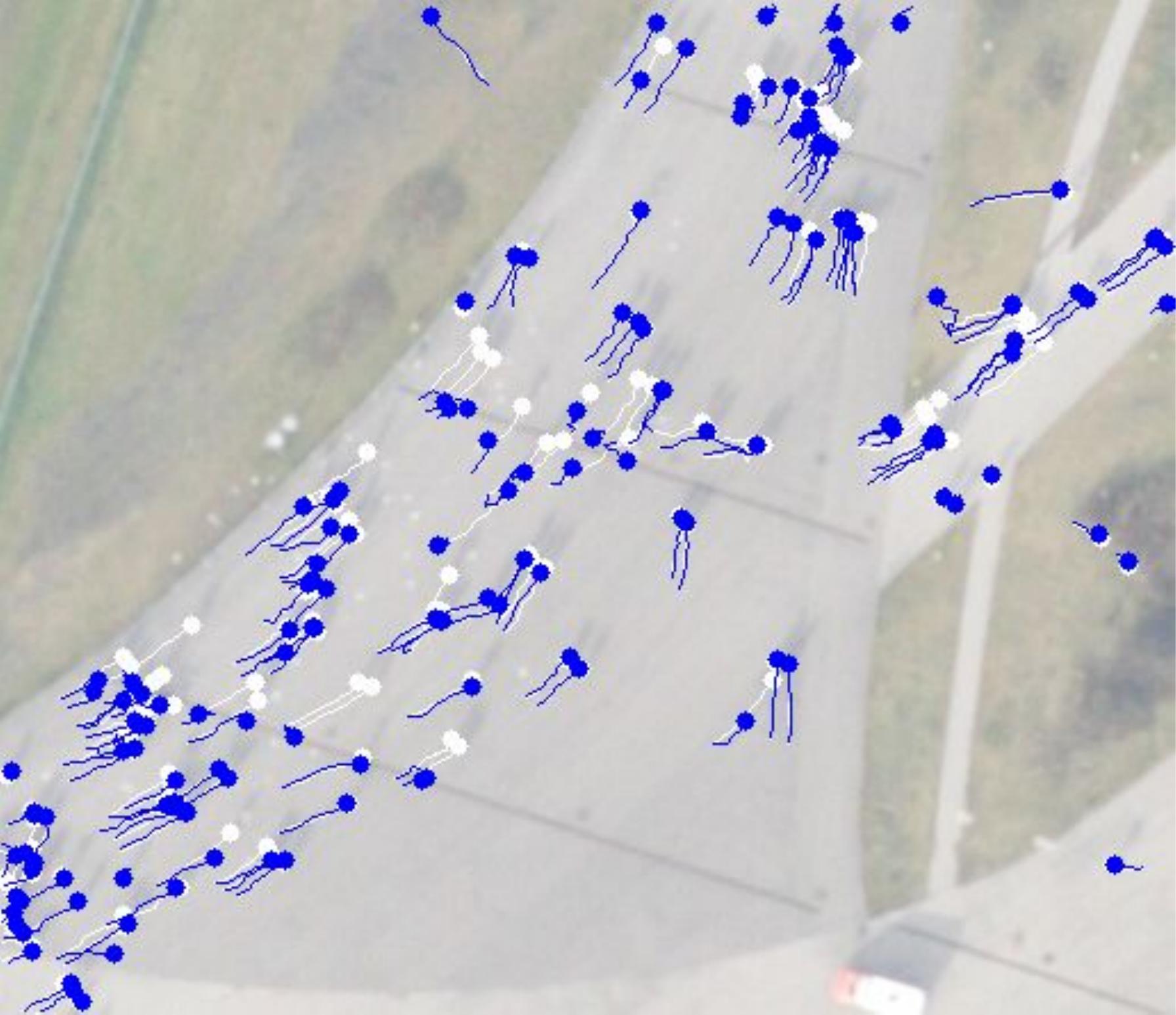}
       \put(5,75){8}
       \end{overpic}}
       \subfloat{\begin{overpic}[width=.49\columnwidth]{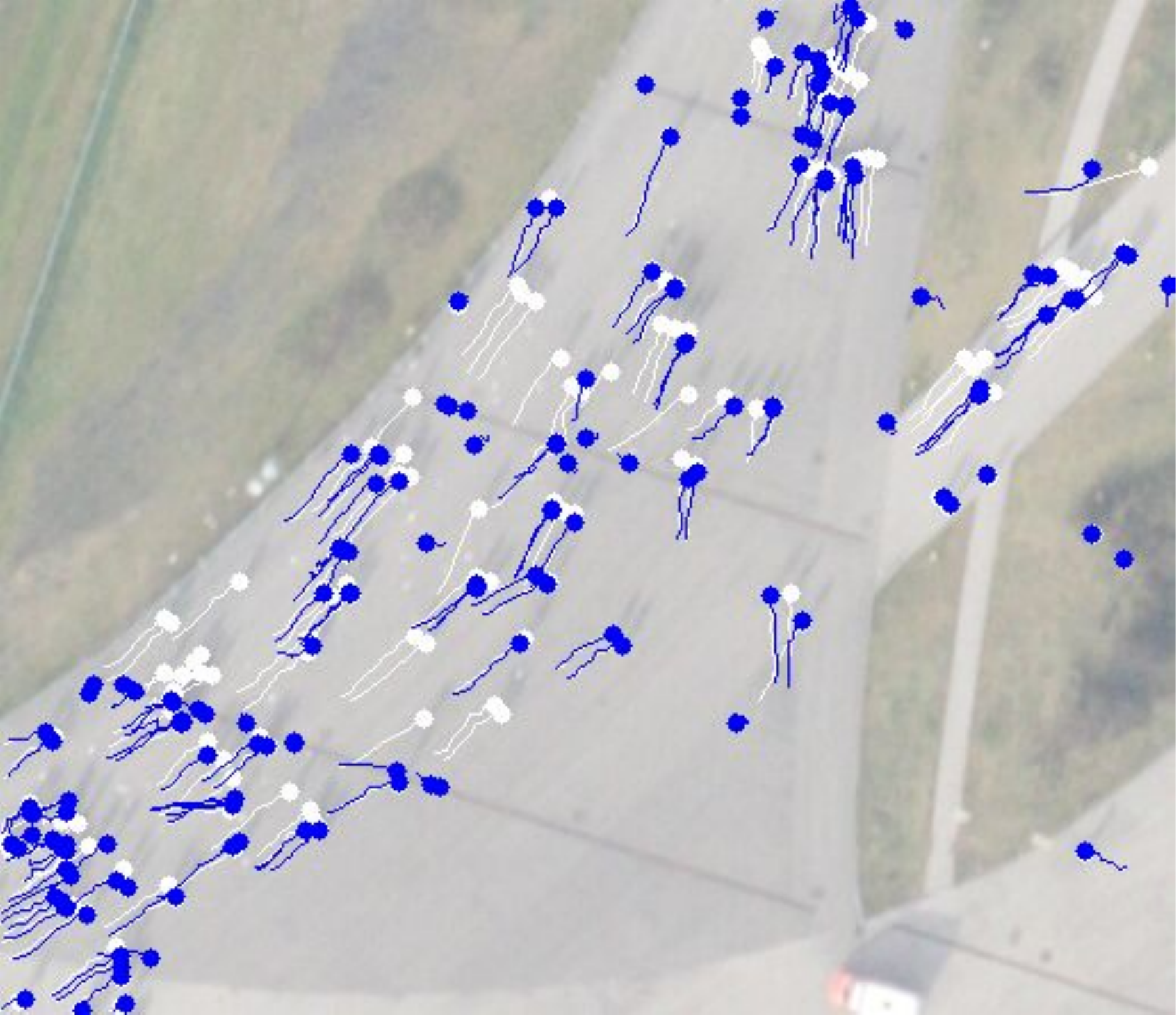}
         \put(5,75){14}
       \end{overpic}}

    \caption{Tracking results by AerialMPTNet (top row) and SMSOT-CNN (bottom row)  on the frames 8 and 14 of the ``AA\_Walking\_ 02" sequence of the KIT~AIS pedestrian dataset. The predictions and ground truth are depicted in blue and white, respectively.}
    \label{fig:smsotcnnWalking}%
\end{figure}

\begin{figure}%
    \centering
   \subfloat{\begin{overpic}[width=.24\columnwidth]{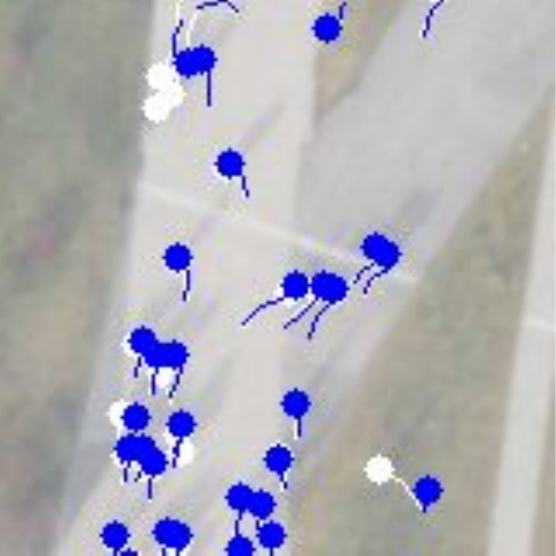}
         \put(5,80){4}
       \end{overpic}}
   \subfloat{\begin{overpic}[width=.24\columnwidth]{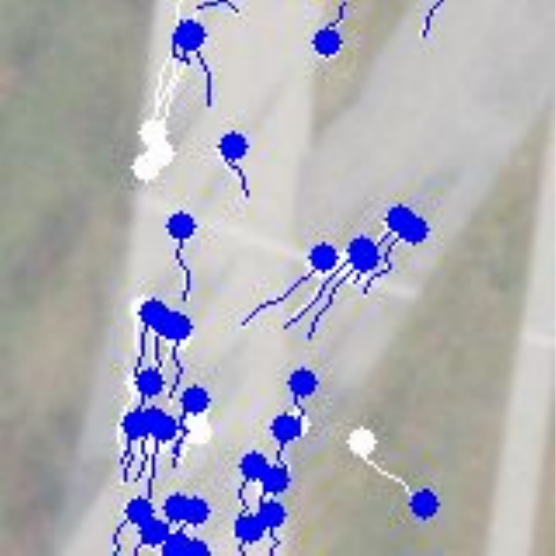}
         \put(5,80){6}
       \end{overpic}}
  \subfloat{\begin{overpic}[width=.24\columnwidth]{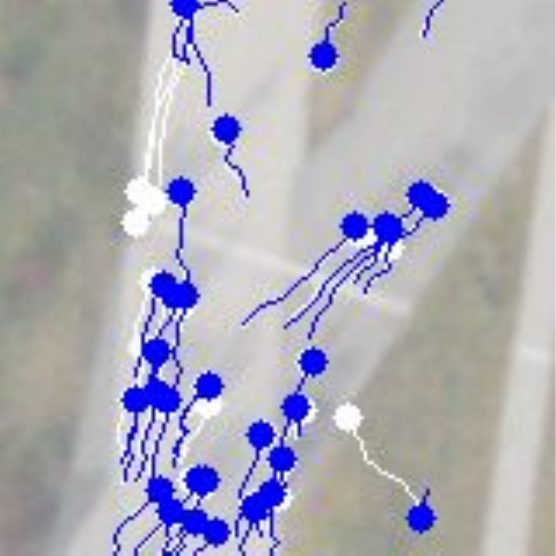}
         \put(5,80){8}
       \end{overpic}}
  \subfloat{\begin{overpic}[width=.24\columnwidth]{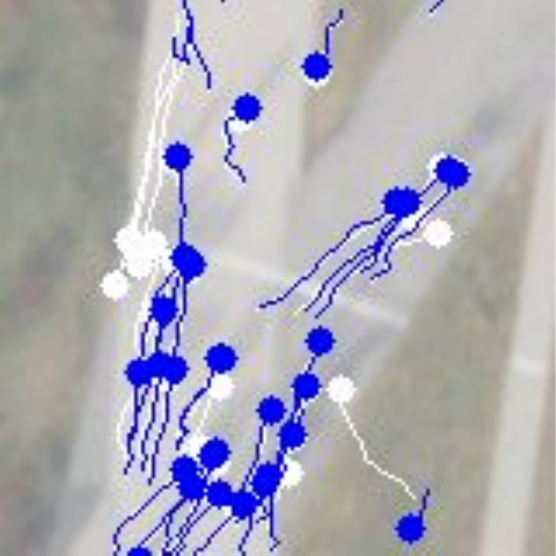}
         \put(5,80){10}
       \end{overpic}}    
    \\[1ex]
   \subfloat{\begin{overpic}[width=.24\columnwidth]{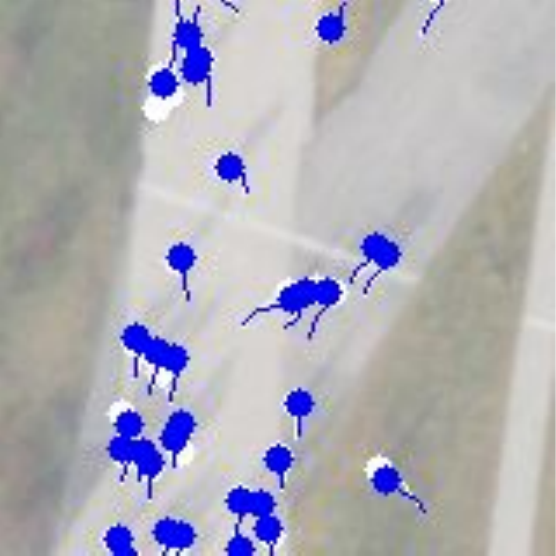}
         \put(5,80){4}
       \end{overpic}}
   \subfloat{\begin{overpic}[width=.24\columnwidth]{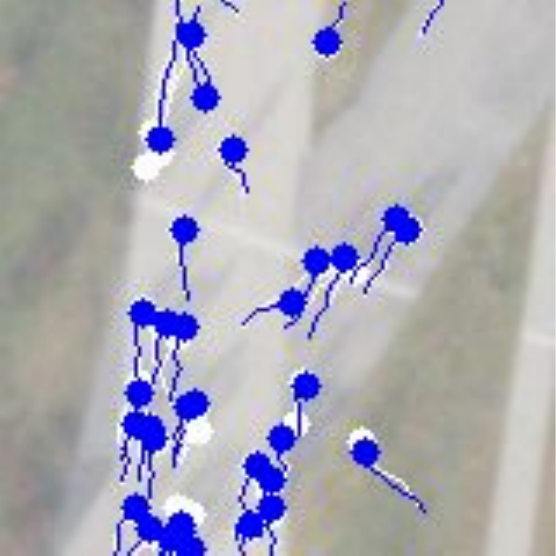}
         \put(5,80){6}
       \end{overpic}}
  \subfloat{\begin{overpic}[width=.24\columnwidth]{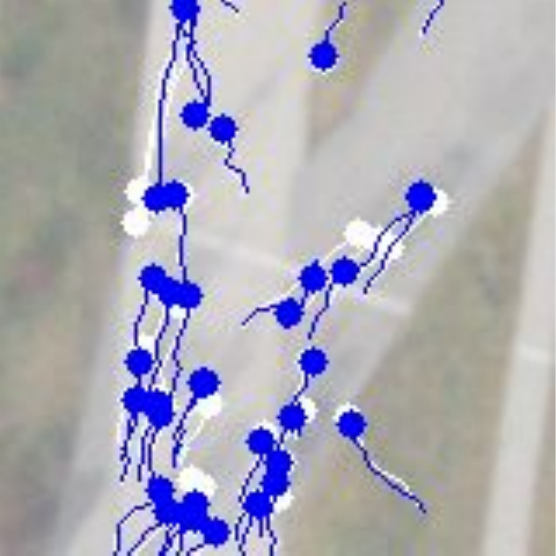}
         \put(5,80){8}
       \end{overpic}}
  \subfloat{\begin{overpic}[width=.24\columnwidth]{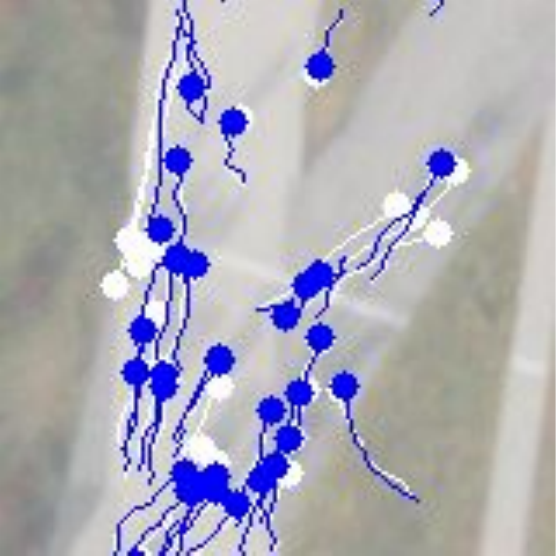}
         \put(5,80){10}
       \end{overpic}}    
    
    \caption{Tracking results by AerialMPTNet (top row) and SMSOT-CNN (bottom row) on the frames 4, 6, 8, and 10 of the ``AA\_Crossing\_02" sequence of the KIT~AIS pedestrian dataset. The predictions and ground truth are depicted in blue and white, respectively.}
    \label{fig:aacrossing}%
\end{figure}

In more details, AerialMPTNet yields the best MOTA among the studied methods on the ``AA\_Walking\_ 02", ``Munich02", and ``RaR\_Snack\_Zone\_02" sequences of the KIT~AIS pedestrian dataset (-16.8, -34.5, and 38.9, respectively.) These sequences are the most complex ones in this dataset with respect to the length and number of objects, thing which could significantly influence the MOTA value. Longer sequences and a higher number of objects usually cause the MOTA value to decrease, as it is more probable that the tracking methods lose track of the objects or confuse their IDs in these cases.
\autoref{fig:smsotcnnWalking} illustrates the tracking results on two frames of the ``AA\_Walking\_ 02" sequence of the KIT~AIS pedestrian dataset by AerialMPTNet and SMSOT-CNN.
Comparing the predictions and ground truth points demonstrates that SMSOT-CNN loses track of a considerably higher number of pedestrians between these two frames. While SMSOT-CNN's predictions are stuck at the diagonal background lines due to their similar appearance features to the pedestrians, AerialMPTNet can easily handle this situation due to the LSTM module. 
We also visualized a cropped part of four frames from the ``AA\_Crossing\_02" sequence of the KIT~AIS pedestrian dataset in~\autoref{fig:aacrossing}. As in the previous example, AerialMPTNet clearly outperforms SMSOT-CNN on the tracking of the pedestrians crossing the background lines. 

On the AerialMPT dataset, AerialMPTNet achieves the best MOTA scores among all studied methods in this paper on the ``Bauma3", ``Bauma6", and ``Witt" sequences (-32.0, -28.4, -65.9), which contain the most complex scenarios regarding crowd density, pedestrian movements, variety of the GSDs, and complexity of the terrain.
However, in contrast to the KIT~AIS pedestrian dataset, the MOTA scores are not correlated with the sequence lengths, indicating the impact of other complexities on the tracking results and the better distribution of complexities over the sequences of the AerialMPT dataset as compared to the KIT~AIS pedestrian dataset.

\begin{figure}%
    \centering
       \subfloat{\begin{overpic}[width=.24\columnwidth]{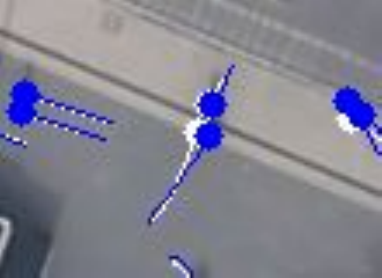}
         \put(5,58){11}
       \end{overpic}}
   \subfloat{\begin{overpic}[width=.24\columnwidth]{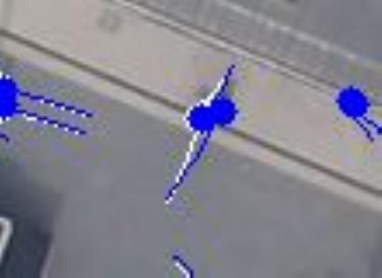}
         \put(5,58){13}
       \end{overpic}}
  \subfloat{\begin{overpic}[width=.24\columnwidth]{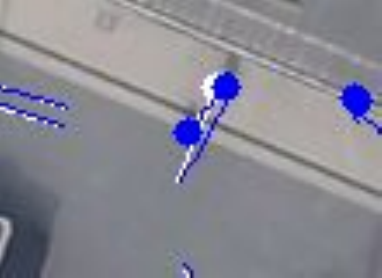}
         \put(5,58){15}
       \end{overpic}}
  \subfloat{\begin{overpic}[width=.24\columnwidth]{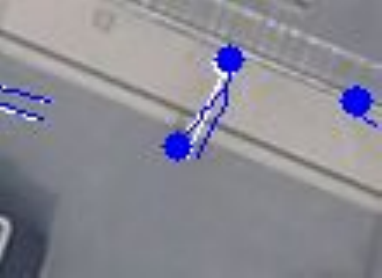}
         \put(5,58){17}
       \end{overpic}}    
    \\[1ex]
   \subfloat{\begin{overpic}[width=.24\columnwidth]{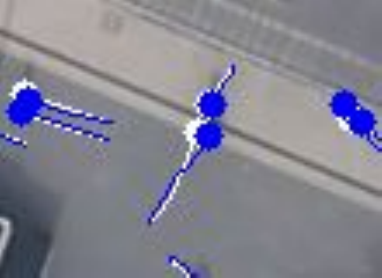}
         \put(5,58){11}
       \end{overpic}}
   \subfloat{\begin{overpic}[width=.24\columnwidth]{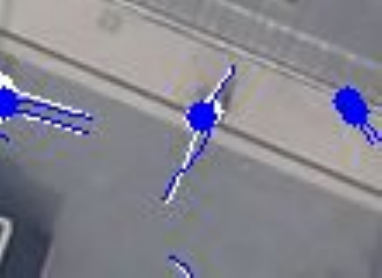}
         \put(5,58){13}
       \end{overpic}}
  \subfloat{\begin{overpic}[width=.24\columnwidth]{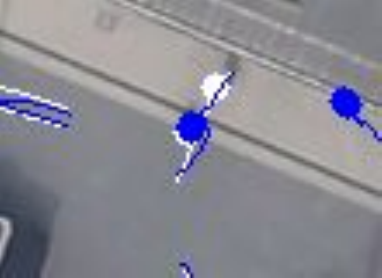}
         \put(5,58){15}
       \end{overpic}}
  \subfloat{\begin{overpic}[width=.24\columnwidth]{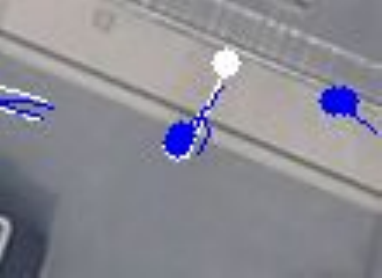}
         \put(5,58){17}
       \end{overpic}}  
    
    \caption{Tracking results by the AerialMPTNet (top row) and SMSOT-CNN (bottom row)  on the frames 11, 13, 15, and 17 of the ``Pasing8" sequence of the AerialMPT dataset. The predictions and ground truth are depicted in blue and white, respectively.}
    \label{fig:aerialmptnetcrossing}%
\end{figure}

\begin{figure}%
    \centering
   \subfloat{\begin{overpic}[width=.24\columnwidth]{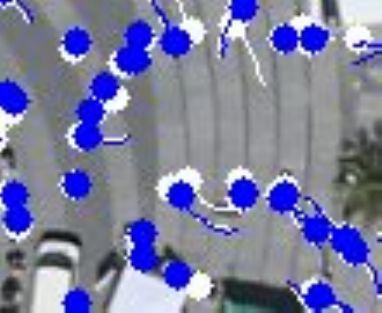}
         \put(1,66){\color{red}21}
       \end{overpic}}
   \subfloat{\begin{overpic}[width=.24\columnwidth]{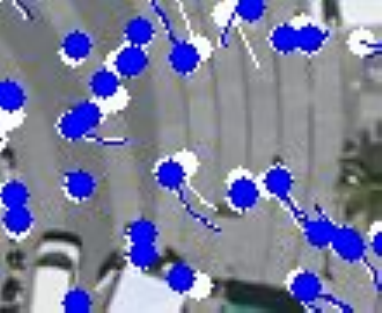}
         \put(1,66){\color{red}23}
       \end{overpic}}
  \subfloat{\begin{overpic}[width=.24\columnwidth]{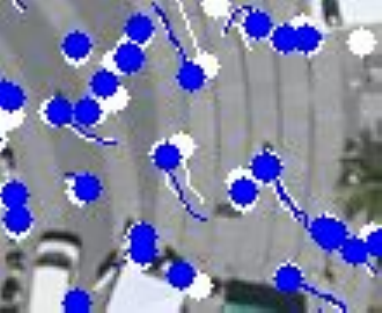}
         \put(1,66){\color{red}25}
       \end{overpic}}
  \subfloat{\begin{overpic}[width=.24\columnwidth]{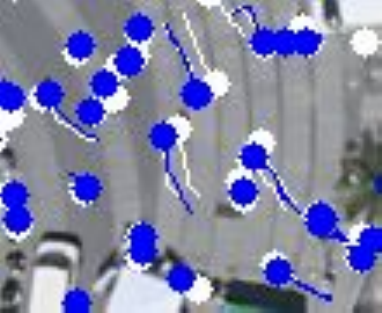}
         \put(1,66){\color{red}27}
       \end{overpic}}    
    \\[1ex]
   \subfloat{\begin{overpic}[width=.24\columnwidth]{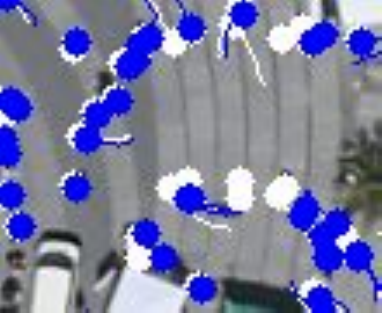}
         \put(1,66){\color{red}21}
       \end{overpic}}
   \subfloat{\begin{overpic}[width=.24\columnwidth]{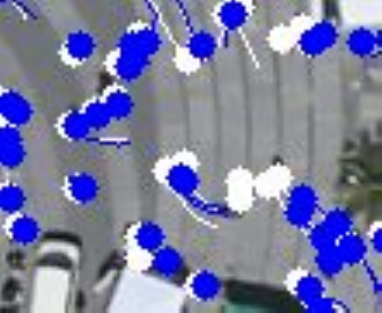}
         \put(1,66){\color{red}23}
       \end{overpic}}
  \subfloat{\begin{overpic}[width=.24\columnwidth]{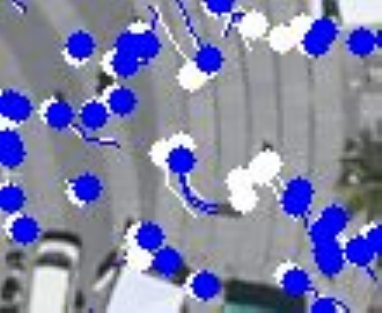}
         \put(1,66){\color{red}25}
       \end{overpic}}
  \subfloat{\begin{overpic}[width=.24\columnwidth]{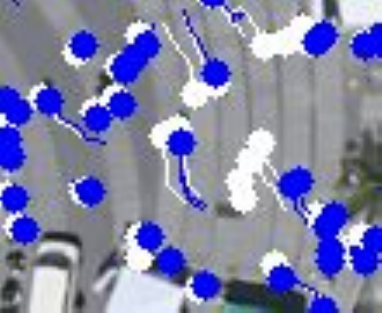}
         \put(1,66){\color{red}27}
       \end{overpic}}  
   
    \caption{Tracking results by the AerialMPTNet (top row) and SMSOT-CNN (bottom row)  on the frames 21, 23, 25, and 27 of the ``Karlsplatz" sequence of the AerialMPT dataset. The predictions and ground truth are depicted in blue and white, respectively.}
    \label{fig:aerialmptnetkarlsplatz}%
\end{figure}

\autoref{fig:aerialmptnetcrossing} exemplifies the role of the LSTM module in enhancing the tracking performance in AerialMPTNet. This figure shows an intersection of two pedestrians in the cropped patches from four frames of the ``Pasing8" sequence of the AerialMPT dataset. According to the results, SMOT-CNN (bottom row) loses one of the pedestrians after their intersection leading to an ID switch. However, AerialMPTNet (top row) can track both pedestrians correctly, mainly relying on the pedestrians' movement histories (their movement directions) provided by the LSTM module.
\autoref{fig:aerialmptnetkarlsplatz} illustrates a case in which the advantage of the GCNN module can be clearly observed. The images are cropped from four frames of the ``Karlsplatz" sequence of the AerialMPT dataset. It can be seen that SMSOT-CNN has difficulties in tracking the pedestrians in such crowded scenarios, where the pedestrians move in various directions. However, AerialMPTNet can handle this scenario mainly based on the pedestrian relationship models provided by the GCNN module.

\begin{figure}%
    \centering
  \subfloat{\begin{overpic}[width=.35\columnwidth, angle =90]{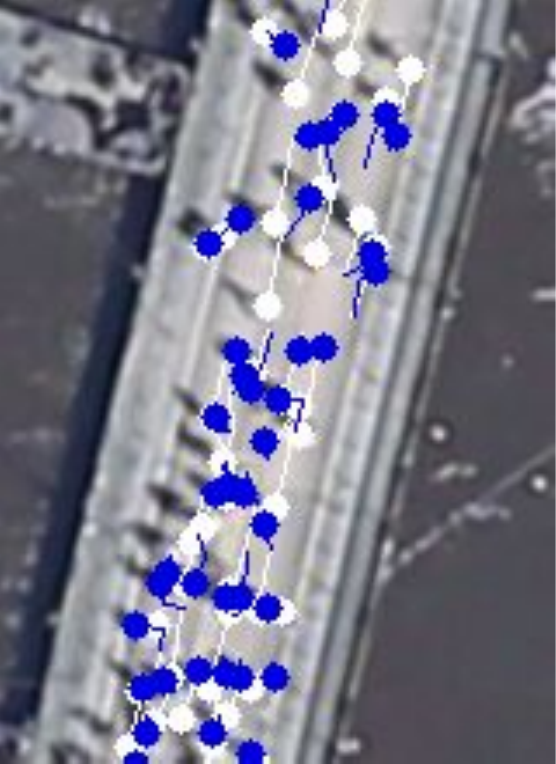}
         \put(90,64){\color{red}3}
       \end{overpic}}
  \subfloat{\begin{overpic}[width=.35\columnwidth, angle =90]{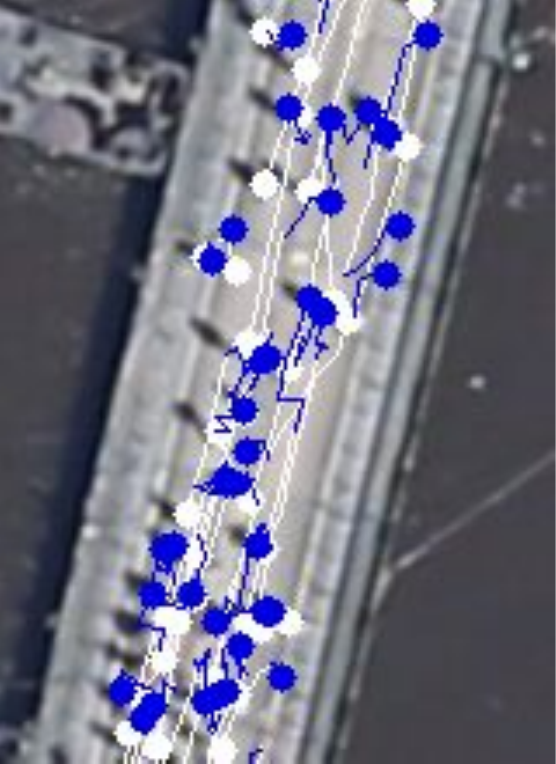}
         \put(90,64){\color{red}6}
       \end{overpic}}    
    \\[1ex]
   \subfloat{\begin{overpic}[width=.35\columnwidth, angle =90]{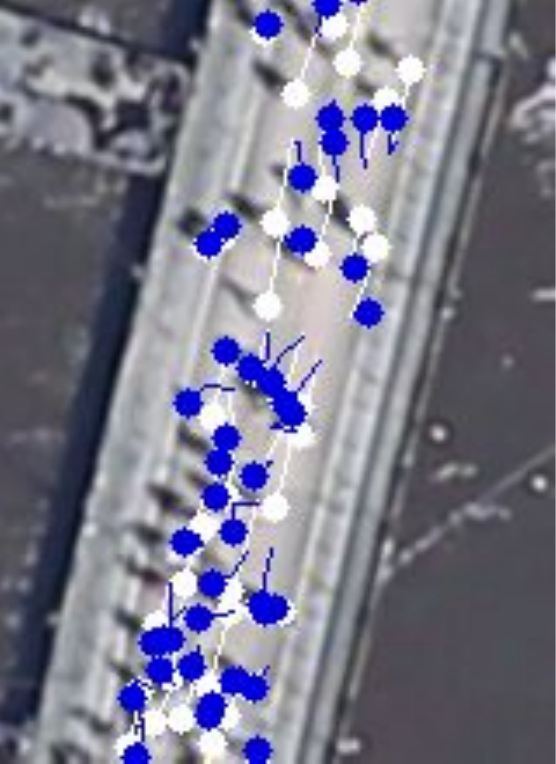}
         \put(90,64){\color{red}3}
       \end{overpic}}
   \subfloat{\begin{overpic}[width=.35\columnwidth, angle =90]{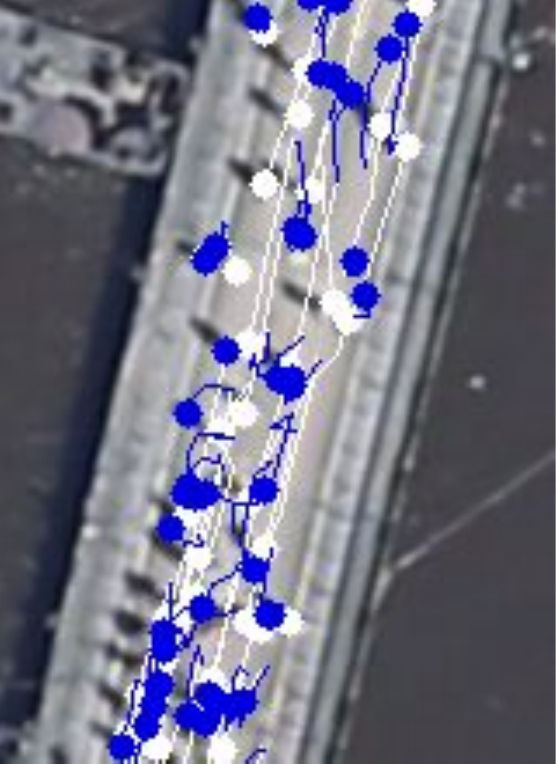}
         \put(90,64){\color{red}6}
       \end{overpic}}

    \caption{Tracking results by AerialMPTNet (top row) and SMSOT-CNN (bottom row)  on the frames 3 and 6 of the ``Witt" sequence of the AerialMPT dataset. The predictions and ground truth are depicted in blue and white, respectively.}
    \label{fig:witt}%
\end{figure}

In addition, there are sequences where both methods reach their limits and perform poorly. \autoref{fig:witt} illustrates the tracking results of AerialMPTNet (top row) and of SMSOT-CNN (bottom row) on two frames of the ``Witt" sequence of the AerialMPT dataset. Comparing the predictions and ground truth object tracks indicates the large number of lost objects by both methods. According to~\autoref{tab:aerialMPTNetresults} and~\autoref{tab:smsotAll}, despite the small number of frames in the ``Witt" sequence, the MOTA scores are low for both methods (-68.6 and -65.9). Further investigations show that these poor performances are caused by the non-adaptive search window size. In the ``Witt" sequence, pedestrians move out of the search window and are lost by the tracker as a consequence. In order to solve this issue, the GSD of the frames as well as the pedestrian velocities should be considered in determining the search window size.

In order to show the complexity of the pedestrian tracking task in the AerialMPT dataset, we report the tracking results of AerialMPTNet on the frames 18 and 10 of the ``Munich02" and ``Bauma3" sequences, respectively, in~\autoref{fig:overview}.
%

\subsubsection{Vehicle Tracking}

\begin{figure}%
    \centering
      \subfloat{\begin{overpic}[width=.54\columnwidth, angle=90]{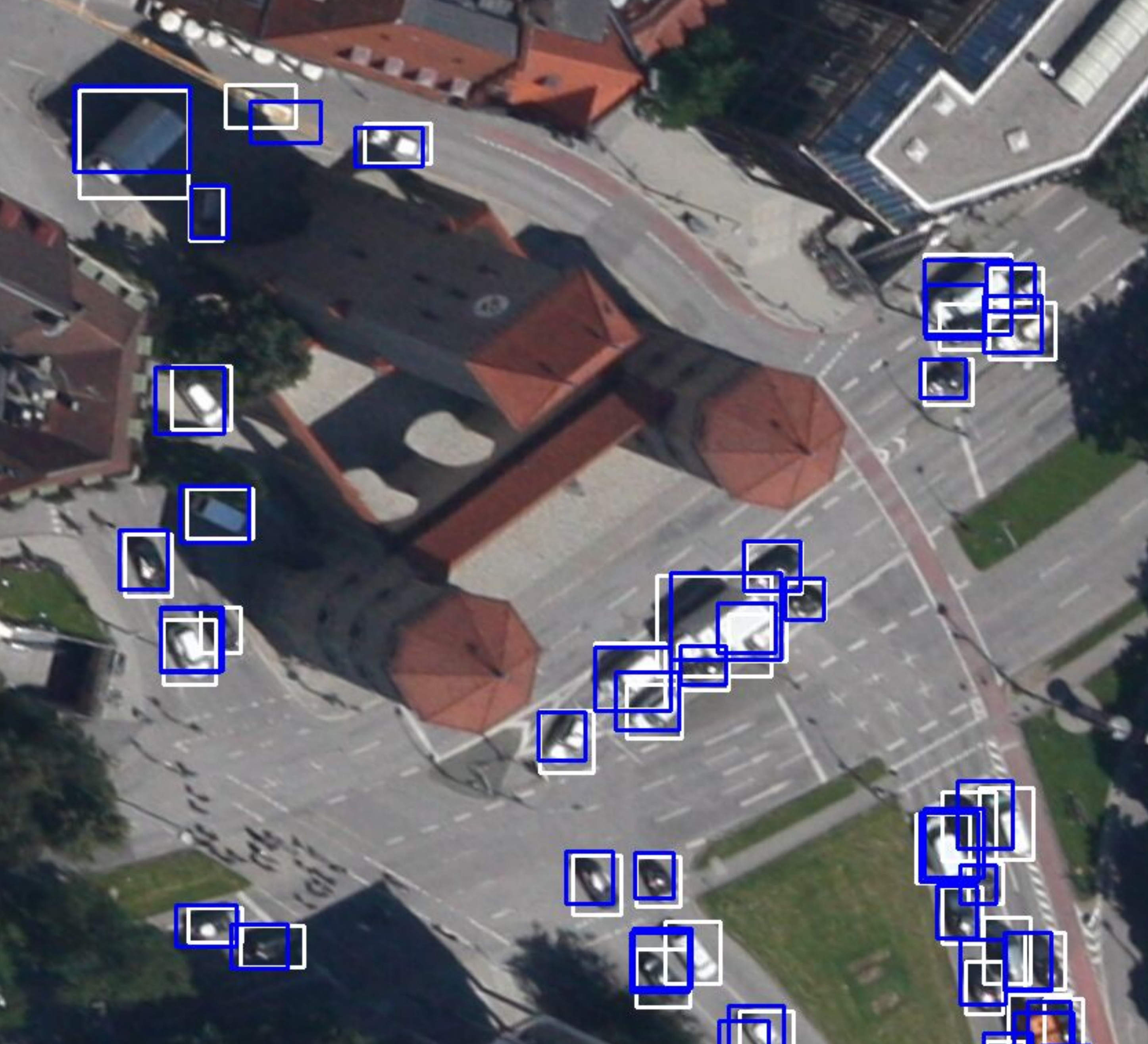}
         \put(3,85){\color{red}4}
       \end{overpic}}    
   \subfloat{\begin{overpic}[width=.54\columnwidth, angle=90]{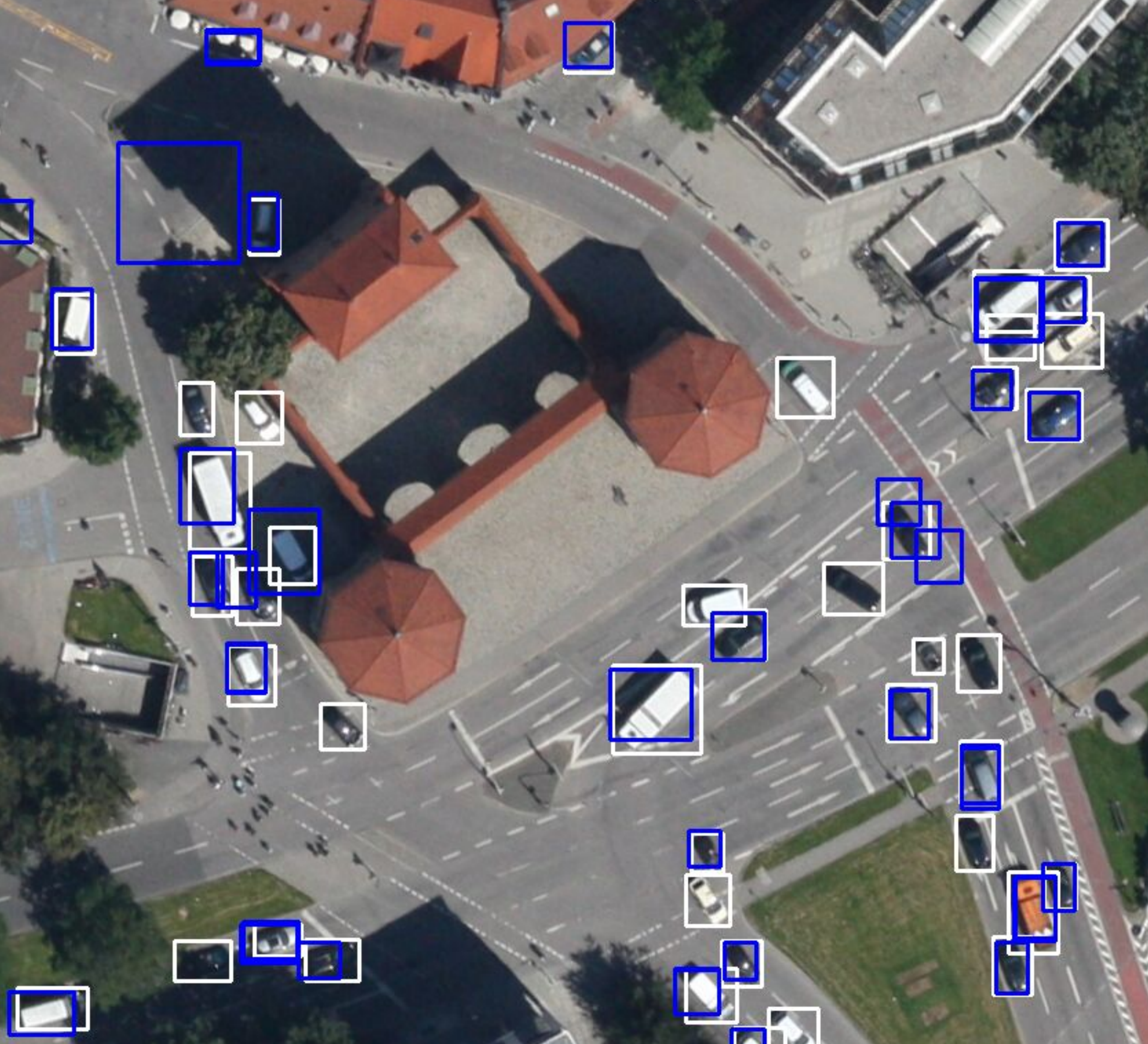}
         \put(3,85){\color{red}31}
       \end{overpic}}
    
    \caption{Tracking results by AerialMPTNet on the frames 4 and 31 of the ``MunichCrossroad02" sequence of the KIT~AIS vehicle dataset. The predictions and ground truth bounding boxes are depicted in blue and white, respectively. Several hindrances such as changing viewing angle, shadows, and occlusions (e.g., by trees) are visible.}
    \label{fig:crossroad2}%
\end{figure}

\begin{figure}%
    \centering
  \subfloat{\begin{overpic}[width=.53\columnwidth, angle =90]{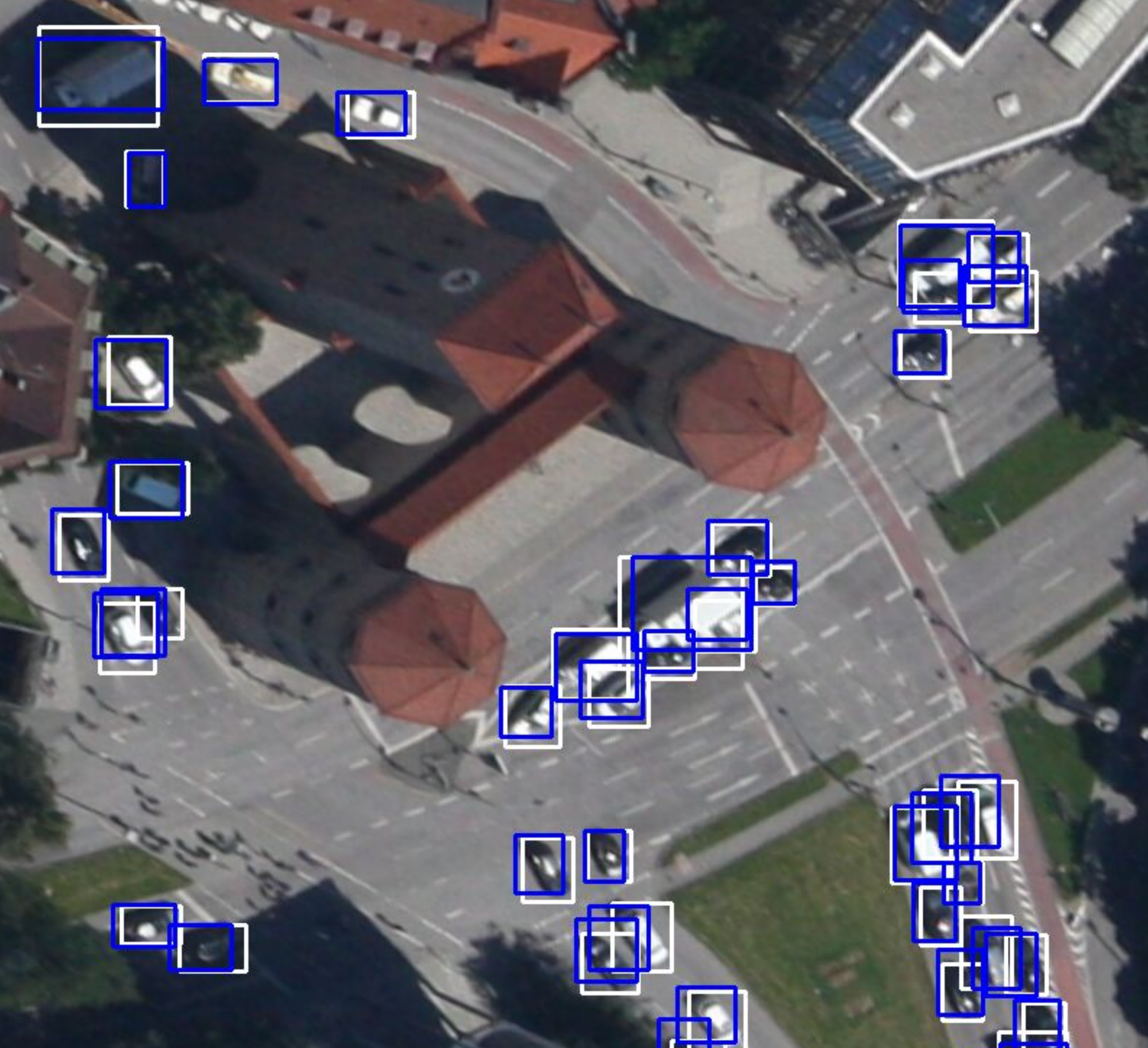}
         \put(3,86){\color{red}2}
       \end{overpic}}
  \subfloat{\begin{overpic}[width=.53\columnwidth, angle =90]{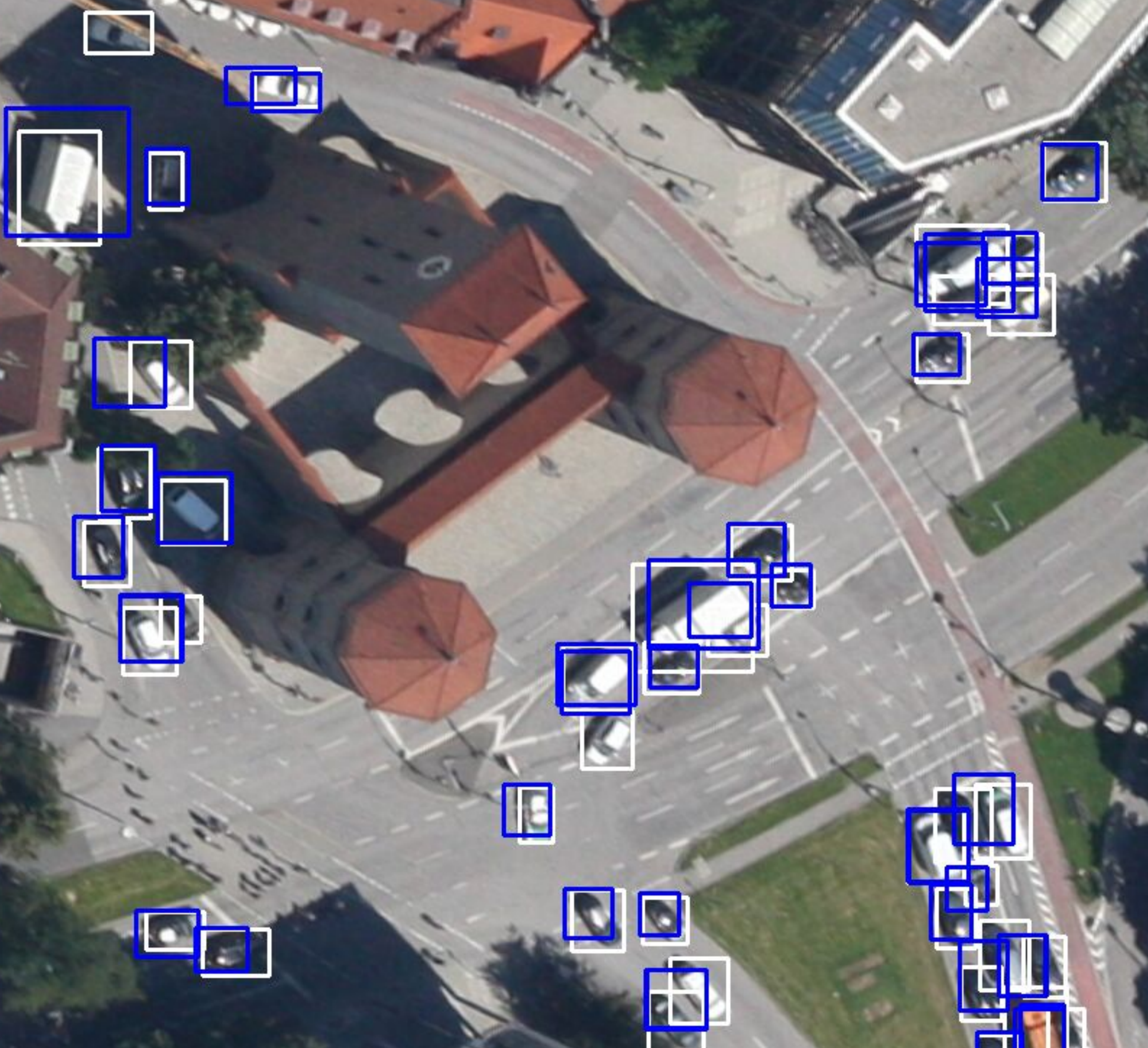}
         \put(3,86){\color{red}8}
       \end{overpic}}    
    \\[1ex]
   \subfloat{\begin{overpic}[width=.53\columnwidth, angle =90]{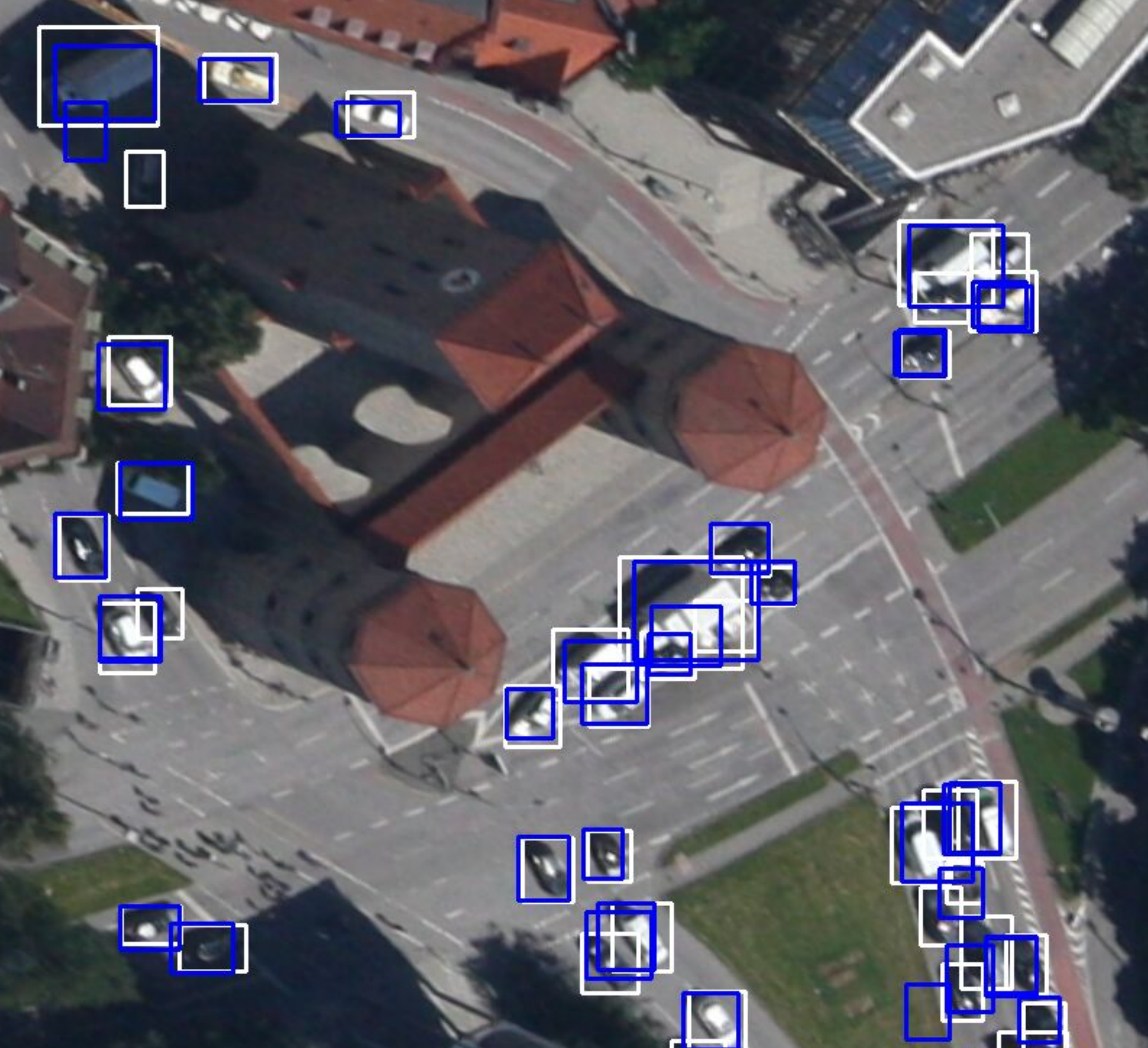}
         \put(3,86){\color{red}2}
       \end{overpic}}
   \subfloat{\begin{overpic}[width=.53\columnwidth, angle =90]{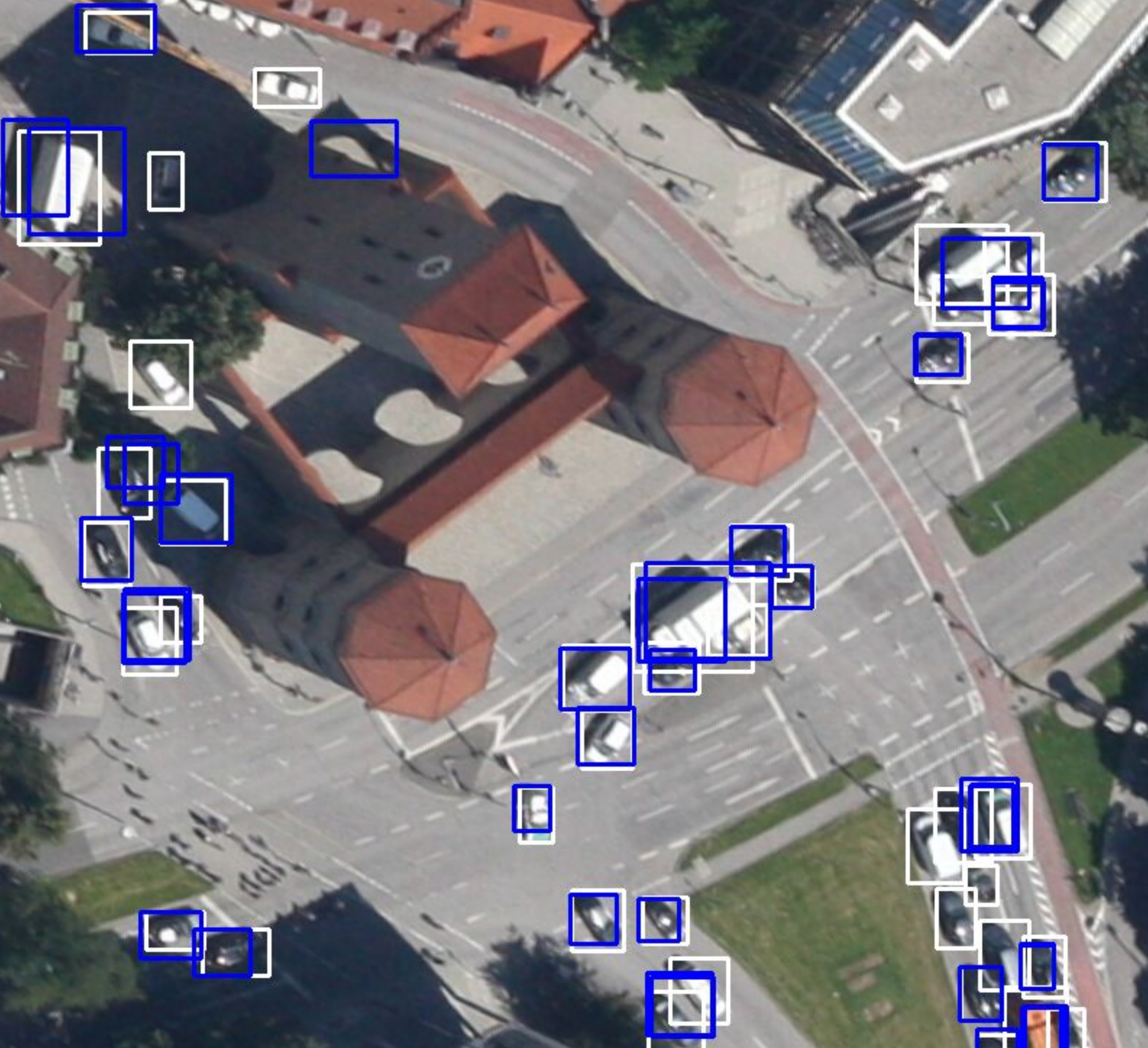}
         \put(3,86){\color{red}8}
       \end{overpic}}

    \caption{Tracking results by AerialMPTNet (top row) and SMSOT-CNN (bottom row)  on the frames 2 and 8 of the ``MunichCrossroad02" sequence of the KIT~AIS vehicle dataset. The predictions and ground truth bounding boxes are depicted in blue and white, respectively.}
    \label{fig:crossroad02dd}%
\end{figure}

According to~\autoref{tab:overallperformance}, AerialMPTNet outperforms SMSOT-CNN also on the KIT~AIS vehicle dataset, although the increase  in performance is lower compared to the pedestrian tracking results. 
Results on different sequences in~\autoref{tab:aerialMPTNetresults} and~\autoref{tab:smsotAll} show that both methods perform poorly on the ``MunichCrossroad02" sequence. \autoref{fig:crossroad2} visualizes the challenges that the tracking methods face in this sequence. For the visualization, we selected an early and a late frame to demonstrate the strong camera movements and changes in the viewing angle, which affect scene arrangements and object appearances. In addition, vehicles are partly or completely occluded by shadows and other objects such as trees. Finally, in this crossroad the movement patterns of the vehicles are complex.
In~\autoref{fig:crossroad02dd}, we compare the performances of AerialMPTNet and SMSOT-CNN on the ``MunichCrossroad02" sequence.  Both methods track AerialMPTNet tracks a few vehicles better than SMSOT-CNN such as the ones located densely at the traffic lights. AerialMPTNet loses track of  a few vehicles which are tracked correctly by SMSOT-CNN. These failures could be solved by a parameter adjustment in our AerialMPTNet.

\begin{figure*}%
    \centering
      \subfloat{\begin{overpic}[width=.49\textwidth]{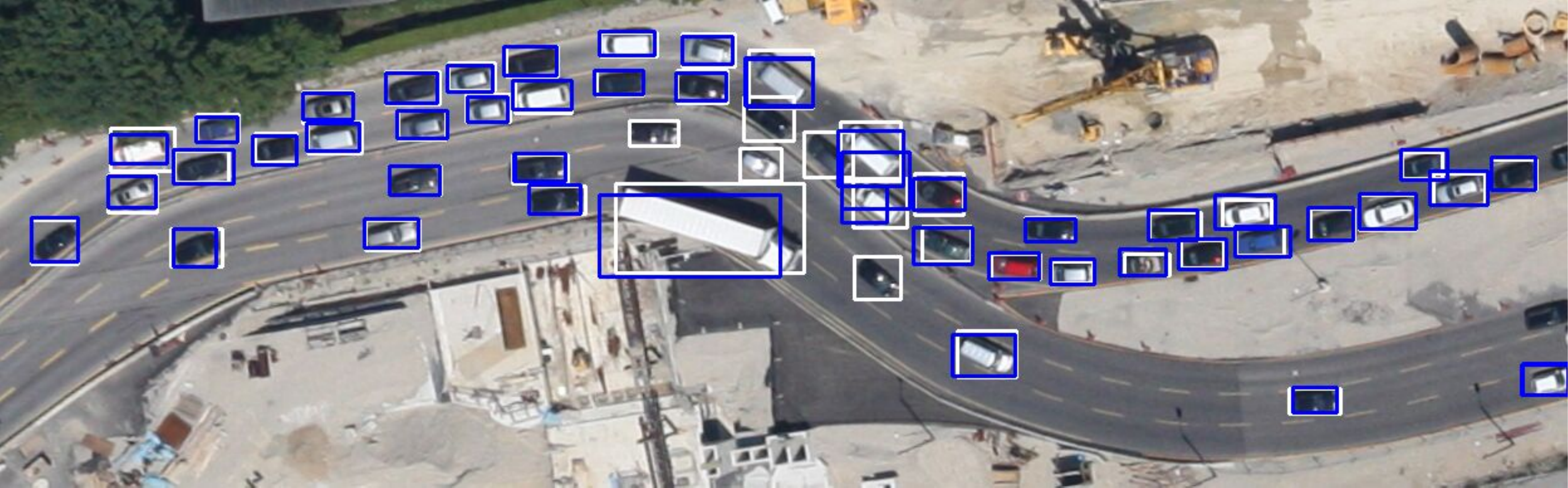}
         \put(2,27){\color{red}20}
       \end{overpic}}
  \subfloat{\begin{overpic}[width=.49\textwidth]{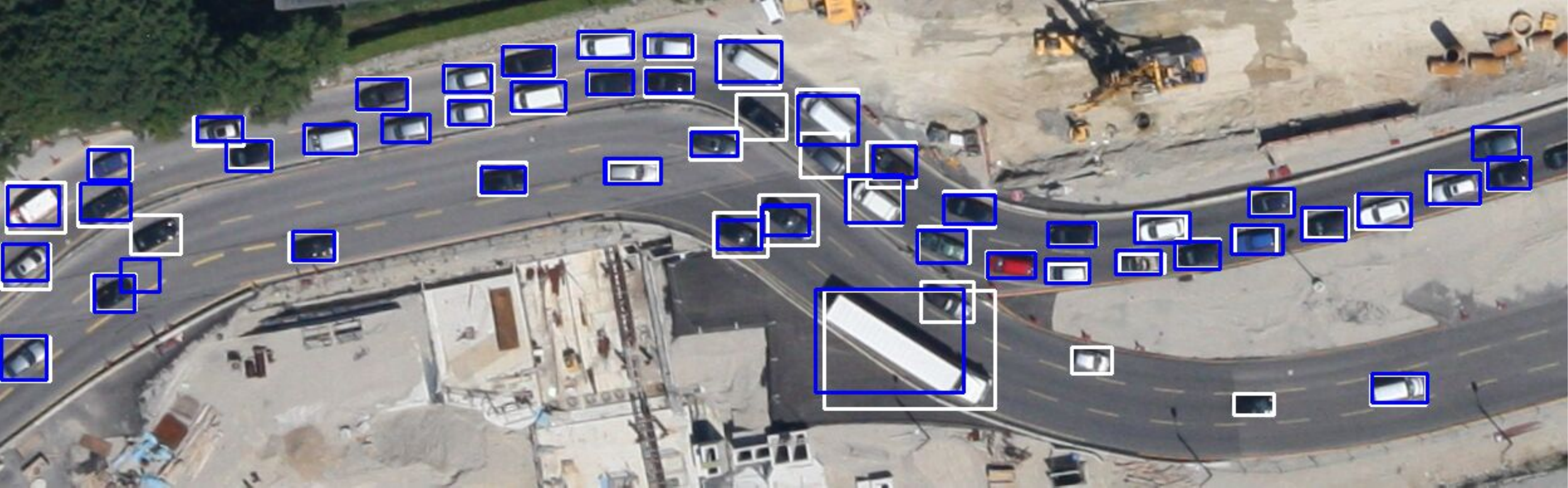}
         \put(2,27){\color{red}29}
       \end{overpic}}
    \\[1ex]       
   \subfloat{\begin{overpic}[width=.49\textwidth]{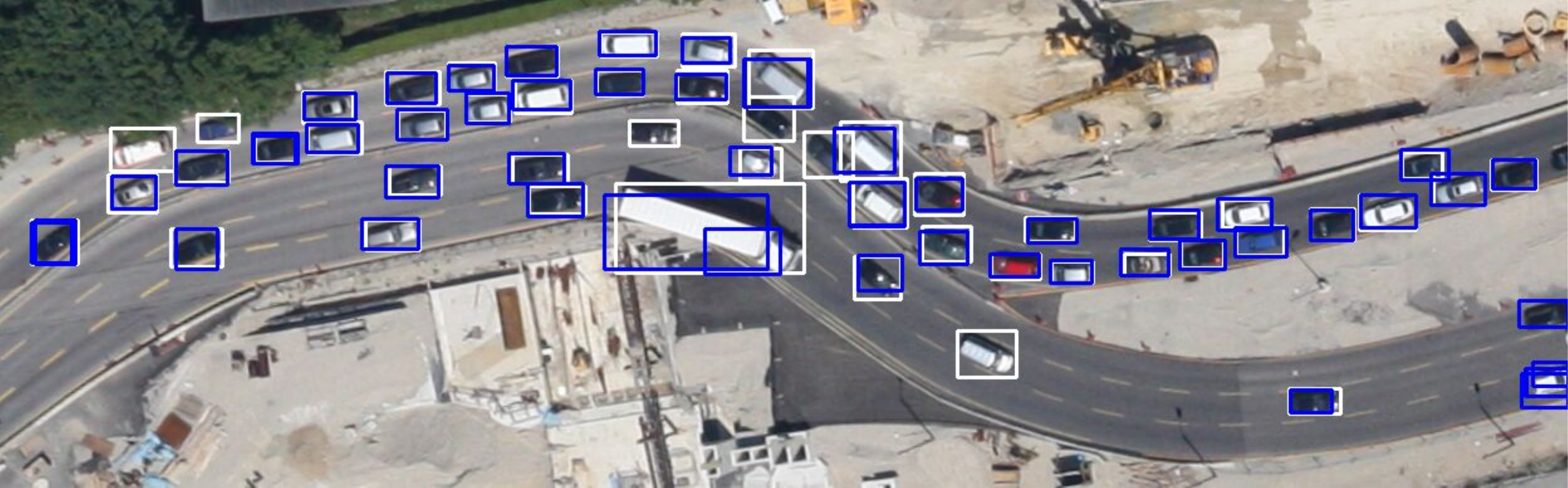}
         \put(2,27){\color{red}20}
       \end{overpic}}
   \subfloat{\begin{overpic}[width=.49\textwidth]{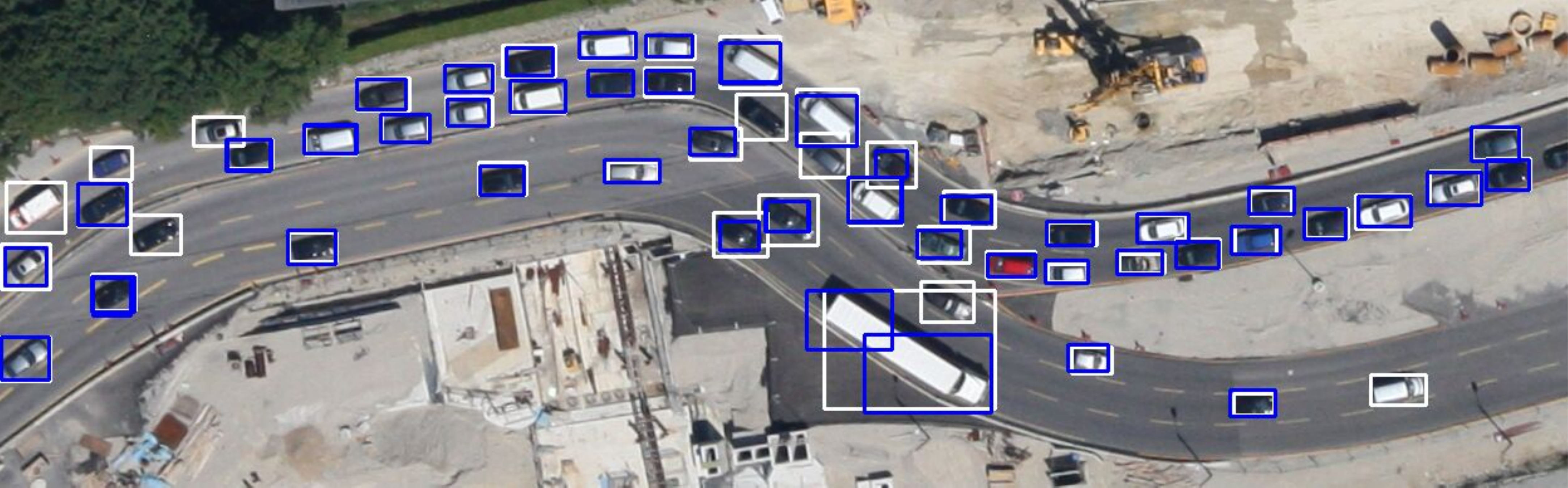}
         \put(2,27){\color{red}29}
       \end{overpic}}
    \caption{Tracking results by AerialMPTNet (top row) and SMSOT-CNN (bottom row)  on the frames 20 and 29 of the ``MunichStreet04" sequence of the KIT~AIS vehicle dataset. The predictions and ground truth bounding boxes are depicted in blue and white, respectively.}
    \label{fig:stree04}%
\end{figure*}

\begin{figure*}%
    \centering
      \subfloat{\begin{overpic}[width=.49\textwidth]{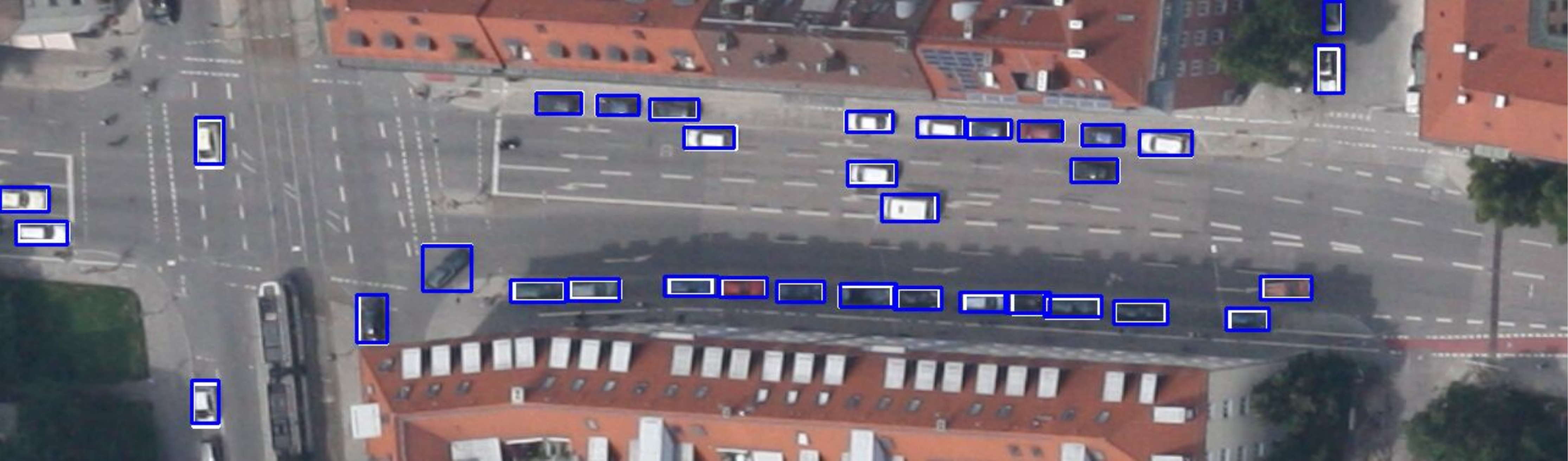}
         \put(2,22){\color{red}1}
       \end{overpic}}
  \subfloat{\begin{overpic}[width=.49\textwidth]{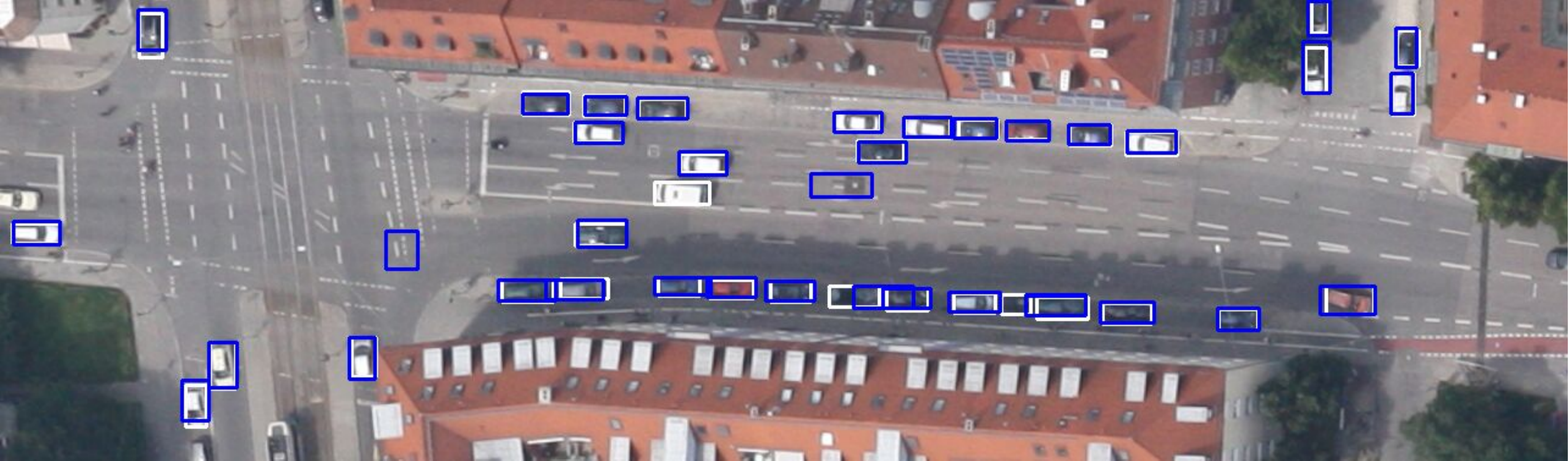}
         \put(2,22){\color{red}7}
       \end{overpic}}
    \\[1ex]       
   \subfloat{\begin{overpic}[width=.49\textwidth]{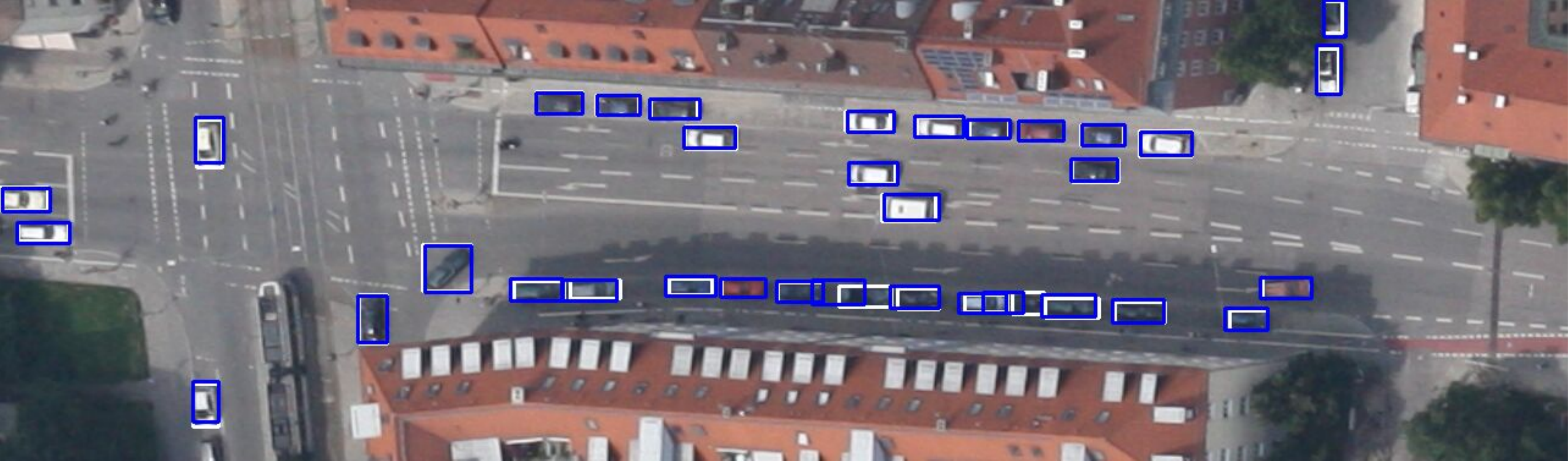}
         \put(2,22){\color{red}1}
       \end{overpic}}
   \subfloat{\begin{overpic}[width=.49\textwidth]{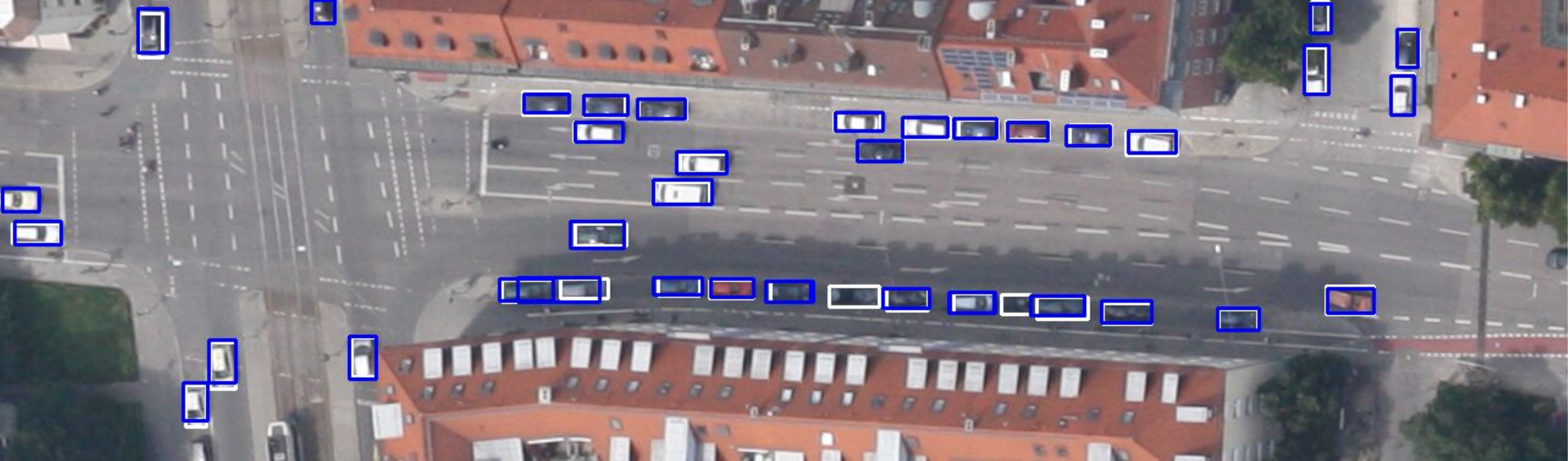}
         \put(2,22){\color{red}7}
       \end{overpic}}
    \caption{Tracking results by AerialMPTNet (top row) and SMSOT-CNN (bottom row)  on the frames 1 and 7 of the ``MunichStreet02" sequence of the KIT~AIS vehicle dataset. The predictions and ground truth bounding boxes are depicted in blue and white, respectively.}
    \label{fig:street02}%
\end{figure*}

In~\autoref{fig:stree04} we compare performances on the ``MunichStreet04" sequence. In this example, AerialMPTNet tracks the long vehicle much better than SMSOT-CNN.
Based on~\autoref{tab:aerialMPTNetresults} and~\autoref{tab:smsotAll}, SMSOT-CNN outperforms our AerialMPTNet on the ``MunichStreet02" sequence. In~\autoref{fig:street02}, we exemplify the existing problems with our AerialMPTNet in this sequence. A background object (in the middle of the scene) has been recognized as a vehicle in frame 7, while the vehicle of interest is lost. A similar failure happens at the intersection. This is due to the parameter configurations of AerialMPTNet. As mentioned before, our method was initially proposed for pedestrian tracking, taking into account the characteristics and challenges of this task. Thus, we believe that by further investigations and parameter tuning, such issues should be solved.

\subsubsection{Localization Preciseness}

\begin{figure*}%
    \centering
    \subfloat[]{\includegraphics[width=.485\textwidth]{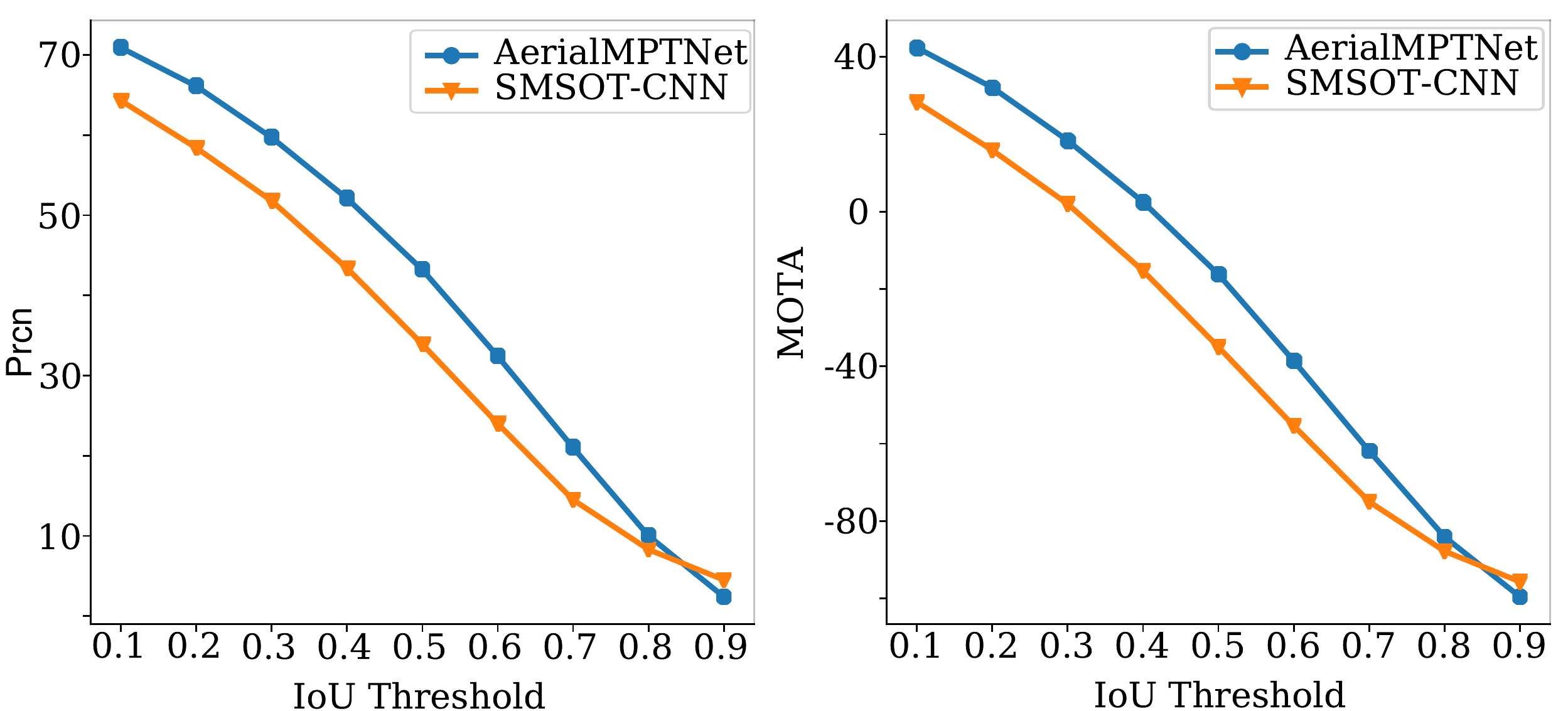}}
    \subfloat[]{\includegraphics[width=.49\textwidth]{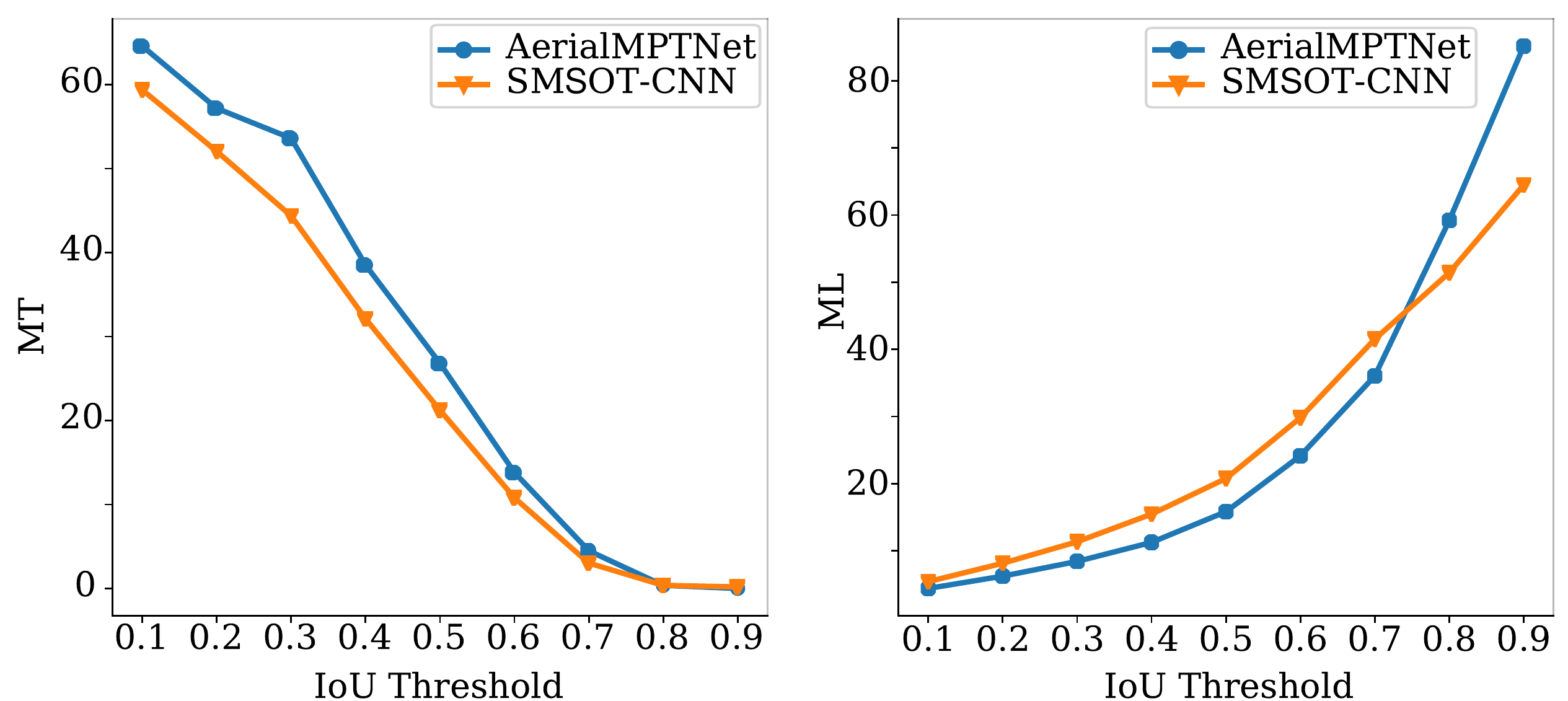}}
    
    \subfloat[]{\includegraphics[width=.485\textwidth]{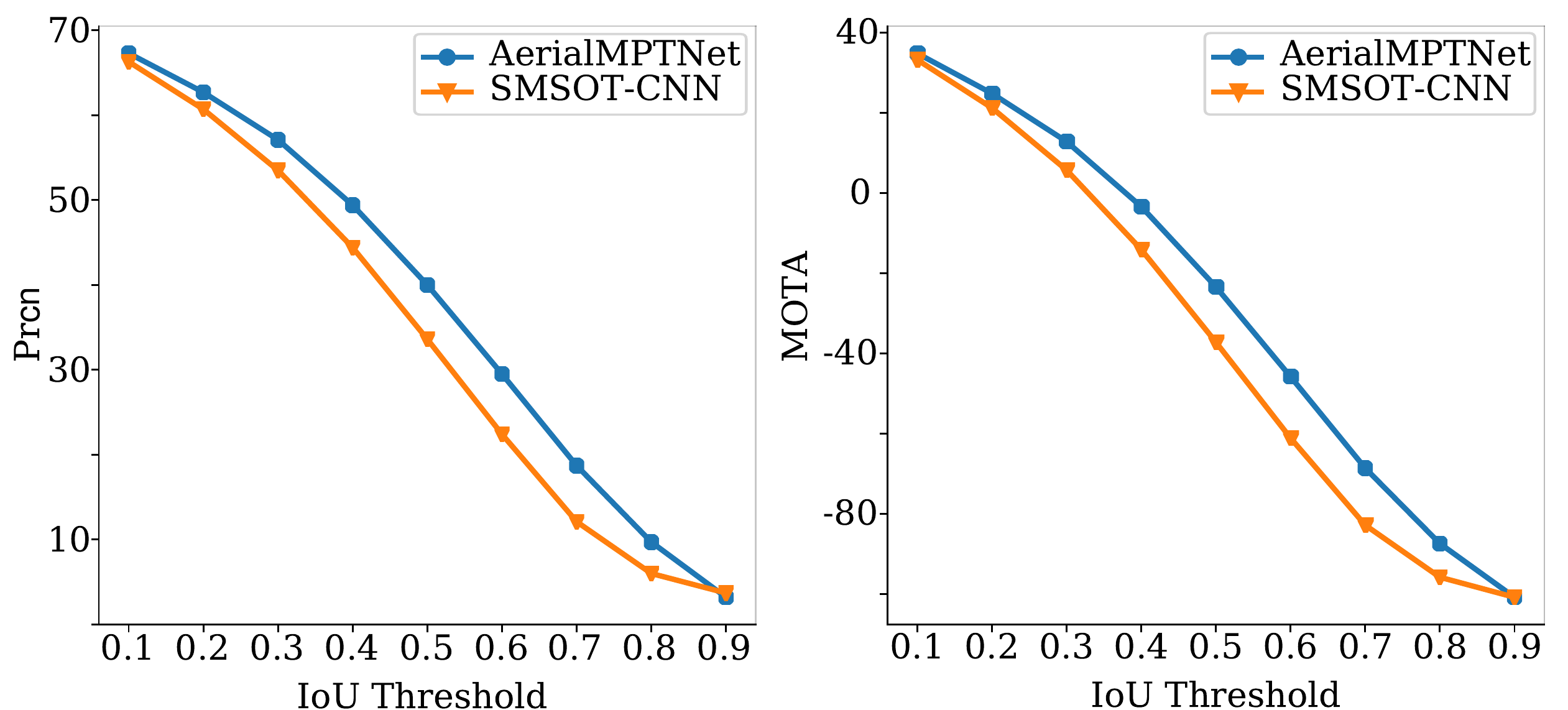}}
    \subfloat[]{\includegraphics[width=.49\textwidth]{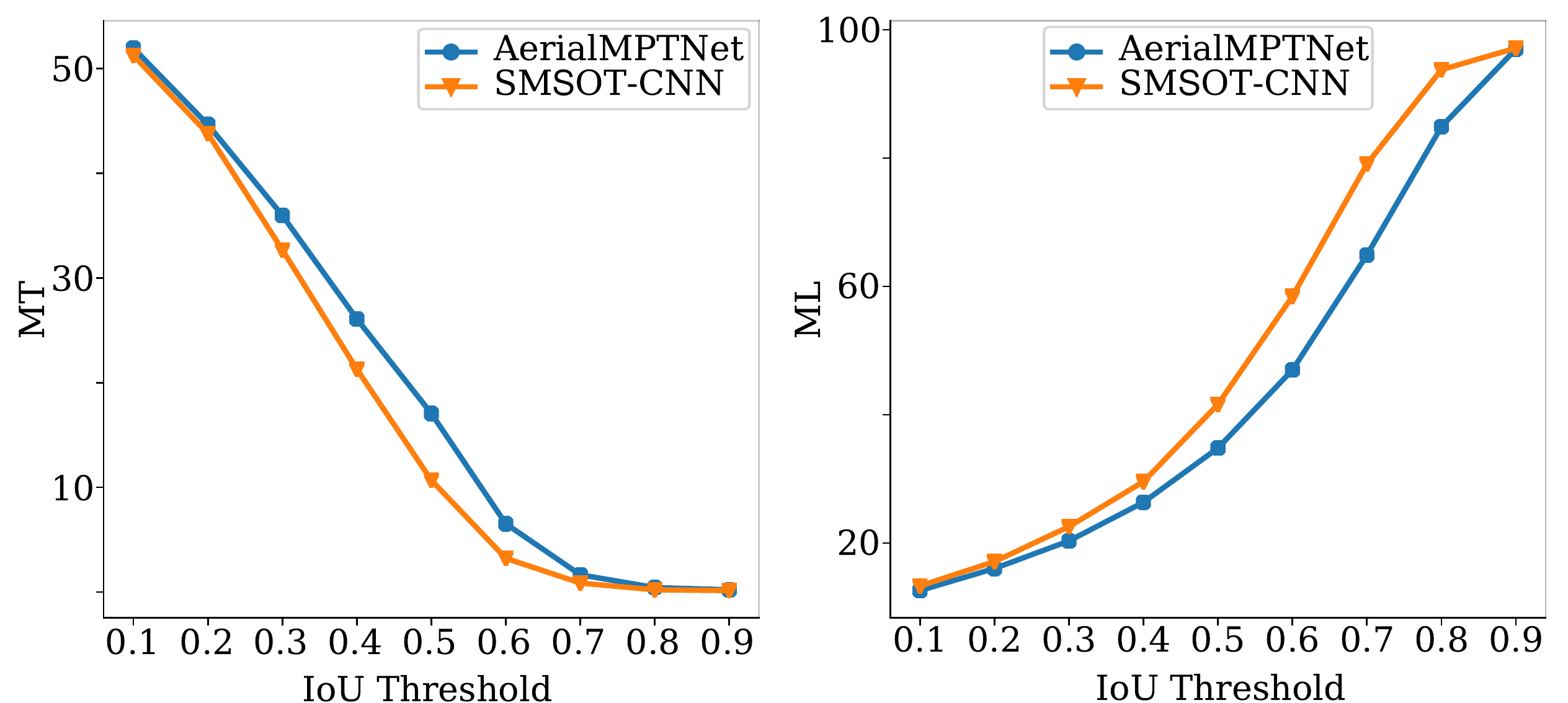}}
    
   \subfloat[]{\includegraphics[width=.485\textwidth]{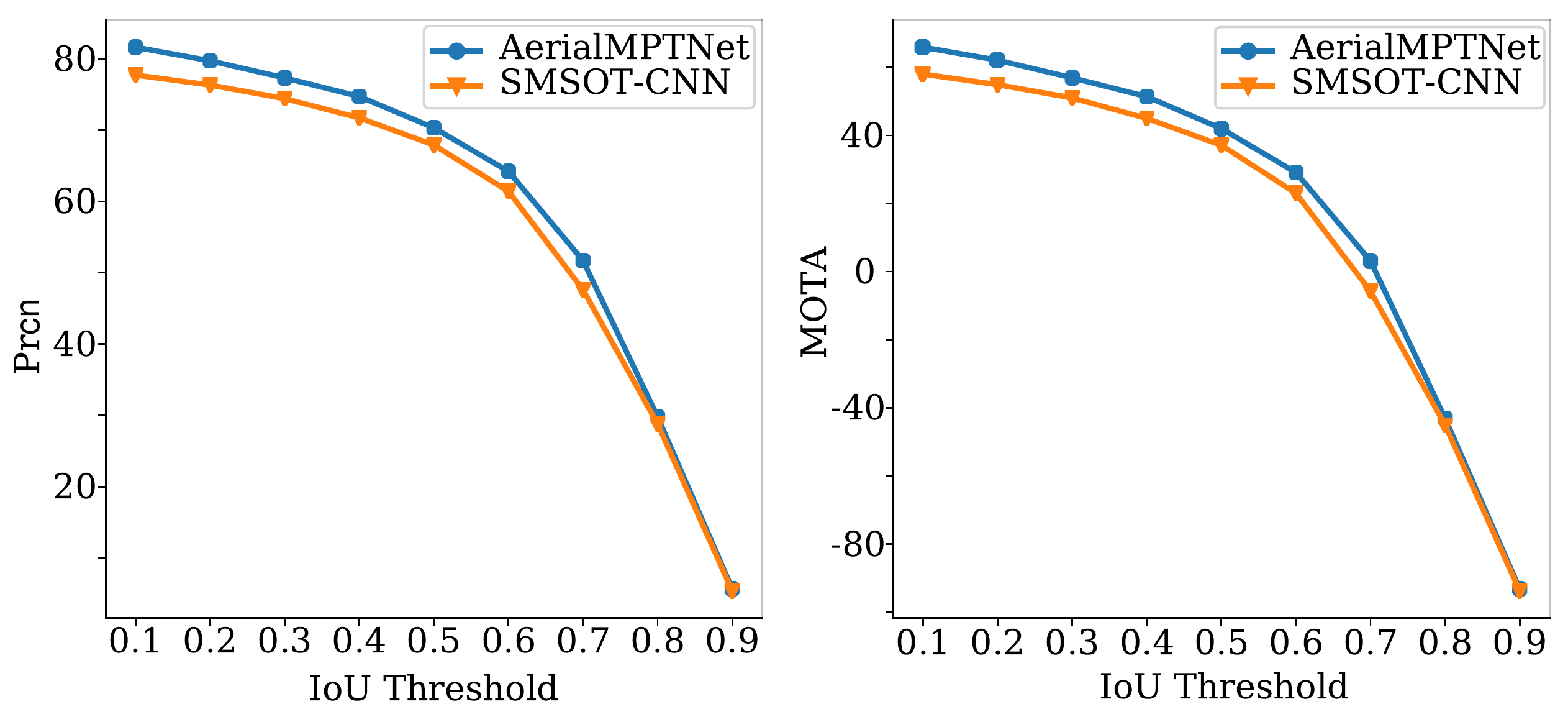}}
    \subfloat[]{\includegraphics[width=.49\textwidth]{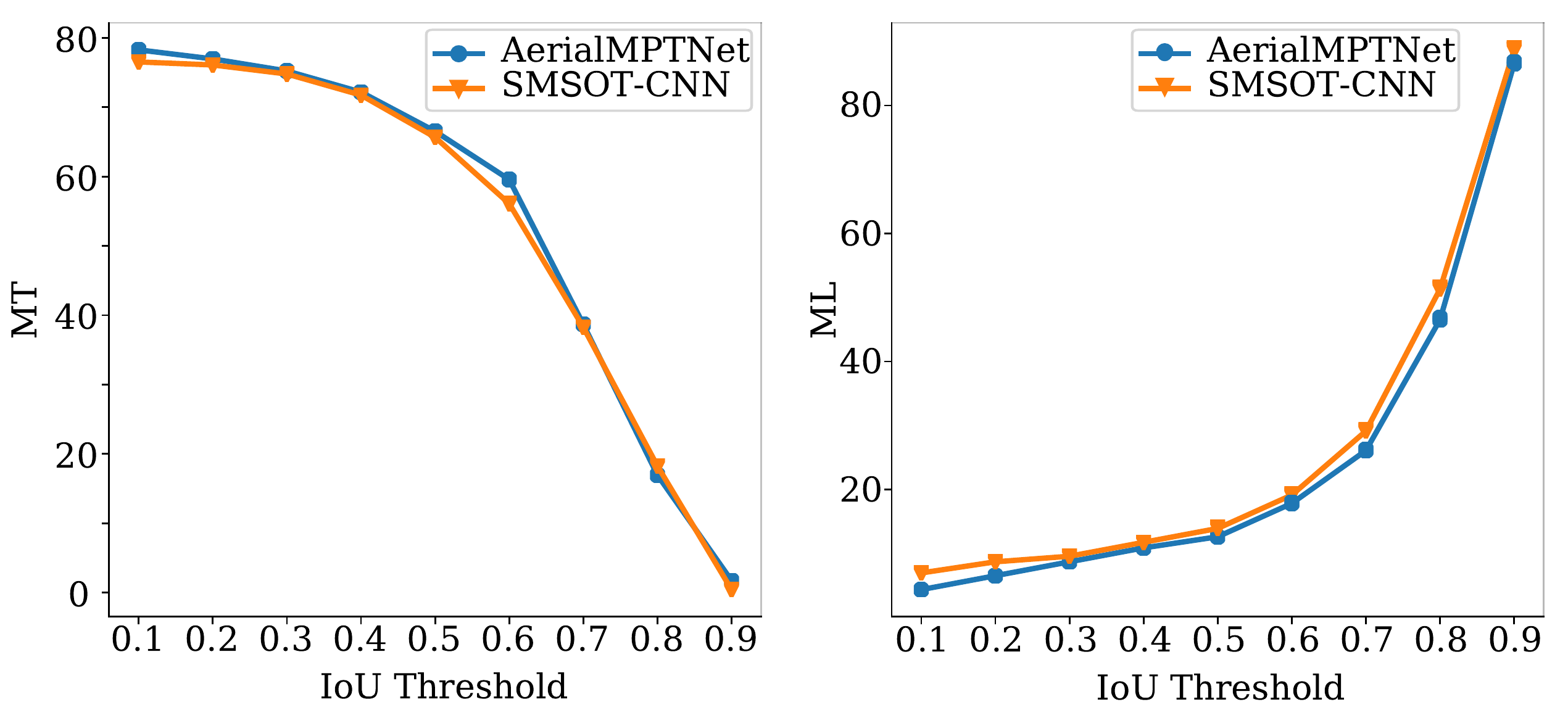}}

    \caption{Comparing the Prcn, MOTA, MT, and ML of the AerialMPTNet and SMSOT-CNN on the KIT~AIS pedestrian (first row), AerialMPT (second row), and KIT~AIS vehicle (third row) datasets by changing the IoU thresholds of the evaluation metrics.}%
\label{fig:MOTAKITPedestrian}
\end{figure*}

In order to evaluate the preciseness of the object locations predicted by AerialMPTNet with respect to SMSOT-CNN, we vary the overlap criterion (IoU threshold) of the evaluation metrics for the Prcn, MOTA, MT, and ML metrics in~\autoref{fig:MOTAKITPedestrian}.
According to the plots, the performance of both methods decreases by increasing the IoU threshold, requiring more overlap between the predicted and ground truth bonding boxes (more precise localization.) For all presented metrics, the preciseness of our ArialMPTNet surpasses that of the SMSOT-CNN.   
However, for the vehicle dataset the performance increase by our AerialMPTNet over SMSOT-CNN is lower than for the case of the pedestrian datasets.

\subsection{AerialMPTNet (with Squeeze-and-Excitation layers)}

In this step we evaluate the improvement achieved by adding SE layers to our AerialMPTNet, as described in Section~\ref{sec:squeeze}. We train the network on our three experimental datasets and report the tracking results in~\autoref{tab:overallperformance}. Using the SE layers in AerialMPTNet$_{SE}$ degrades the results marginally for most of the metrics on the KIT~AIS pedestrian and vehicle datasets as compared to AerialMPTNet. For the vehicle dataset, the SE layers improves the number of the mostly lost (ML) and partially tracked (PT) vehicles by 0.9\% and 3.9\%, respectively.
On the AerialMPT dataset, however, the network behaviour is totally different. AerialMPTNet$_{SE}$ outperforms AerialMPTNet for most of the metrics. SE layers improve MOTA and MOTP by 2 and 0.1 points, respectively. Moreover, the number of mostly tracked (MT) pedestrians increases by 1.7\%.
These inconstant behaviours could be due to the different image quality and contrast of the datasets. Since the images of the AerialMPT dataset are characterized by a higher quality, the adaptive channel weighting would be more meaningful.

\subsection{Training with OHEM}

We evaluate the influence of Online Hard Example Mining (OHEM) on the training of our AerialMPTNet as described in Section~\ref{sec:ohem}. The results are compared to those of the AerialMPTNet with its standard training procedure in~\autoref{tab:overallperformance}.
The use of OHEM in the training procedure reduces the performance marginally on both pedestrian datasets. For example, MOTA decreases by 5 and 1.7 points for the KIT~AIS pedestrian and  AerialMPT datasets, respectively. For the KIT AIS vehicle dataset, however, results show small improvements in the tracking results. For instance, MOTA rises by 1.8 points and the number of mostly tracked objects increases by 1.4\%.
We argue that pedestrian movement is highly complex and therefore, providing in input a similar situation multiple times to the tracker based on OHEM does not help the performance. For the vehicles, however, since they mostly moves in straight paths, OHEM can improve the training by retrying the failure cases. This is the first experiment on the benefits of OHEM in regression-based tracking. Further experiments have to be conducted in order to better understand the underlying reasons.

\subsection{Huber Loss Function}

We assess the effects of loss function in the tracking performance by using the Huber loss~\cite{huber1992robust} instead of the traditional $L1$ loss function. The Huber loss is a mixture of the $L1$ and $L2$ losses, both commonly used for regression problems, and combines their strengths. The $L1$ loss measures the Mean Absolute Error (MAE) between the output of the network \(x\) and the ground truth \(\hat{x}\):
\begin{equation}
    L1(x,\hat{x}) = \sum_i|x_i-\hat{x}_i|
\end{equation}

The L2 loss calculates the Mean Squared Error (MSE) between the network output and the ground truth value:
\begin{equation}
    L2(x,\hat{x}) = \sum_i(x_i-\hat{x}_i)^2
\end{equation}
The $L1$ loss is less affected by outliers with respect to the $L2$ loss. 
The Huber loss acts as a MSE when the error is small, and as a MAE when the error is large:
\begin{equation}
    L_H(x,\hat{x}) =  \sum_i z_i, \\
\end{equation}
\[
    z_i = \begin{cases}
    0.5(x_i-\hat{x}_i)^2, & \text{$if~~|x_i-\hat{x}_i|<1 $}\\
    |x_i-\hat{x}_i|-0.5, & \text{$otherwise$}
    \end{cases}
\]
The Huber loss is more robust to outliers with respect to $L2$ and improves the $L1$ loss for the missing minima at the end of the training.

\autoref{tab:huber} compares results obtained by $L1$ and Huber loss functions. The model trained with the $L1$ loss outperforms the one trained with the Huber loss in general on all three datasets. There are a few metrics for which the Huber loss shows an improvement over $L1$, such as MT in the vehicle dataset or IDS in the AerialMPT dataset; however, these are marginal. Altogether, we can conclude that the $L1$ loss is a better option for our method in these tracking scenarios.

\lossresults

\section{Comparing AerialMPTNet to Other Methods}

In this section, we compare the results of our AerialMPTNet with a set of traditional methods including KCF, Median Flow, CSRT, and MOSSE as well as DL-based methods such as Tracktor++, Stacked-DCFNet, and SMSOT-CNN.
\autoref{tab:overallperformance} reports the results of different tracking methods on the KIT~AIS and AerialMPT datasets. 
In general, the DL-based methods outperform the traditional ones, with MOTA scores varying between -16.2 and -48.8 rather then between -55.9 and -85.8, respectively. The percentages of mostly tracked and mostly lost objects vary between 0.8\% and 9.6\% for the DL-based methods, while they lie between 36.5\% and 78.3\% for the traditional ones. 

\subsection{Pedestrian Tracking}
Among the traditional methods, CSRT is the best performing one on the AerialMPT and KIT~AIS pedestrian datasets, with MOTA values of -55.9 and -64.6. CSRT mostly tracks 9.6\% and 2.9\%, and of the pedestrians while it mostly loses 39.4\% and 59.3\% of the objects in these datasets. 
The DL-based methods, apart from Tracktor++, track much more pedestrians mostly ($>$13.8\%) and lose much less pedestrians ($<$23.6\%) with respect to traditional methods. The poor performances of Tracktor++ is due to its limitations in working with small objects.
AerialMPTNet outperforms all other methods according to most of the adopted figures of merit on the pedestrian datasets with significantly larger MOTA values (-16.2 and -23.4) and competitive MOTP (69.6 and 69.7) values.
It mostly tracks 5.9\% and 4.6\% more pedestrians and loses 5.2\% and 6.8\% less pedestrians with respect to the best performing previous method, SMSOT-CNN on the KIT~AIS and AerialMPT pedestrian datasets, respectively.

\subsection{Vehicle Tracking}

As~\autoref{tab:overallperformance} demonstrates, the DL-based methods and CSRT outperform KCF, Median Flow, and MOSSE significantly, with average MOTA value of 42.9 versus -30.9. The DL-based methods and CSRT are also better with respect to the number of mostly tracked and mostly lost vehicles, varying between 30.0\% and 69.1\% and between 22.6\% and 12.6\%, respectively. These values for KCF, MOSSE, and Median Flow are between 19.6\% and 32.2\% and between 50.4\% and 27.8\%.  
Among the DL-based methods, Stacked-DCFNet has the best performance in terms of MOTA and MOTP, outperforming AerialMPTNet by 4.6 and 5.7 points, respectively. While the number of mostly tracked vehicles by Stacked-DCFNet is 2.6\% larger than in the case of AerialMPTNet, it mostly loses 3.1\% more vehicles. 
The performance of Tracktor++ increases significantly compared to the pedestrian scenarios, due to the ability of its object detector in detecting vehicles. Tracktor++ achieves a competitive MOTA of 37.1 without any ground truth initialization. 
The best performing method in terms of MOTA, MT, and ML is CSRT. It outperforms all other methods with a MOTA of 51.1 and MOTP of 80.7.

\begin{figure*}%
    \centering
    \subfloat[]{\includegraphics[width=.27\textwidth]{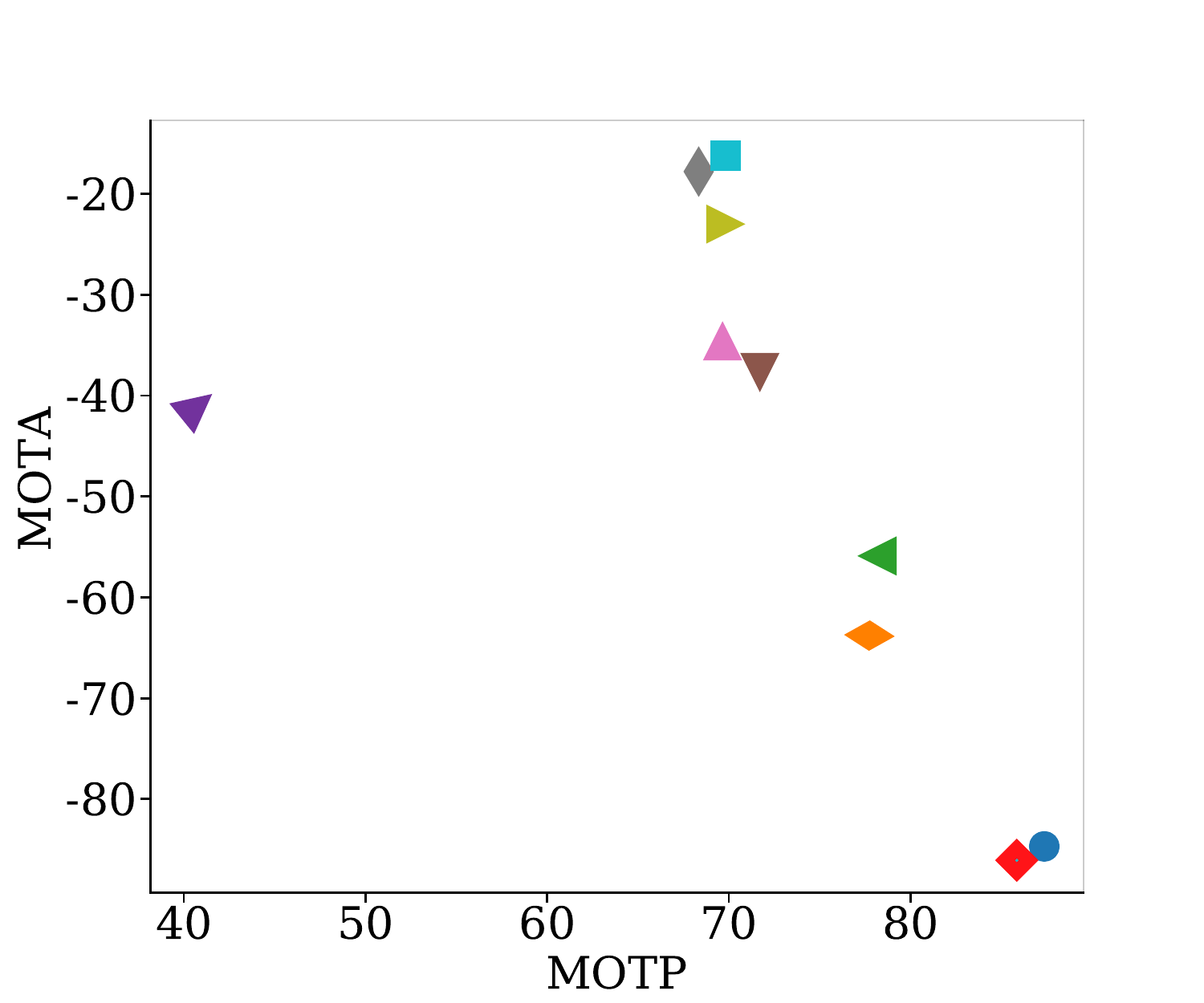}}
    \subfloat[]{\includegraphics[width=.27\textwidth]{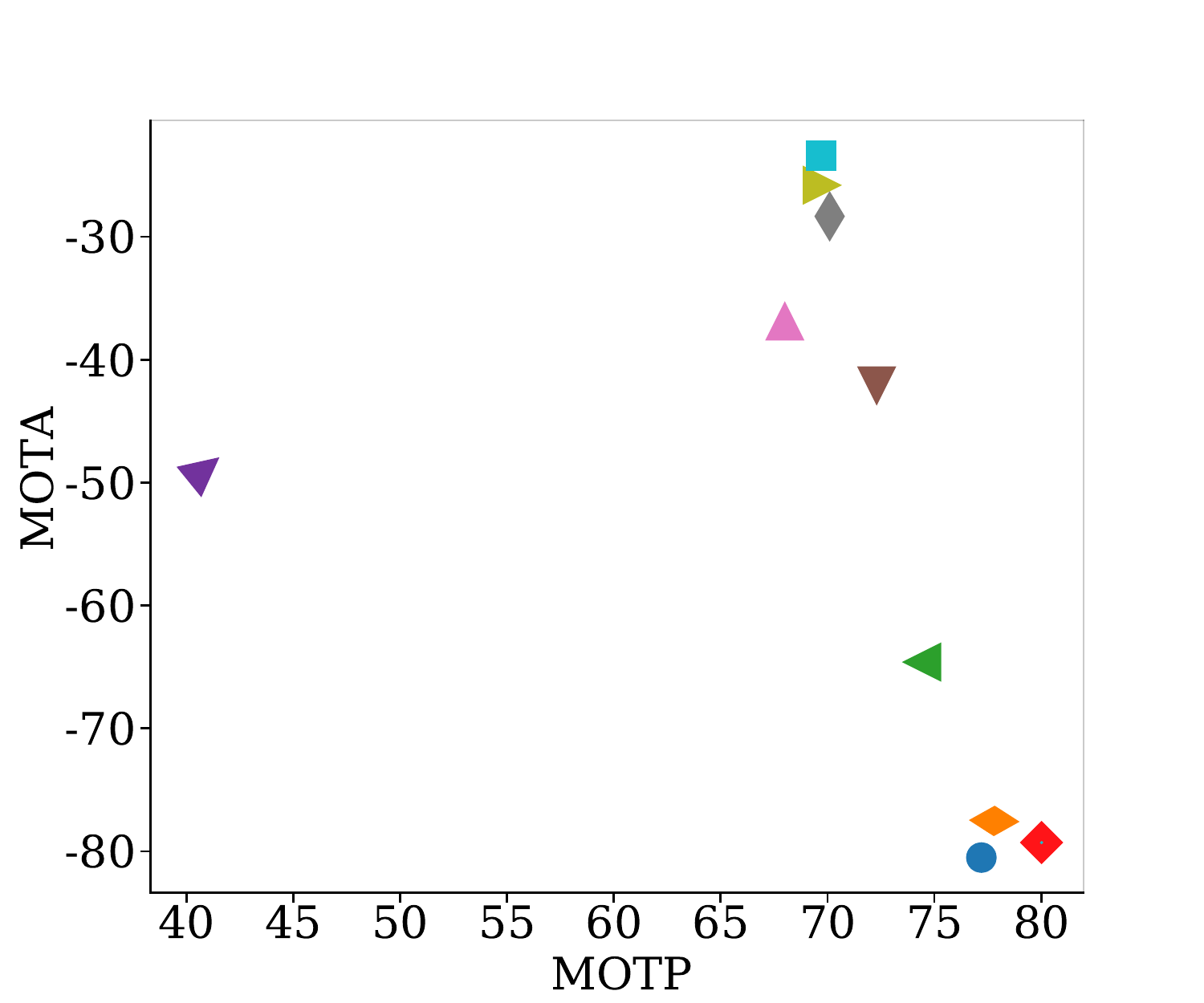}}
    \subfloat[]{\includegraphics[width=.268\textwidth]{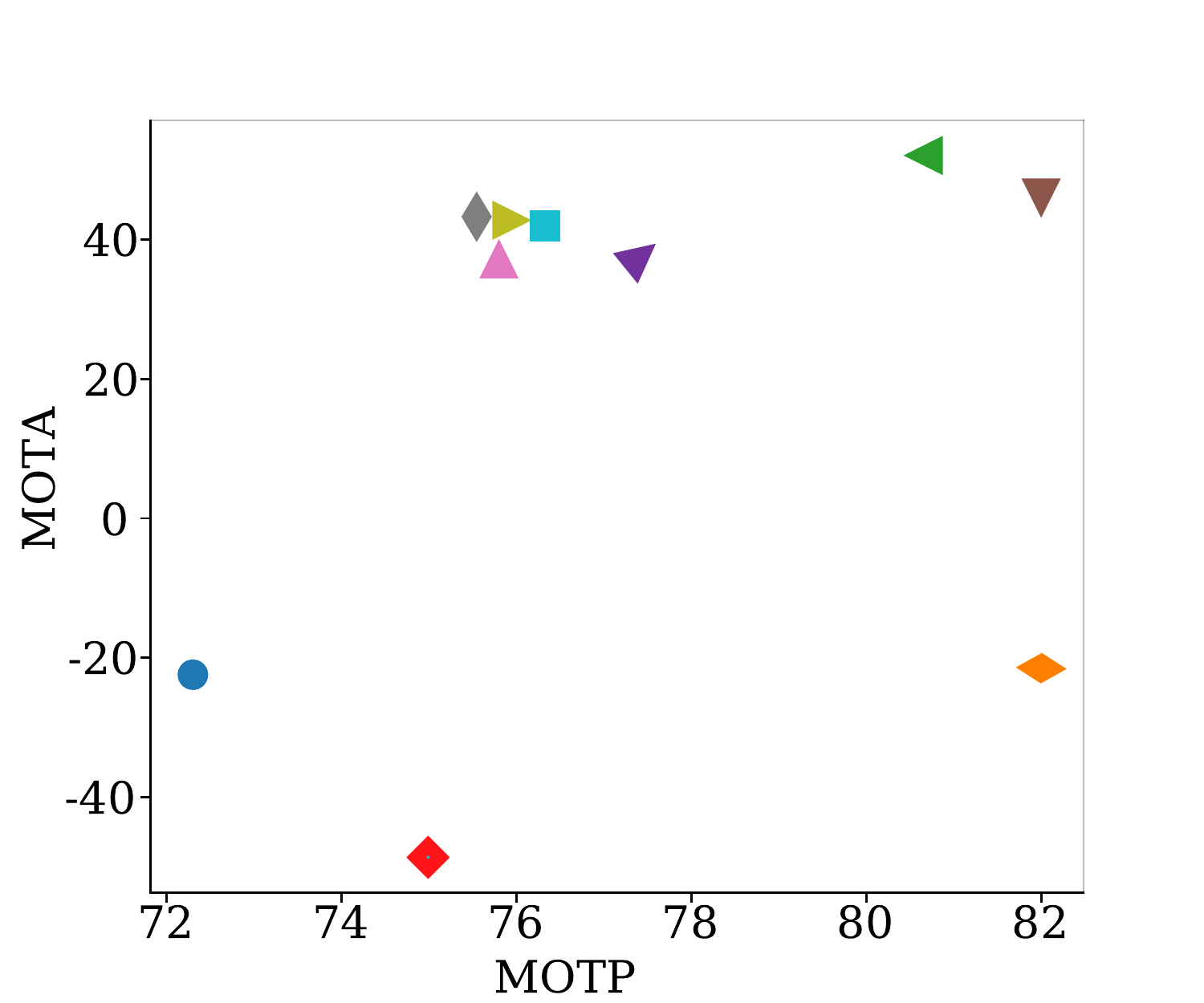}}
    \subfloat[]{\includegraphics[width=.16\textwidth]{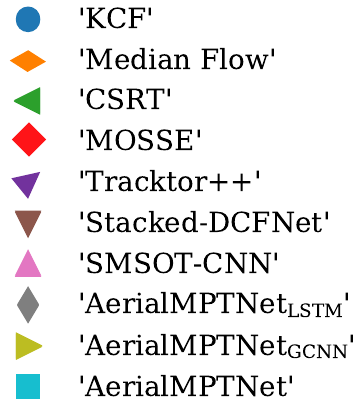}}
    
    \caption{Ranking the tracking methods based on their MOTA and MOTP values on the (a) KIT~AIS pedestrian, (b) AerialMPT, and (c) KIT~AIS vehicle datasets.}
    \label{fig:ranking}
\end{figure*}

We rank the studied tracking methods based on their MOTA and MOTP values in~\autoref{fig:ranking}, with the diagrams offering a clear overview on their performance. AerialMPTNet appears the best method in terms of MOTA for both pedestrian datasets, and achieves competitive MOTP values. Median Flow, for example, achieves a very high MOTP values; however, because of the low number of matched track-object pairs after the first frame, it is not able to track many objects. Hence, the MOTP value solely is not a good performance indicator. For the KIT~AIS vehicle dataset, AerialMPTNet shows worse performance than the other methods according to the MOTA and MOTP values. CSRT and Stacked-DCFNet, however, perform favorably for vehicle tracking.
\section{Conclusion and Future Works}\label{sec:conclusion}

In this paper, we investigate the challenges posed by the tracking of pedestrians and vehicles in aerial imagery by applying a number of traditional and DL-based \gls{sot} and \gls{mot} methods on three aerial \gls{mot} datasets. We also describe our proposed DL-based aerial \gls{mot} method, the so-called AerialMPTNet. Our proposed network fuses appearance, temporal, and graphical information for a more accurate and stable tracking by employing a \gls{snn}, a \gls{lstm}, and a \gls{gcnn} module.
The influence of \gls{se} and \gls{ohem} on the performance of AerialMPTNet is investigated, as well as the impact of adopting an $L1$ rather than a Huber loss function.
%
%
An extensive qualitative and quantitative evaluation shows that the proposed AerialMPTNet outperforms both traditional and state-of-the-art DL-based \gls{mot} methods for the pedestrian datasets, and achieves competitive results for the vehicle dataset. On the one hand, it is verified that \gls{lstm} and \gls{gcnn} modules enhance the tracking performance; on the other hand, the use of \gls{se} and \gls{ohem} significantly helps only in some cases, while degrading the tracking results in other cases. The comparison of $L1$ and Huber loss shows that $L1$ is a better option for most of the scenarios in our experimental datasets.
We believe that the present paper can promote research on aerial \gls{mot} by providing a deep insight into its challenges and opportunities, and pave the path for future works in this domain.
In the future, within the framework of AerialMPTNet, the search area size can be adapted to the image GSDs and object velocities and accelerations. Additionally, the \gls{snn} module can be modified in orer to improve the appearance features extraction.
The training process of most DL-based tracking methods relies on common loss functions, which do not correlate with tracking evaluation metrics such as MOTA and MOTP, as they are usually differentiable. Recently, differentiable proxies of MOTA and MOTP have been proposed~\cite{xutrain}, which can be also investigated for the aerial \gls{mot} scenarios.


\section*{Acknowledgment}

The authors would like to thank...

\ifCLASSOPTIONcaptionsoff
  \newpage
\fi



\printbibliography

@article{kim2018residual,
  title={Residual LSTM attention network for object tracking},
  author={Kim, Hong-In and Park, Rae-Hong},
  journal={Ieee Signal Processing Letters},
  volume={25},
  number={7},
  pages={1029--1033},
  year={2018},
  publisher={IEEE}
}

@article{held_learning_2016,
	title = {Learning to {Track} at 100 {FPS} with {Deep} {Regression} {Networks}},
	url = {http://arxiv.org/abs/1604.01802},
	language = {en},
	urldate = {2019-11-04},
	journal = {arXiv:1604.01802 [cs]},
	author = {Held, David and Thrun, Sebastian and Savarese, Silvio},
	month = 08,
	year = {2016},
	note = {arXiv: 1604.01802},
}

@article{bahmanyar2019multiple,
  title={MULTIPLE VEHICLES AND PEOPLE TRACKING IN AERIAL IMAGERY USING STACK OF MICRO SINGLE-OBJECT-TRACKING CNNS},
  author={Bahmanyar, R and Azimi, SM and Reinartz, P},
  journal={The International Archives of Photogrammetry, Remote Sensing and Spatial Information Sciences},
  volume={42},
  pages={163--170},
  year={2019},
  publisher={Copernicus GmbH}
}

@inproceedings{lin2017focal,
  title={Focal loss for dense object detection},
  author={Lin, Tsung-Yi and Goyal, Priya and Girshick, Ross and He, Kaiming and Doll{\'a}r, Piotr},
  booktitle={Proceedings of the IEEE international conference on computer vision},
  pages={2980--2988},
  year={2017}
}

@inproceedings{bahmanyar2019mrcnet,
  title={MRCNet: Crowd Counting and Density Map Estimation in Aerial and Ground Imagery},
  author={Bahmanyar, Reza and Vig, Elenora and Reinartz, Peter},
  booktitle={Proceedings of BMVC’s Workshop on Object Detection and Recognition for Security Screenin (BMVC-ODRSS)},
  year={2019}
}

@inproceedings{ren2015faster,
  title={Faster r-cnn: Towards real-time object detection with region proposal networks},
  author={Ren, Shaoqing and He, Kaiming and Girshick, Ross and Sun, Jian},
  booktitle={Advances in neural information processing systems},
  pages={91--99},
  year={2015}
}

@inproceedings{reilly2010detection,
  title={Detection and tracking of large number of targets in wide area surveillance},
  author={Reilly, Vladimir and Idrees, Haroon and Shah, Mubarak},
  booktitle={European conference on computer vision},
  pages={186--199},
  year={2010},
  organization={Springer}
}

@inproceedings{bewley_simple_2016,
	title = {Simple online and realtime tracking},
	doi = {10.1109/ICIP.2016.7533003},
	booktitle = {2016 {IEEE} {International} {Conference} on {Image} {Processing} ({ICIP})},
	author = {Bewley, Alex and Ge, Zongyuan and Ott, Lionel and Ramos, Fabio and Upcroft, Ben},
	month = sep,
	year = {2016},
	note = {ISSN: 2381-8549},
	pages = {3464--3468}
}

@inproceedings{wojke_simple_2017,
	title = {Simple online and realtime tracking with a deep association metric},
	doi = {10.1109/ICIP.2017.8296962},
	booktitle = {2017 {IEEE} {International} {Conference} on {Image} {Processing} ({ICIP})},
	author = {Wojke, Nicolai and Bewley, Alex and Paulus, Dietrich},
	month = sep,
	year = {2017},
	note = {ISSN: 2381-8549},
	pages = {3645--3649}
}

@article{jadhav_aerial_2019,
	title = {Aerial multi-object tracking by detection using deep association networks},
	url = {http://arxiv.org/abs/1909.01547},
	language = {en},
	urldate = {2019-11-05},
	journal = {arXiv:1909.01547 [cs]},
	author = {Jadhav, Ajit and Mukherjee, Prerana and Kaushik, Vinay and Lall, Brejesh},
	month = sep,
	year = {2019},
	note = {arXiv: 1909.01547}
}

@article{lu_deep_nodate,
	title = {Deep {Regression} {Tracking} with {Shrinkage} {Loss}},
	language = {en},
	author = {Lu, Xiankai and Ma, Chao and Ni, Bingbing and Yang, Xiaokang and Reid, Ian and Yang, Ming-Hsuan},
	pages = {17}
}

@article{wang_dcfnet_2017,
	title = {{DCFNet}: {Discriminant} {Correlation} {Filters} {Network} for {Visual} {Tracking}},
	shorttitle = {{DCFNet}},
	url = {http://arxiv.org/abs/1704.04057},
	language = {en},
	urldate = {2019-11-07},
	journal = {arXiv:1704.04057 [cs]},
	author = {Wang, Qiang and Gao, Jin and Xing, Junliang and Zhang, Mengdan and Hu, Weiming},
	month = apr,
	year = {2017},
	note = {arXiv: 1704.04057}
}

@inproceedings{kalal2010forward,
  title={Forward-backward error: Automatic detection of tracking failures},
  author={Kalal, Zdenek and Mikolajczyk, Krystian and Matas, Jiri},
  booktitle={2010 20th International Conference on Pattern Recognition},
  pages={2756--2759},
  year={2010},
  organization={IEEE}
}

@article{henriques2014high,
  title={High-speed tracking with kernelized correlation filters},
  author={Henriques, Jo{\~a}o F and Caseiro, Rui and Martins, Pedro and Batista, Jorge},
  journal={IEEE transactions on pattern analysis and machine intelligence},
  volume={37},
  number={3},
  pages={583--596},
  year={2014},
  publisher={IEEE}
}

@article{hare2015struck,
  title={Struck: Structured output tracking with kernels},
  author={Hare, Sam and Golodetz, Stuart and Saffari, Amir and Vineet, Vibhav and Cheng, Ming-Ming and Hicks, Stephen L and Torr, Philip HS},
  journal={IEEE transactions on pattern analysis and machine intelligence},
  volume={38},
  number={10},
  pages={2096--2109},
  year={2015},
  publisher={IEEE}
}

@article{cuevas2005kalman,
  title={Kalman filter for vision tracking},
  author={Cuevas, Erik V and Zaldivar, Daniel and Rojas, Raul},
  year={2005}
}

@article{cuevas2007particle,
  title={Particle filter in vision tracking},
  author={Cuevas, Erik and Zaldivar, Daniel and Rojas, Raul},
  journal={e-Gnosis},
  number={5},
  pages={1--11},
  year={2007},
  publisher={Universidad de Guadalajara}
}

@inproceedings{okuma2004boosted,
  title={A boosted particle filter: Multitarget detection and tracking},
  author={Okuma, Kenji and Taleghani, Ali and De Freitas, Nando and Little, James J and Lowe, David G},
  booktitle={European conference on computer vision},
  pages={28--39},
  year={2004},
  organization={Springer}
}

@inproceedings{butenuth2011integrating,
  title={Integrating pedestrian simulation, tracking and event detection for crowd analysis},
  author={Butenuth, Matthias and Burkert, Florian and Schmidt, Florian and Hinz, Stefan and Hartmann, Dirk and Kneidl, Angelika and Borrmann, Andr{\'e} and Sirmacek, Beril},
  booktitle={2011 IEEE International Conference on Computer Vision Workshops (ICCV Workshops)},
  pages={150--157},
  year={2011},
  organization={IEEE}
}

@inproceedings{huang2008robust,
  title={Robust object tracking by hierarchical association of detection responses},
  author={Huang, Chang and Wu, Bo and Nevatia, Ramakant},
  booktitle={European Conference on Computer Vision},
  pages={788--801},
  year={2008},
  organization={Springer}
}

@techreport{remoteSensing2008,
    title={Remote Sensing Data: Applications and Benefits},
    author={U.S. House Hearing, 110 Congress},
    institution= {Subcommittee on Space and Aeronautics, Committee on Science and Technology},
    year= {2008},
    month = {April},
    note= {Serial No. 110-91, retrieved January 2, 2020: https://www.govinfo.gov/content/pkg/CHRG-110hhrg41573/html/CHRG-110hhrg41573.html}
}

@article{meng2012object,
  title={Object tracking using high resolution satellite imagery},
  author={Meng, Lingfei and Kerekes, John P},
  journal={IEEE Journal of Selected Topics in Applied Earth Observations and Remote Sensing},
  volume={5},
  number={1},
  pages={146--152},
  year={2012},
  publisher={IEEE}
}

@inproceedings{hager1996real,
  title={Real-time tracking of image regions with changes in geometry and illumination},
  author={Hager, Gregory D and Belhumeur, Peter N},
  booktitle={Proceedings CVPR IEEE Computer Society Conference on Computer Vision and Pattern Recognition},
  pages={403--410},
  year={1996},
  organization={IEEE}
}

@inproceedings{briechle2001template,
  title={Template matching using fast normalized cross correlation},
  author={Briechle, Kai and Hanebeck, Uwe D},
  booktitle={Optical Pattern Recognition XII},
  volume={4387},
  pages={95--102},
  year={2001},
  organization={International Society for Optics and Photonics}
}

@article{avidan2007ensemble,
  title={Ensemble tracking},
  author={Avidan, Shai},
  journal={IEEE transactions on pattern analysis and machine intelligence},
  volume={29},
  number={2},
  pages={261--271},
  year={2007},
  publisher={IEEE}
}

@inproceedings{zheng2016mars,
  title={Mars: A video benchmark for large-scale person re-identification},
  author={Zheng, Liang and Bie, Zhi and Sun, Yifan and Wang, Jingdong and Su, Chi and Wang, Shengjin and Tian, Qi},
  booktitle={European Conference on Computer Vision},
  pages={868--884},
  year={2016},
  organization={Springer}
}

@inproceedings{bergmann2019tracking,
  title={Tracking without bells and whistles},
  author={Bergmann, Philipp and Meinhardt, Tim and Leal-Taixe, Laura},
  booktitle={Proceedings of the IEEE International Conference on Computer Vision},
  pages={941--951},
  year={2019}
}

@inproceedings{alahi2016social,
  title={Social lstm: Human trajectory prediction in crowded spaces},
  author={Alahi, Alexandre and Goel, Kratarth and Ramanathan, Vignesh and Robicquet, Alexandre and Fei-Fei, Li and Savarese, Silvio},
  booktitle={Proceedings of the IEEE conference on computer vision and pattern recognition},
  pages={961--971},
  year={2016}
}

@inproceedings{xue2018ss,
  title={SS-LSTM: A hierarchical LSTM model for pedestrian trajectory prediction},
  author={Xue, Hao and Huynh, Du Q and Reynolds, Mark},
  booktitle={2018 IEEE Winter Conference on Applications of Computer Vision (WACV)},
  pages={1186--1194},
  year={2018},
  organization={IEEE}
}

@inproceedings{vemula2018social,
  title={Social attention: Modeling attention in human crowds},
  author={Vemula, Anirudh and Muelling, Katharina and Oh, Jean},
  booktitle={2018 IEEE international Conference on Robotics and Automation (ICRA)},
  pages={1--7},
  year={2018},
  organization={IEEE}
}

@inproceedings{bertinetto2016fully,
  title={Fully-convolutional siamese networks for object tracking},
  author={Bertinetto, Luca and Valmadre, Jack and Henriques, Joao F and Vedaldi, Andrea and Torr, Philip HS},
  booktitle={European conference on computer vision},
  pages={850--865},
  year={2016},
  organization={Springer}
}

@article{marvasti2019deep,
  title={Deep Learning for Visual Tracking: A Comprehensive Survey},
  author={Marvasti-Zadeh, Seyed Mojtaba and Cheng, Li and Ghanei-Yakhdan, Hossein and Kasaei, Shohreh},
  journal={arXiv preprint arXiv:1912.00535},
  year={2019}
}

@inproceedings{wojke2017simple,
  title={Simple online and realtime tracking with a deep association metric},
  author={Wojke, Nicolai and Bewley, Alex and Paulus, Dietrich},
  booktitle={2017 IEEE international conference on image processing (ICIP)},
  pages={3645--3649},
  year={2017},
  organization={IEEE}
}

@inproceedings{xiang2015learning,
  title={Learning to track: Online multi-object tracking by decision making},
  author={Xiang, Yu and Alahi, Alexandre and Savarese, Silvio},
  booktitle={Proceedings of the IEEE international conference on computer vision},
  pages={4705--4713},
  year={2015}
}

@article{everaerts2008use,
  title={The use of unmanned aerial vehicles (UAVs) for remote sensing and mapping},
  author={Everaerts, Jurgen and others},
  journal={The International Archives of the Photogrammetry, Remote Sensing and Spatial Information Sciences},
  volume={37},
  number={2008},
  pages={1187--1192},
  year={2008},
  publisher={International Soc. for Photogrammetry and Remote Sensing (ISPRS)}
}

@article{liu2015fast,
  title={Fast multiclass vehicle detection on aerial images},
  author={Liu, Kang and Mattyus, Gellert},
  journal={IEEE Geoscience and Remote Sensing Letters},
  volume={12},
  number={9},
  pages={1938--1942},
  year={2015},
  publisher={IEEE}
}

@inproceedings{wang2015visual,
  title={Visual tracking with fully convolutional networks},
  author={Wang, Lijun and Ouyang, Wanli and Wang, Xiaogang and Lu, Huchuan},
  booktitle={Proceedings of the IEEE international conference on computer vision},
  pages={3119--3127},
  year={2015}
}

@article{qi2015unsupervised,
  title={Unsupervised ship detection based on saliency and S-HOG descriptor from optical satellite images},
  author={Qi, Shengxiang and Ma, Jie and Lin, Jin and Li, Yansheng and Tian, Jinwen},
  journal={IEEE Geoscience and Remote Sensing Letters},
  volume={12},
  number={7},
  pages={1451--1455},
  year={2015},
  publisher={IEEE}
}

@article{benedek2009detection,
  title={Detection of object motion regions in aerial image pairs with a multilayer Markovian model},
  author={Benedek, Csaba and Szir{\'a}nyi, Tam{\'a}s and Kato, Zoltan and Zerubia, Josiane},
  journal={IEEE Transactions on Image Processing},
  volume={18},
  number={10},
  pages={2303--2315},
  year={2009},
  publisher={IEEE}
}

@inproceedings{bolme2010visual,
  title={Visual object tracking using adaptive correlation filters},
  author={Bolme, David S and Beveridge, J Ross and Draper, Bruce A and Lui, Yui Man},
  booktitle={2010 IEEE computer society conference on computer vision and pattern recognition},
  pages={2544--2550},
  year={2010},
  organization={IEEE}
}

@article{zhang2017deep,
  title={Deep reinforcement learning for visual object tracking in videos},
  author={Zhang, Da and Maei, Hamid and Wang, Xin and Wang, Yuan-Fang},
  journal={arXiv preprint arXiv:1701.08936},
  year={2017}
}

@inproceedings{song2018vital,
  title={Vital: Visual tracking via adversarial learning},
  author={Song, Yibing and Ma, Chao and Wu, Xiaohe and Gong, Lijun and Bao, Linchao and Zuo, Wangmeng and Shen, Chunhua and Lau, Rynson WH and Yang, Ming-Hsuan},
  booktitle={Proceedings of the IEEE Conference on Computer Vision and Pattern Recognition},
  pages={8990--8999},
  year={2018}
}

@inproceedings{he2016deep,
  title={Deep residual learning for image recognition},
  author={He, Kaiming and Zhang, Xiangyu and Ren, Shaoqing and Sun, Jian},
  booktitle={Proceedings of the IEEE conference on computer vision and pattern recognition},
  pages={770--778},
  year={2016}
}

@inproceedings{szegedy2016rethinking,
  title={Rethinking the inception architecture for computer vision},
  author={Szegedy, Christian and Vanhoucke, Vincent and Ioffe, Sergey and Shlens, Jon and Wojna, Zbigniew},
  booktitle={Proceedings of the IEEE conference on computer vision and pattern recognition},
  pages={2818--2826},
  year={2016}
}

@inproceedings{ristani2016performance,
  title={Performance measures and a data set for multi-target, multi-camera tracking},
  author={Ristani, Ergys and Solera, Francesco and Zou, Roger and Cucchiara, Rita and Tomasi, Carlo},
  booktitle={European Conference on Computer Vision},
  pages={17--35},
  year={2016},
  organization={Springer}
}

@inproceedings{held2016learning,
  title={Learning to track at 100 fps with deep regression networks},
  author={Held, David and Thrun, Sebastian and Savarese, Silvio},
  booktitle={European Conference on Computer Vision},
  pages={749--765},
  year={2016},
  organization={Springer}
}

@inproceedings{li2018high,
  title={High performance visual tracking with siamese region proposal network},
  author={Li, Bo and Yan, Junjie and Wu, Wei and Zhu, Zheng and Hu, Xiaolin},
  booktitle={Proceedings of the IEEE Conference on Computer Vision and Pattern Recognition},
  pages={8971--8980},
  year={2018}
}

@inproceedings{boudoukh2009visual,
  title={Visual tracking of object silhouettes},
  author={Boudoukh, Guy and Leichter, Ido and Rivlin, Ehud},
  booktitle={2009 16th IEEE International Conference on Image Processing (ICIP)},
  pages={3625--3628},
  year={2009},
  organization={IEEE}
}

@article{zhang2016robust,
  title={Robust visual tracking via convolutional networks without training},
  author={Zhang, Kaihua and Liu, Qingshan and Wu, Yi and Yang, Ming-Hsuan},
  journal={IEEE Transactions on Image Processing},
  volume={25},
  number={4},
  pages={1779--1792},
  year={2016},
  publisher={IEEE}
}

@INPROCEEDINGS{Schmidt2011,
  author = {Florian Schmidt and Stefan Hinz},
  title = {A Scheme for the Detection and Tracking of People Tuned for Aerial
	Image Sequences},
  booktitle = {Photogrammetric Image Analysis (PIA)},
  year = {2011},
  editor = {Uwe Stilla and Franz Rottensteiner and Helmut Mayer and Boris Jutzi
	and Matthias Butenuth},
  number = {6952},
  series = {LNCS},
  pages = {257--270},
  address = {Munich, Germany},
  month = oct,
  organization = {ISPRS},
  publisher = {Springer, Heidelberg},
  doi = {10.1007/978-3-642-24393-6_22},  
}

@article{milan2016mot16,
  title={MOT16: A benchmark for multi-object tracking},
  author={Milan, Anton and Leal-Taix{\'e}, Laura and Reid, Ian and Roth, Stefan and Schindler, Konrad},
  journal={arXiv preprint arXiv:1603.00831},
  year={2016}
}

@article{kingma2014adam,
  title={Adam: A method for stochastic optimization},
  author={Kingma, Diederik P and Ba, Jimmy},
  journal={arXiv preprint arXiv:1412.6980},
  year={2014}
}

@inproceedings{xutrain,
  title={How To Train Your Deep Multi-Object Tracker},
  author={Xu, Yihong and Osep, Aljosa and Ban, Yutong and Horaud, Radu and Leal-Taix{\'e}, Laura and Alameda-Pineda, Xavier},
  booktitle={Computer Vision and Pattern Recognition},
  year={2020}
}

@article{rakha2007characterizing,
  title={Characterizing driver behavior on signalized intersection approaches at the onset of a yellow-phase trigger},
  author={Rakha, Hesham and El-Shawarby, Ihab and Setti, Jos{\'e} Reynaldo},
  journal={IEEE Transactions on Intelligent Transportation Systems},
  volume={8},
  number={4},
  pages={630--640},
  year={2007},
  publisher={IEEE}
}

@inproceedings{zhang2017real,
  title={Real-time vehicle detection and tracking in video based on faster R-CNN},
  author={Zhang, Yongjie and Wang, Jian and Yang, Xin},
  booktitle={Journal of Physics: Conference Series},
  volume={887},
  number={1},
  pages={012068},
  year={2017},
  organization={IOP Publishing}
}

@book{brunelli2009template,
  title={Template matching techniques in computer vision: theory and practice},
  author={Brunelli, Roberto},
  year={2009},
  publisher={John Wiley \& Sons}
}

@article{chahyati2017tracking,
  title={Tracking people by detection using CNN features},
  author={Chahyati, Dina and Fanany, Mohamad Ivan and Arymurthy, Aniati Murni},
  journal={Procedia Computer Science},
  volume={124},
  pages={167--172},
  year={2017},
  publisher={Elsevier}
}

@inproceedings{huang2017learning,
  title={Learning policies for adaptive tracking with deep feature cascades},
  author={Huang, Chen and Lucey, Simon and Ramanan, Deva},
  booktitle={Proceedings of the IEEE International Conference on Computer Vision},
  pages={105--114},
  year={2017}
}

@inproceedings{wang2016stct,
  title={Stct: Sequentially training convolutional networks for visual tracking},
  author={Wang, Lijun and Ouyang, Wanli and Wang, Xiaogang and Lu, Huchuan},
  booktitle={Proceedings of the IEEE conference on computer vision and pattern recognition},
  pages={1373--1381},
  year={2016}
}

@incollection{huber1992robust,
  title={Robust estimation of a location parameter},
  author={Huber, Peter J},
  booktitle={Breakthroughs in statistics},
  pages={492--518},
  year={1992},
  publisher={Springer}
}

@inproceedings{lukezic2017discriminative,
  title={Discriminative correlation filter with channel and spatial reliability},
  author={Lukezic, Alan and Vojir, Tomas and Cehovin Zajc, Luka and Matas, Jiri and Kristan, Matej},
  booktitle={Proceedings of the IEEE Conference on Computer Vision and Pattern Recognition},
  pages={6309--6318},
  year={2017}
}

@article{wang2017dcfnet,
  title={Dcfnet: Discriminant correlation filters network for visual tracking},
  author={Wang, Qiang and Gao, Jin and Xing, Junliang and Zhang, Mengdan and Hu, Weiming},
  journal={arXiv preprint arXiv:1704.04057},
  year={2017}
}

@article{rastogi2011design,
  title={Design implications of walking speed for pedestrian facilities},
  author={Rastogi, Rajat and Thaniarasu, Ilango and Chandra, Satish},
  journal={Journal of transportation engineering},
  volume={137},
  number={10},
  pages={687--696},
  year={2011},
  publisher={American Society of Civil Engineers}
}

@article{finnis2006field,
  title={Field observations of factors influencing walking speeds},
  author={Finnis, KK and Walton, D},
  journal={Ergonomics},
  year={2006}
}

@article{strayer2004profiles,
  title={Profiles in driver distraction: Effects of cell phone conversations on younger and older drivers},
  author={Strayer, David L and Drew, Frak A},
  journal={Human factors},
  volume={46},
  number={4},
  pages={640--649},
  year={2004},
  publisher={Sage Publications Sage UK: London, England}
}

@inproceedings{hu2018squeeze,
  title={Squeeze-and-excitation networks},
  author={Hu, Jie and Shen, Li and Sun, Gang},
  booktitle={Proceedings of the IEEE conference on computer vision and pattern recognition},
  pages={7132--7141},
  year={2018}
}

@article{lin2013network,
  title={Network in network},
  author={Lin, Min and Chen, Qiang and Yan, Shuicheng},
  journal={arXiv preprint arXiv:1312.4400},
  year={2013}
}

@article{fiaz2019handcrafted,
  title={Handcrafted and deep trackers: Recent visual object tracking approaches and trends},
  author={Fiaz, Mustansar and Mahmood, Arif and Javed, Sajid and Jung, Soon Ki},
  journal={ACM Computing Surveys (CSUR)},
  volume={52},
  number={2},
  pages={1--44},
  year={2019},
  publisher={ACM New York, NY, USA}
}

@article{kalman1960new,
  title={A new approach to linear filtering and prediction problems},
  author={Kalman, Rudolph Emil},
  year={1960}
}

@article{montecarlo2014,
  title={Overview of Bayesian sequential Monte Carlo methods for group and extended object tracking},
  author={Lyudmila Mihaylova; Avishy Y. Carmi; François Septier; Amadou Gning; Sze Kim Pang; Simon Godsilll},
  year={2014}
}

@inproceedings{ma2015hierarchical,
  title={Hierarchical convolutional features for visual tracking},
  author={Ma, Chao and Huang, Jia-Bin and Yang, Xiaokang and Yang, Ming-Hsuan},
  booktitle={Proceedings of the IEEE international conference on computer vision},
  pages={3074--3082},
  year={2015}
}

@article{kuhn1955hungarian,
  title={The Hungarian method for the assignment problem},
  author={Kuhn, Harold W},
  journal={Naval research logistics quarterly},
  volume={2},
  number={1-2},
  pages={83--97},
  year={1955},
  publisher={Wiley Online Library}
}

@inproceedings{yokoyama2005contour,
  title={A contour-based moving object detection and tracking},
  author={Yokoyama, Masayuki and Poggio, Tomaso},
  booktitle={2005 IEEE International Workshop on Visual Surveillance and Performance Evaluation of Tracking and Surveillance},
  pages={271--276},
  year={2005},
  organization={IEEE}
}

@inproceedings{sadeghian2017tracking,
  title={Tracking the untrackable: Learning to track multiple cues with long-term dependencies},
  author={Sadeghian, Amir and Alahi, Alexandre and Savarese, Silvio},
  booktitle={Proceedings of the IEEE International Conference on Computer Vision},
  pages={300--311},
  year={2017}
}

@inproceedings{redmon2016you,
  title={You only look once: Unified, real-time object detection},
  author={Redmon, Joseph and Divvala, Santosh and Girshick, Ross and Farhadi, Ali},
  booktitle={Proceedings of the IEEE conference on computer vision and pattern recognition},
  pages={779--788},
  year={2016}
}

@inproceedings{dalal2005histograms,
  title={Histograms of oriented gradients for human detection},
  author={Dalal, Navneet and Triggs, Bill},
  booktitle={2005 IEEE computer society conference on computer vision and pattern recognition (CVPR'05)},
  volume={1},
  pages={886--893},
  year={2005},
  organization={IEEE}
}

@inproceedings{kraus2020aerialmptnet,
    title={{AerialMPTNet:} Multi-Pedestrian Tracking in Aerial Imagery Using Temporal and Graphical Features},
    author={Maximilian Kraus and Seyed Majid Azimi and Emec Ercelik and Reza Bahmanyar and Peter Reinartz and Alois Knoll},
    year={2020},
    booktitle={International Conference on Pattern Recognition (ICPR)},
}

@INPROCEEDINGS{Shrivastava2016ohem,

  author={A. {Shrivastava} and A. {Gupta} and R. {Girshick}},
  booktitle={2016 IEEE Conference on Computer Vision and Pattern Recognition (CVPR)}, 
  title={Training Region-Based Object Detectors with Online Hard Example Mining}, 
  year={2016},
  volume={},
  number={},
  pages={761-769},}

\end{document}